  \documentclass[final,5p,times,twocolumn]{elsarticle}

\usepackage{amssymb}
\usepackage{comment}
\usepackage{hyperref}
\usepackage{amsfonts}
\usepackage{appendix}
\usepackage{amssymb}
\usepackage{amsthm}
\usepackage{latexsym,amsmath}
\usepackage{graphicx}
\usepackage{float}
\usepackage{subfig}
\usepackage{multirow}
\usepackage{multicol} 
\usepackage{booktabs}
\usepackage{afterpage}
\usepackage{longtable}
\usepackage{xcolor}
\usepackage{comment}
\usepackage{verbatim}
\usepackage{url}
\usepackage{makecell}
\usepackage{caption}

\journal{.}

\begin{document}

\begin{frontmatter}



\title{Review of deep learning models for crypto price prediction: implementation and evaluation}


\author[inst1]{Jingyang Wu} 

\author[inst1]{Xinyi Zhang} 

\author[inst1]{Fangyixuan Huang } 

\author[inst1]{Haochen Zhou} 

\author[inst1]{Rohtiash Chandra}

\affiliation[inst1]{UNSW Sydney, Sydney, Australia}

\begin{abstract} 
There has been much interest in accurate cryptocurrency price forecast models by investors and researchers. Deep Learning models are prominent machine learning techniques that have transformed various fields and have shown potential for finance and economics. Although various deep learning models have been explored for cryptocurrency price forecasting, it is not clear which models are suitable due to high market volatility. In this study, we review the literature about deep learning for cryptocurrency price forecasting and evaluate novel deep learning models for cryptocurrency stock price prediction. Our deep learning models include variants of long short-term memory (LSTM) recurrent neural networks, variants of convolutional neural networks (CNNs), and the Transformer model.  We evaluate univariate and multivariate approaches for multi-step ahead predicting of cryptocurrencies close-price.  We also carry out volatility analysis on the four cryptocurrencies which reveals significant fluctuations in their prices throughout the COVID-19 pandemic. Additionally, we investigate the prediction accuracy of two scenarios identified by different training sets for the models. First, we use the pre-COVID-19 datasets to model cryptocurrency close-price forecasting during the early period of COVID-19. Secondly, we utilise data from the COVID-19 period to predict prices for 2023 to 2024.  Our results show that the convolutional LSTM with a multivariate approach provides the best prediction accuracy in two major experimental settings.
 Our results also indicate that the multivariate deep learning models exhibit better performance in forecasting four different cryptocurrencies when compared to the univariate models. 
\end{abstract}


\end{frontmatter}


\section{Introduction}
\label{sec:sample1}
The traditional financial ecosystem is implemented through a complex set of policies and structural mechanisms that financial institutions utilise to engender currency within an economy \cite{bose2019financial}. The core of this ecosystem is the central bank, treasury, and commercial banking entities which are classified under three primary monetary frameworks: commodity-based \cite{frankel2008fiscal}, commodity-backed \cite{hayek1990denationalisation}, and fiat currency systems \cite{gross2019money}. Triggered by the flaws in these institutions such as inflationary propensities and transactional inefficiencies \cite{boyd2001impact}, the digitisation of currency has become a revolution \cite{milkau2015digitalisation}. Cryptocurrencies aim to rectify the existing system imperfections \cite{chaum1983blind}, such as inflation,  financial stability, transactional efficiency,  and reduced operational costs. A \textit{cryptocurrency} is a \textit{peer-to-peer} digital exchange system where cryptographic techniques are employed to create and distribute units of currency among participants \cite{nakamoto2008bitcoin,manimuthu2019literature}. The cryptocurrency market has seen rapid and unpredictable changes over its relatively brief existence \cite{farell2015analysis}. The security of the cryptocurrency market is ensured by a technology called blockchain \cite{eyal2017blockchain}, which  provides a comprehensive security. In the present year (2024), there are over 5,000 cryptocurrencies and 5.8 million active users in the cryptocurrency industry \cite{jang2017empirical}. Due to its inherent nature of mixing cryptography with a monetary unit, Bitcoin (BTC) became one of the most popular cryptocurrency and received attention in fields such as computer science, economics and cryptography \cite{saad2019toward}. Satoshi Nakamoto pseudonymously introduced Bitcoin and released it as an  open source software in January 2009 \cite{nakamoto2008bitcoin}. The cryptocurrency ecosystem encompassing Bitcoin and Altcoins with tokens such as Civic and BitDegree, marks a significant stride towards a decentralized financial system. The  cryptocurrency ecosystem refers to the broader infrastructure and community that encompasses various cryptocurrencies and blockchain projects, whereas a "token" such as  BitDegree and Civic, serves specific functions within these ecosystems, often facilitating access to services or representing certain assets. Unlike Bitcoin, which is primarily a digital currency intended for transactions and value storage, BitDegree serves a distinct purpose by focusing on education, offerring tokens as incentives within its educational platform. Nevertheless, due to its decentralised nature and absence of governmental support, the cryptocurrency market is susceptible to significant fluctuations in value and the formation of pricing bubbles \cite{corbet2018datestamping}.

The inherent volatility of cryptocurrencies featuring transaction volume fluctuations and price variability, complicates the predictive analysis of cryptocurrency prices \cite{bhosale2018volatility}. However, volatility \cite{katsiampa2019empirical} makes it a profitable market for speculation as the sourse of potencial gain. The prominent cryptocurrencies such as Bitcoin (BTC), Ethereum (ETH), and Litecoin (LTC)  differ based on valuation, transaction speed, usage, and volatility \cite{elendner2016cross}. Identifying the precise catalysts for these price trends in the cryptocurrency domain remains elusive due to the sector's pronounced volatility. Nevertheless, the market value of cryptocurrencies is projected to increase in the future, with an expected compound annual growth rate of 11.1\% \cite{seabe2023forecasting}. Meanwhile, the financial audit sector is evolving to integrate cryptocurrencies as a valid transaction medium. Investors have encountered challenges in previous instances due to price bubbles resulting in extreme fluctuations \cite{kyriazis2020systematic}. In order to surmount these obstacles, it is imperative to have a dependable model that can aid market participants in identifying trends and generating accurate predictions. Predicting cryptocurrency prices with precision is difficult due to its sensitivity to multiple factors, including government policies, technology advancements, public perception, and world events \cite{ammer2022deep}. Muarry et al. \cite{murray2023forecasting} highlights the inherent difficulties in predicting the pricing of cryptocurrencies because of their high volatility, decentralised nature, and other distinctive features such as transaction speed and variations in their ecosystems. 

Several researchers are affirming the correlation between cryptocurrencies and other domains such as the economics, finance, the internet, and even politics.  
Wang et al. \cite{wang2023machine} presented an analysis using machine learning models and revealed a strong correlation between cryptocurrencies and their intrinsic features (e.g., lagged volatility, previous trading information). Kyriazis \cite{kyriazis2019survey} studied spillover effects in cryptocurrency markets, emphasising Bitcoin’s role using statistical models such as vector autoregression (VAR) \cite{stock2001vector} and generalized autoregressive conditional heteroskedasticity (GARCH) \cite{duan1995garch} to explain inter-market dynamics. Huynh et al. \cite{huynh2020small} revealed that gold can be used as a reliable tool to reduce the risk associated with unpredictable changes in the cryptocurrency market when utilized as a separate form of currency. However, investors are enthusiastic and also cautious due  to the highly volatile cryptocurrency market. Machine learning, along with deep learning models are promising for cryptocurrency due to prediction capabilities and the ability to model multimodal \cite{baltruvsaitis2018multimodal}, spatiotemporal data \cite{wang2020deep}, and time series forecasting \cite{lim2021time}. 

Machine learning and deep learning models have shown great potential in temporal forecasting problems for various domains, such as climate extremes \cite{jacques2022deep}, energy \cite{mahjoub2022predicting}, and financial time series \cite{chandra2021bayesian}. Deep learning models can assist in forecasting future cryptocurrency prices, although there are challenges due to nonlinear and volatile nature of the time series. Many researchers are keen to use long short-term memory (LSTM) and its variants to predict cryptocurrencies \cite{livieris2020ensemble,ferdiansyah2019lstm,wu2018new}. Deep learning methods such as LSTM recurrent neural networks \cite{hochreiter1997long,yu2019review}, convolutional neural networks (CNN) \cite{jiang2017cryptocurrency}, and Transformer models \cite{sridhar2021multi} are also promising for predicting cryptocurrencies. Chandra et al. \cite{chandra2021evaluation} led a comparative analysis of various deep learning models for multi-step forward time series prediction. 
A myriad of factors, both internal and external, such as the trading volume, market beta, and volatility, play a critical role in determining cryptocurrency value. Therefore, we need to utilise cryptocurrencies that are highly correlated for deep learning models and access univariate and multivariate deep learning models.

In this paper, we provide a detailed review of the literature on crypto-price forecasting using deep learning models and then evaluate novel deep learning models for cryptocurrency price forecasting. Specifically, we utilise variants of long short-term memory (LSTM) recurrent neural networks, variants of convolutional neural networks (CNNs), and the Transformer model.  We evaluate univariate and multivariate approaches for multi-step ahead predicting of cryptocurrencies close-price. Our results show that the univariate  LSTM model variants perform best for cryptocurrency predictions. We also carry out volatility analysis on the four cryptocurrencies and investigate the prediction accuracy of two scenarios identified by different training sets for the models. First, we use the pre-COVID-19 datasets to model cryptocurrency close-price forecasting during the early period of COVID-19. Secondly, we utilise data from the COVID-19 period to predict prices for 2023 to 2024. We investigate the effect of univariate and multivariate models, where the multivariate model features the Gold price, close, open, and high price of the crypto being predicted and a highlighted correlated crypto price index. 


The rest of the paper is organised as follows. Chapter \ref{ch2} provides a comprehensive overview and analysis of previous research and literature relating to the topic. Chapter \ref{ch3} provides the framework that compares  selected deep learning models. Chapter \ref{ch4}  presents the experiments and results. Chapter \ref{ch5} presents a discussion, and Chapter \ref{ch6} concludes the paper.


\section{Review } \label{ch2}

Forecasting financial time series is highly favoured by researchers in both academic and financial sectors due to its wide applications and significant influence. Machine learning and deep learning have paved the way for numerous models, leading to a large body of published research. Among these areas of interest, cryptocurrency price prediction stands out. This section offers an in-depth overview of how machine learning and deep learning are applied to financial time series forecasting, especially for predicting cryptocurrency prices, without using complex terminology.

\subsection{Financial time series prediction}
Financial time series forecasting had an emphasis on predicting asset prices \cite{tsay2005analysis}. Although there are diverse methodologies, the key focus has been on predicting the future movements of the underlying asset with deep learning models \cite{fischer2018deep}. This field covers a variety of subjects including forecasting of stock prices, index prediction, forex price prediction, as well as predictions for commodities (such as oil and gold), bond prices, volatility, and cryptocurrency prices \cite{sezer2020financial}. Despite the wide range of topics, the underlying principles applied in these forecasts remain uniformly applicable across all categories.

Research within financial time series forecasting is broadly segregated into two categories based on precise price forecasting and trend (directional movement) forecasting \cite{plakandaras2015market}. Although exact price prediction aligns with regression tasks, the primary goal in numerous financial forecasting projects is not the accurate prediction of prices, but rather the correct identification of price movement direction. This shifts the emphasis towards trend prediction, or determining the directional change in prices, marking it as a more critical area of investigation compared to pinpoint price forecasting. Hence, trend prediction is approached as a classification issue. Some analyses focus on binary outcomes, addressing only upward/downward movements \cite{nabipour2020predicting}, while others incorporate a third class (neutral option), thus constituting a 3-class problem \cite{kong2021predicting}.

In recent years, researchers have utilised machine learning and deep learning for the analysis of financial time series data.  Nabipour et al. \cite{nabipour2020predicting} conducted a comparative analysis of deep learning models (simple recurrent neural network (RNNs) \cite{elman1990finding} and   LSTM networks \cite{hochreiter1997long}) with machine learning models for stock market trend prediction, demonstrating the superior accuracy of deep learning. Mehtab et al. \cite{mehtab2021stock} enhanced NIFTY-50  \footnote{\url{https://www.nseindia.com/}} Indian stock index prediction using LSTM models with the grid-searching and walk-forward validation and achieved notable accuracy.  NIFTY-50 represents the weighted average of 50 of the top companies listed on the National Stock Exchange (NSE) of India. Rezaei et al. \cite{rezaei2021stock} combined deep learning with frequency decomposition methods, including  empirical mode decomposition (EMD) \cite{rilling2003empirical}, and complete ensemble empirical mode decomposition (CEEMD) \cite{torres2011complete} to predict stock prices and demonstrated  effectiveness of  CEEMD-CNN-LSTM and EMD-CNN-LSTM. Jing et al. \cite{jing2021hybrid} developed a hybrid model that merges deep learning with investor sentiment analysis, utilising CNN for sentiment classification and LSTM for stock price prediction, demonstrating enhanced predictive accuracy for stock prices. Mehtab and Sen \cite{mehtab2022stock} used a blend of machine learning and deep learning models  with walk-forward validation and grid-search technique for precise short-term forecasting rather than long-term trends of NIFTY-50, offering valuable insights for short-term traders. Li and Pan \cite{li2022novel} enhanced stock price prediction accuracy by employing an ensemble deep learning model that leveraged stock prices and news data, using LSTM and gated recurrent unit (GRU) networks. Kanwal et al. \cite{kanwal2022bicudnnlstm} introduced a hybrid deep learning model combining bidirectional LSTM and one-dimensional CNN for stock price prediction, achieving higher accuracy and efficiency on five distinct stock datasets. Swathi et al. \cite{swathi2022optimal} presented a novel model for stock price prediction, leveraging Twitter sentiment analysis with an impressive accuracy of 94.73\%, showcasing its effectiveness over traditional and other deep learning methods. Ben Ameur et al. \cite{ben2023forecasting} utilized deep learning models (LSTM, GRU, RNN, and CNNs) to  forecast commodity prices for the Bloomberg Index, demonstrating LSTM models superior performance. Baser et al. \cite{baser2023gold} evaluated gold price prediction using tree-based models, including Decision Trees, AdaBoost, Random Forest, Gradient Boosting, and XGBoost. They  demonstrated XGBoosts superior accuracy through technical indicators analysis. Deepa et al. \cite{deepa2023machine} used statistical and machine learning  models for prediction of cotton prices in India and reported that boosted decision tree regression provided the highest accuracy. 
Zhao and Yang \cite{zhao2023deep} proposed an integrated deep learning framework for stock price movement prediction, which combined sentiment analysis with deep learning models and got enhanced prediction accuracy by incorporating both market data and investor sentiment.

Table \ref{t20} provides a list of sample studies focused on using traditional statistical and machine learning methods to predict cryptocurrency trends. We report various models with error measures such as the mean absolute error (MAE), mean absolute percentage error (MAPE) and root mean squared error(RMSE). We also mention the time periods of data used in these literatures.


\begin{table*}[htbp!]
    \centering
    \footnotesize 
    \begin{tabular}{c c c c c}
        \hline \hline
        Methods & Data & \makecell[c]{Target\\predictor}& \makecell[c]{Time range\\(month/day/year)} & Metric \\
        \hline
        \makecell[c]{LSTM\\grid-search\cite{mehtab2021stock}}& NIFTY 50 index & \makecell[c]{Price\\ prediction} & \makecell[c]{12/29/2014-\\07/31/2020} & RMSE \\   
        \hline
        \makecell[c]{MLP,\\RNN,\\LSTM\cite{nabipour2020predicting}} & Stock market trends & \makecell[c]{Trend\\ prediction} & \makecell[c]{11/01/2009-\\11/01/2019} & \makecell[c]{F1-Score\\Accuracy\\ROC-AUC} \\    
        \hline
        \makecell[c]{LSTM, CNN,\\empirical mode\\decomposition,\\CEEMD\cite{rezaei2021stock}} & Stock prices & \makecell[c]{Price\\ prediction} & \makecell[c]{01/01/2010-\\09/01/2019} & \makecell[c]{RMSE\\MAE\\MAPE}  \\    
        \hline
        \makecell[c]{CNN, LSTM\cite{jing2021hybrid}} & Stock prices & \makecell[c]{Price\\ prediction} & \makecell[c]{01/01/2017-\\07/01/2019} & \makecell[c]{MAPE}  \\
        \hline
        \makecell[c]{Linear Regression,\\Bagging, XGBoost\\Ranform Forests\\MLP, SVM, LSTM\cite{mehtab2022stock}} & NSE stock prices & \makecell[c]{Short-term price\\prediction} & - & \makecell[c]{Comparative\\ analysis}  \\
        \hline
        \makecell[c]{LSTM with\\Sentiment\\Analysis\cite{li2022novel}} & Stock prices & \makecell[c]{Price\\ Prediction} & \makecell[c]{12/31/2017-\\06/01/2018} & \makecell[c]{MSE\\Precision\\Recall\\F1-Score} \\
        \hline
        \makecell[c]{BD-LSTM,\\1D-CNN\cite{kanwal2022bicudnnlstm}} & Stock prices & \makecell[c]{Price\\prediction} & \makecell[c]{01/01/2000-\\12/31/2020} & \makecell[c]{Accuracy\\Efficiency} \\
        \hline
        \makecell[c]{Teaching\\Learning Based\\Optimization\\LSTM\cite{swathi2022optimal}} & Stock prices & \makecell[c]{Price\\prediction} & \makecell[c]{-} & \makecell[c]{Accuracy\\Precision\\Recall\\F1-Score}  \\
        \hline
        \makecell[c]{LSTM,\\Gated Recurrent Units,\\RNN, CNN\cite{ben2023forecasting}} & \makecell[c]{Bloomberg\\Commodity\\Index} & \makecell[c]{Price\\prediction} & \makecell[c]{01/01/2002-\\12/31/2020} & Accuracy  \\
        \hline
        \makecell[c]{Decision Tree,\\AdaBoost,\\Random Forests,\\Gradient Boosting, XGBoost\cite{baser2023gold}} & Gold prices & \makecell[c]{Price\\ prediction} & \makecell[c]{11/18/2011-\\01/01/2019} & \makecell[c]{MAE\\MSE\\RMSE\\R2 Score} \\
        \hline
        \makecell[c]{Logistic Regression,\\Bayesian Linear Regression,\\Boosted Decision\\Tree Regression,\\Random  Forest Regression,\\Poisson Regression\cite{deepa2023machine}} & \makecell[c]{Agriculture\\material\\prices} & \makecell[c]{Price\\prediction} & - & \makecell[c]{MAE\\RMSE\\RAE\\R square}  \\
        \hline
        \makecell[c]{LSTM, Ensemble CNN,\\Denoising Autoencoder,\\Sentiment Analysis\cite{zhao2023deep}} & \makecell[c]{Stock prices\\and sentiment} & \makecell[c]{Price\\movement} & \makecell[c]{01/01/2002-\\12/31/2020} & \makecell[c]{RMSE}  \\
        \hline \hline
    \end{tabular}
    \caption{Sample studies focused on financial time series forecasting}
    \label{t20}
\end{table*}

\subsection{Cryptocurrency prediction}

Some researchers have employed machine learning models, such as  simple neural networks (SNN) also known as backpropagation and artificial neural networks \cite{almeida2015bitcoin}, support vector machines (SVM) \cite{mallqui2019predicting},  genetic algorithm-based SNN \cite{quek2022new},  and neuroevolution of augmenting topologies (NEAT) \cite{radityo2017prediction} which  evolves both architecture and neural network parameters.

Next, we review some of the machine learning models that are pivotal in predicting cryptocurrency prices. Greaves and Au \cite{greaves2015using} demonstrated the superiority of neural networks over linear regression, logistic regression, and support vector machines (SVM) \cite{cortes1995support} for Bitcoin price prediction. Sovbetov \cite{sovbetov2018factors} examined the effect of market factors by using autoregressive distributed lag (ARDL) and the S\&P50 Index on various cryptocurrencies. Guo et al. \cite{guo2018bitcoin} improved short-term Bitcoin volatility forecasting with temporal mixture models, outperforming traditional methods. Akcora et al. \cite{akcora2018forecasting} investigated the predictive Granger causality of chainlets and identify certain types of chainlets that exhibit the highest predictive influence on Bitcoin price and investment risk.  Roy et al. \cite{roy2018bitcoin}  used ARIMA, Autoregressive, and Moving Average  models  in forecasting short-run volatility in Bitcoin's weighted costs. Derbentsev et al. \cite{derbentsev2019forecasting} compared binary autoregressive tree (BART),  ARIMA, and autoregressive fractional integrated moving average (ARFIMA) models for forecasting Bitcoin, Ethereum, and Ripple prices where  BART had best accuracy. Kumar et al. \cite{aanandhi2021cryptocurrency} and Latif et al. \cite{latif2023comparative} examine the effectiveness of LSTM and ARIMA models in the short-term prediction of BTC prices, demonstrating that while ARIMA models can capture the general trend, LSTM models excel in predicting both the direction and magnitude of price movements, highlighting the potential of deep learning in financial market predictions. Maleki et al. \cite{maleki2023bitcoin} used machine learning models including linear regression, gradient boosting regressor(GBR), support vector regressor(SVR),  random forest regressor (RFR) and ARIMA  in predicting Bitcoin prices, suggesting new investment strategies in the cryptocurrency market.

Table \ref{t21} provides a list of studies focused on using traditional statistical and machine learning methods to predict cryptocurrency trends. The table reports metrics such as the mean absolute error (MAE), mean absolute percentage error (MAPE) and root mean squared error(RMSE). We also mention the time periods of data used in these papers.

\begin{table*}[htbp]
    \centering
    \footnotesize
    \begin{tabular}{c c c c c}
    \hline \hline
    Methods & \makecell[c]{Cryptocurrency\\(type)} & \makecell[c]{Target\\predictor} & \makecell[c]{Time range\\(month/day/year)} & Metric \\
    \hline
    \makecell[c]{Linear regression,\\Logistic regression,\\Neural Networks, SVM\cite{greaves2015using}} & BTC & Future price & \makecell[c]{prior-\\07/04/2013} & \makecell[c]{MSE\\Accuracy} \\
    \hline
    \makecell[c]{Autoregressive\\Distributed Lag\cite{sovbetov2018factors} } & \makecell[c]{BTC\\ETH\\Dash\\LTC\\Monero} & \makecell[c]{Short-Long\\term price} & \makecell[c]{01/01/2010-\\01/12/018} & \makecell[c]{ADF \\test price} \\
    \hline
    \makecell[c]{Temporal\\mixture models\cite{guo2018bitcoin} } & BTC & \makecell[c]{Short-term\\volatility} & \makecell[c]{09/01/2015-\\04/01/2017} & \makecell[c]{RMSE\\MAE} \\
    \hline
    \makecell[c]{k-Chainlets\cite{akcora2018forecasting}} & BTC & Close Price & \makecell[c]{01/01/2009-\\01/01/2018} & \makecell[c]{RMSE\\ wallet gain} \\
    \hline
    \makecell[c]{ARIMA,\\Autoregressive,\\Movingaverage\cite{roy2018bitcoin}} & BTC & Market price & \makecell[c]{07/31/2013-\\08/01/2017} & \makecell[c]{RMSE} \\
    \hline
    \makecell[c]{BART,\\ARIMA,\\ARFIMA\cite{derbentsev2019forecasting} } & \makecell[c]{BTC\\Ripple\\ETH} & \makecell[c]{Short-term \\price} & \makecell[c]{01/01/2017-\\03/01/2019} & RMSE \\
    \hline
    \makecell[c]{ARIMA,\\LSTM\cite{aanandhi2021cryptocurrency}} & ETH & Close Price & \makecell[c]{01/01/2016-\\12/31/2021} & \makecell[c]{MSE} \\
    \hline
    \makecell[c]{ARIMA,\\LSTM\cite{latif2023comparative}} & BTC & \makecell[c]{Short-term \\price} & \makecell[c]{12/21/2020-\\12/21/2021} & \makecell[c]{MAPE\\MAE\\RMSE} \\
    \hline
    \makecell[c]{Logistic Regression,\\Gradient boosting regressor,\\SVR,\\Random forest regressor,\\ARIMA\cite{maleki2023bitcoin}} & \makecell[c]{BTC\\ETH\\ZEC\\LTC} & Close price & \makecell[c]{04/01/2018-\\03/31/2019} & \makecell[c]{MSE\\MAPE\\MAE\\AIC\\BIC} \\
    \hline \hline
    \end{tabular}
    \caption{Sample studies focused on using machine learning and time series methods in cryptocurrency forecasting}
    \label{t21}
\end{table*}

\subsection{Deep learning models for cryptocurrency prediction}

In recent years,  deep learning models have been prominent in the prediction of cryptocurrencies, as follows. Jiang and Liang \cite{jiang2017cryptocurrency} combined CNNs with reinforcement learning \cite{kaelbling1996reinforcement} for portfolio management utilising historical cryptocurrency pricing data to allocate assets optimally within specified portfolio constraints. Wu et al. \cite{wu2018new} improved Bitcoin prediction accuracy by using autoregressive  characteristics in an LSTM network, outperforming standard LSTM. Lee et al. \cite{lee2018generating} introduced a novel approach employing inverse reinforcement learning coupled with agent-based modeling for Bitcoin price prediction. Ly et al. \cite{ly2018applying} employed LSTM networks to predict Bitcoin trends, demonstrating the models' capability to forecast price changes and classify market movements with varying degrees of accuracy.  Saad et al. \cite{saad2019toward} found LSTM to be the most accurate in forecasting Bitcoin prices compared to various machine learning models
Patel et al. Lucarelli and Borrotti \cite{lucarelli2019deep} investigated automated cryptocurrency trading using deep reinforcement learning, employing double deep Q-learning networks trained by Sharpe ratio rewards, which outperformed traditional models in Bitcoin trading. Lahmiri and Bekiros \cite{lahmiri2019cryptocurrency} compared  LSTM  networks with generalized regression neural networks (GRNN) to forecast cryptocurrency prices, revealing the chaotic dynamics and fractality in digital currencies' time series.  Patel et al. \cite{patel2020deep} introduced a hybrid LSTM with gated recurrent unit model for Litecoin and Monero and achieved more accuracy than a simple LSTM model. Livieris et al. \cite{livieris2020ensemble} combined deep learning and ensemble learning to forecast trends and prices of Bitcoin, Ethereum, and Ripple. LSTM, bidirectional LSTM, and CNN models demonstrated the capability to deliver precise and dependable predictions.  Marne et al. \cite{marne2020predicting} used RNN and LSTM models to predict Bitcoin prices that showed better results than machine learning models. Nasekin and Chen \cite{nasekin2020deep} analysed cryptocurrency investor sentiment using CNN for sentiment classification and index construction from StockTwits messages. Sridhar and Sanagavarapu \cite{sridhar2021multi} employed a Transformer model for Dogecoin price prediction demonstrating the model's capability to capture both short-term and long-term dependencies effectively. Betancourt and Chen \cite{betancourt2021deep} propose the utilization of \textit{deep reinforcement learning} (DRL) \cite{arulkumaran2017deep}  for the dynamic management of cryptocurrency asset portfolios, accommodating portfolios comprising an evolving number of cryptocurrency assets. Shahbazi and Byun \cite{shahbazi2021improving} applied reinforcement learning for forecasting Litecoin and Monero market values. D’Amato et al. \cite{d2022deep} employed a Jordan RNN to enhance the prediction of cryptocurrency volatility, demonstrating superior accuracy over traditional machine learning models for Bitcoin, Ripple, and Ethereum. Schnaubelt \cite{schnaubelt2022deep} applied reinforcement learning to develop cryptocurrency trading strategies. Parekh et al. \cite{parekh2022dl} combined LSTM and sentiment analysis to predict cryptocurrency prices. The study integrated market sentiments from social media for enhanced forecasting accuracy. Kim et al. \cite{kim2022deep} applied a self-attention-based multiple  LSTM model and improved the prediction accuracy for Bitcoin. Goutte et al. \cite{goutte2023deep} used LSTM networks with technical analysis to enhance cryptocurrency trading strategies, particularly focusing on Bitcoin.

Table \ref{t22} provides an overview of sample research that focuses on applying deep learning techniques to forecast the trend of cryptocurrencies. The table reports metrics such as the mean absolute error (MAE), mean absolute percentage error (MAPE) and root mean squared error(RMSE). We also mention the time periods of data used in these papers.

\begin{table*}[htbp!]
    \centering
    \footnotesize
    \begin{tabular}{c c c c c}
    \hline \hline
    \makecell[c]{Deep learning\\techniques} & \makecell[c]{Cryptocurrency\\(type)} & \makecell[c]{Target\\predictor} & \makecell[c]{Time range\\(month/day/year)} & Metric \\
    \hline
    \makecell[c]{CNN,\\Reinforcement Learning\cite{jiang2017cryptocurrency}} & \makecell[c]{BTC\\ETH\\XRP} & Close Price & \makecell[c]{01/01/2018-\\02/28/2019} & \makecell[c]{RMSE\\ Accuracy\\AUC\\F1} \\
    \hline
    \makecell[c]{LSTM with\\autoregressive\\characteristics\cite{wu2018new}} & BTC & \makecell[c]{Short-Long\\term price} & \makecell[c]{01/01/2018-\\07/28/2018} & \makecell[c]{MSE\\RMSE\\MAPE} \\
    \hline
    \makecell[c]{inverse Reinforcement \\Learning\\Agent-based Model\cite{lee2018generating}} & \makecell[c]{BTC} & \makecell[c]{Close Price} & \makecell[c]{09/01/2016-\\07/31/2017} & \makecell[c]{MAE\\ MSE\\RMSE\\MAPE} \\
    \hline
    \makecell[c]{RNN, LSTM\cite{ly2018applying}} & BTC & \makecell[c]{Trend\\Prediction} & - & - \\
    \hline
    \makecell[c]{Reinforcement Learning,\\LSTM,\\Conjugate Gradient\cite{saad2019toward}} & \makecell[c]{BTC\\ETC} & \makecell[c]{Close price} & \makecell[c]{10/01/2015-\\05/01/2018} & \makecell[c]{RMSE\\MAE} \\
    \hline
    \makecell[c]{Deep Reinforecemt\\Learning\cite{lucarelli2019deep} } & \makecell[c]{BTC} & \makecell[c]{Trading} & \makecell[c]{-} & \makecell[c]{Profit-based\\Metrics} \\
    \hline
    \makecell[c]{Chaotic\\Neural\\Networks\cite{lahmiri2019cryptocurrency}} & \makecell[c]{BTC\\DASH\\XRP} & \makecell[c]{Price\\ Forecasting} & \makecell[c]{07/16/2010-\\10/01/2018} & - \\
    \hline
    \makecell[c]{LSTM with\\Gated Recurrent Units\cite{patel2020deep}} & \makecell[c]{LTC\\Monero} & \makecell[c]{Short-Long\\term price} & \makecell[c]{30/01/2015-\\02/23/2020} & \makecell[c]{Accuracy} \\
    \hline
    \makecell[c]{LSTM,\\BD-LSTM,\\CNN\cite{livieris2020ensemble}} & \makecell[c]{BTC\\ETH\\XRP} & \makecell[c]{Short-Long\\term price} & \makecell[c]{01/01/2018-\\08/31/2019} & \makecell[c]{MSE\\ RMSE\\MAPE} \\
    \hline
    \makecell[c]{RNN, LSTM\cite{marne2020predicting}} & \makecell[c]{BTC} & \makecell[c]{Close Price} & \makecell[c]{01/01/2014-\\01/31/2019} & RMSE \\
    \hline
    \makecell[c]{CNN\cite{nasekin2020deep}} & \makecell[c]{Various} & \makecell[c]{Sentiment\\Analysis} & \makecell[c]{03/01/2013-\\05/31/2018} & - \\
    \hline
    Transformer\cite{sridhar2021multi} & DOGE & Close Price & \makecell[c]{07/05/2019-\\04/28/2021} & \makecell[c]{Accuracy\\R-squared} \\
    \hline
    \makecell[c]{Deep Reinforment\\Learning\cite{betancourt2021deep}} & \makecell[c]{Various} & \makecell[c]{Portfolio\\Management} & \makecell[c]{08/17/2017-\\11/01/2019} & \makecell[c]{Total Return\\Sharpe Ratio} \\
    \hline
    Reinforment Learning\cite{shahbazi2021improving} & \makecell[c]{LTC\\Monero} & \makecell[c]{Price Prediction} & \makecell[c]{2016-2020} & \makecell[c]{MAE\\MSE\\RMSE\\MAPE}   \\
    \hline
    Jordan RNN\cite{d2022deep} & \makecell[c]{BTC\\XRP\\ETH} & \makecell[c]{Volatility} & \makecell[c]{04/28/2013-\\12/15/2019} & \makecell[c]{MSE\\MAPE} \\
    \hline
    \makecell[c]{Deep Reinforcement\\Learning\cite{schnaubelt2022deep}} & \makecell[c]{Various} & \makecell[c]{Limit\\Order\\Placement} & \makecell[c]{01/01/2018-\\06/30/2019} & \makecell[c]{Total Return\\Sharpe Ratio} \\
    \hline
    \makecell[c]{LSTM\\sentiment analysis\cite{parekh2022dl}} & \makecell[c]{Dash\\BTC-Cash} & \makecell[c]{Price\\Prediction} & - & \makecell[c]{MSE\\MAE\\MAPE} \\
    \hline \hline
    \end{tabular}
    \caption{Sample studies focused on deep learning methods in cryptocurrency forecasting}
    \label{t22}
\end{table*}

\subsection{Cryptocurrency volatility and prediction}

Several researchers concentrate on analyzing and predicting the volatility of cryptocurrencies. Volatility in the cryptocurrencies market is a significant factor that influences numerous decisions in business and finance \cite{yen2021economic}. Recently, there has been identification of volatility spillovers between the cryptocurrency market and other financial markets \cite{d2022deep}. Katsiampa \cite{katsiampa2019empirical} employed an Asymmetric Diagonal BEKK model to examine the volatility dynamics in the cryptocurrency market, revealing significant interdependencies and responsiveness to major news in the volatility levels of major cryptocurrencies such as Bitcoin, Ether, Ripple, Litecoin, and Stellar Lumen. Woebbeking \cite{woebbeking2021cryptocurrency} developed the CVX index using a model-free approach derived from cryptocurrency option prices, unveiling that cryptocurrency volatility often diverges from traditional financial markets, and is distinctly reactive to major market events. Yen and Cheng \cite{yen2021economic} utilized stochastic volatility models to analyze the impact of the Economic Policy Uncertainty (EPU) index on cryptocurrency volatility, finding that China's EPU uniquely predicts the volatility of Bitcoin and Litecoin, suggesting these cryptocurrencies might serve as hedging tools against EPU risks. Cross, Hou, and Trinh \cite{cross2021returns} utilized a time-varying parameter model to explore the returns and volatility of cryptocurrencies during the 2017–18 bubble, highlighting a significant risk premium effect in Litecoin and Ripple and identifying adverse news effects as key drivers of the 2018 crash across Bitcoin, Ethereum, Litecoin, and Ripple. Ftiti, Louhichi, and Ben Ameur \cite{ftiti2023cryptocurrency} utilized heterogeneous autoregressive (HAR) models with high-frequency data to explore cryptocurrency volatility during the COVID-19 pandemic. Their findings underscore the predictive superiority of models incorporating both positive and negative semi-variances, especially during the crisis, suggesting these models can effectively capture the asymmetric dynamics of market volatility. Yin, Nie, and Han \cite{yin2021understanding} applied the Generalized Autoregressive Conditional Heteroskedasticity - Mixed Data Sampling (GARCH-MIDAS) model to explore the influence of oil market shocks on the volatility of Bitcoin, Ethereum, and Ripple. Their analysis revealed that oil market shocks, both supply and demand types, significantly affect the long-term volatility of these cryptocurrencies, thereby suggesting potential hedging capabilities against oil-induced economic uncertainties.

There is also some research on the prediction of cryptocurrency volatility. Catania, Grassi, and Ravazzolo \cite{catania2018predicting} employed a score-driven Generalized Hyperbolic Skew Student's t (GHSKT) model to analyze and predict the volatility of Bitcoin, Ethereum, Litecoin, and Ripple. They demonstrated that accounting for long memory and asymmetric reactions to past shocks enhances the model’s predictive accuracy significantly across various forecast horizons. Catania and Grassi \cite{catania2022forecasting} further employed the GHSKT model demonstrating that the model's ability to incorporate higher-order moments and leverage effects significantly enhances the accuracy of volatility forecasts across various cryptocurrencies. Ma et al. \cite{ma2020cryptocurrency} employed the Markov Regime-Switching Mixed Data Sampling (MRS-MIDAS) model to forecast cryptocurrency volatility, particularly focusing on Bitcoin. They enhanced the standard MIDAS approach by incorporating jump-driven time-varying transition probabilities, which allowed the model to capture dynamic changes in volatility states influenced by market jumps.

Table \ref{t23} provides an overview of sample research that focuses on cryptocurrency volatility and prediction. The table reports metrics such as the mean absolute error (MAE), mean absolute percentage error (MAPE) and root mean squared error(RMSE). We also mention the time periods of data used in these papers.


\begin{table*}[htbp!]
    \centering
    \footnotesize
   
    \begin{tabular}{c c c c c}
    \hline \hline
    Methods & Data & \makecell[c]{Target\\predictor} & \makecell[c]{Time range\\(month/day/year)} & \makecell[c]{Metric} \\
    \hline
    \makecell[c]{Asymmetric Diagonal,\\BEKK model\cite{katsiampa2019empirical}} & \makecell[c]{BTC\\ETH\\XRP\\LTC\\XLM} & \makecell[c]{Volatility\\dynamics} & \makecell[c]{08/07/2015-\\02/10/2018} & \makecell[c]{Past squared errors,\\past conditional volatility} \\
    \hline
    \makecell[c]{Model-free volatility,\\CVX index\cite{woebbeking2021cryptocurrency}} & \makecell[c]{BTC} & \makecell[c]{Volatility\\dynamics} & \makecell[c]{02/06/2020-\\07/06/2021} & \makecell[c]{Not specified} \\
    \hline
    \makecell[c]{Stochastic volatility\\models\cite{yen2021economic}} & \makecell[c]{BTC\\LTC} & \makecell[c]{EPU impact on\\cryptocurrency\\volatility} & \makecell[c]{02/2014-\\06/2019} & \makecell[c]{Not specified} \\
    \hline
    \makecell[c]{Time-varying\\parameter\\stochastic volatility\\ model\cite{cross2021returns}} & \makecell[c]{BTC\\ETH\\LTC\\XRP} & \makecell[c]{Returns and\\volatility dynamics} & \makecell[c]{01/2017-\\01/2019} & \makecell[c]{Forecast accuracy\\MSFE\\ALPL} \\
    \hline
    \makecell[c]{Heterogeneous\\autoregressive\\ models\cite{ftiti2023cryptocurrency}} & \makecell[c]{BTC\\ETH\\ETC\\XRP} & \makecell[c]{Volatility\\forecasting} & \makecell[c]{04/01/2018-\\06/30/2020} & \makecell[c]{MSE\\MAE\\MAPE} \\
    \hline
    \makecell[c]{GARCH-MIDAS\\model\cite{yin2021understanding}} & \makecell[c]{BTC\\ETH\\XRP} & \makecell[c]{Impact of oil\\market shocks\\on volatility} & \makecell[c]{04/28/2013-\\12/31/2018} & \makecell[c]{MAE\\MAPE\\RMSE} \\
    \hline
    \makecell[c]{Score-driven\\Generalized Hyperbolic\\Skew Student's t\\ model\cite{catania2018predicting}} & \makecell[c]{BTC\\ETH\\LTC\\XRP} & \makecell[c]{Volatility\\forecasting} & \makecell[c]{04/29/2013-\\12/01/2017} & \makecell[c]{Quasi-Like\\loss function} \\
    \hline
    \makecell[c]{Score-driven\\Generalized Hyperbolic\\Skew Student's t\\ model\cite{catania2022forecasting}} & \makecell[c]{BTC\\ETH\\LTC\\XRP} & \makecell[c]{Volatility\\forecasting} & \makecell[c]{-} & \makecell[c]{MSE\\MAE\\MAPE} \\
    \hline
    \makecell[c]{Markov\\Regime-Switching,\\Mixed Data\\ Sampling model\cite{ma2020cryptocurrency}} & \makecell[c]{BTC} & \makecell[c]{Volatility\\forecasting} & \makecell[c]{03/01/2013-\\09/29/2018} & \makecell[c]{Quasi-Like\\loss function\\MSE\\MAE} \\
    \hline
    \makecell[c]{GARCH-MIDAS\\model\cite{wei2023cryptocurrency}} & \makecell[c]{Gold\\Silver} & \makecell[c]{Cryptocurrency\\ uncertainty impact\\on precious metal\\ volatility} & \makecell[c]{01/02/2014-\\05/13/2022} & \makecell[c]{Diebold-Mariano test\\R-square\\Model Confidence Set test\\Direction-of-Change rate test} \\
    \hline \hline
    \end{tabular}
    \caption{Sample studies focused on cryptocurrency volatility and prediction}
    \label{t23}
\end{table*}

 \section{Methodology: Implementation and Evaluation}
\subsection{Conventional models}
\subsubsection{ARIMA}
The ARIMA model, often known as the Box-Jenkins model \cite{box2015time}, is a commonly used statistical/econometric model for forecasting time series data. The ARIMA model consists of three components: autoregressive (AR), integrated (I), and moving average (MA). The integrated component represents the amount of differencing required to transform the series data into a stationary representation. The autoregressive component describes the relationship between the present value of a time series and its previous values, capturing their correlation. The moving average component indicates the correlation between the current observation and its previous error term. This component assists the model in capturing stochastic variations in the time series.  The three components constitute the three parameters $p$, $d$, and $q$ in the model. $p$ represents the number of lag observations in the autoregressive part. $d$ is the order of differencing, which forms the integrated part, and $q$ is the number of lagged forecast errors in the moving average component.

\subsubsection{Multilayer perceptron}
\begin{figure}
    \centering
    \includegraphics[width=3in,height=2in]{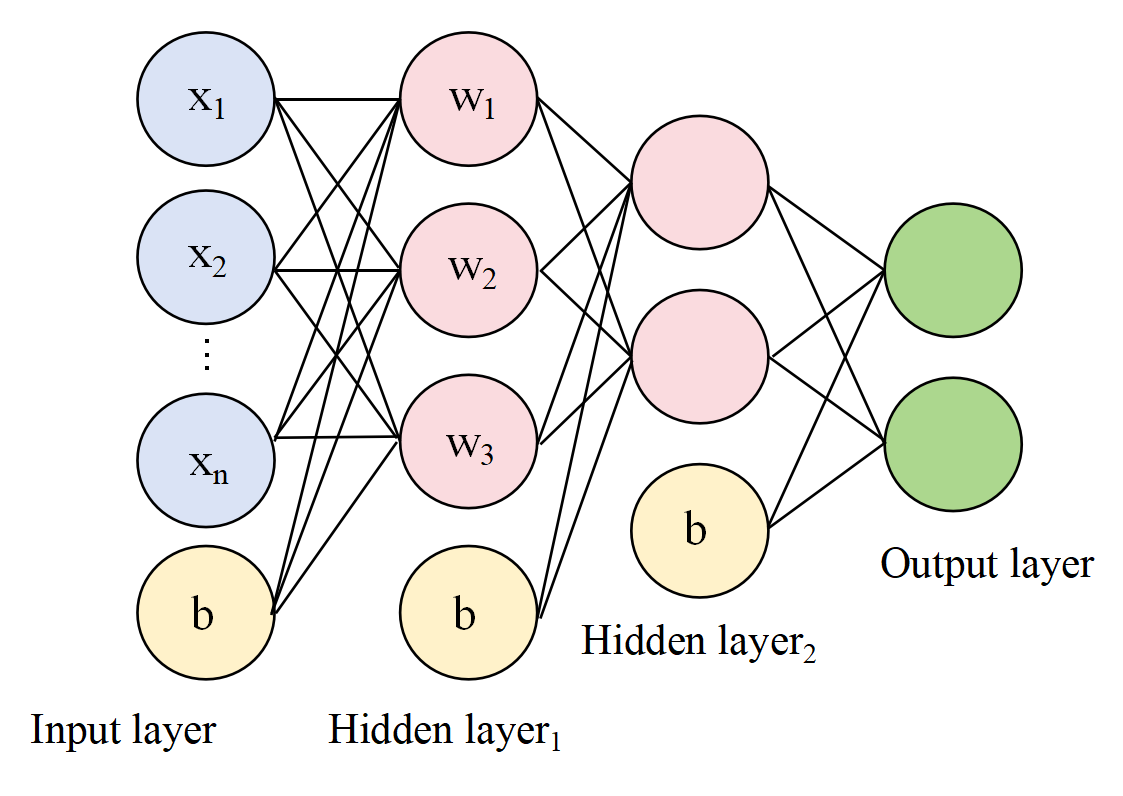}
    \caption{Architecture of a multilayer perceptron showing the input, hidden and output layers and interconnections between them, also the parameters in each layer(weight, bias and input). }
    \label{fmlp}
\end{figure}

A simple neural network, also known as the \textit{multilayer perceptron} is a machine learning model that features an input layer, an output layer and at least one hidden layer. Figure \ref{fmlp} illustrates the architecture of the MLP. The MLP need to use a training algorithm to update the weights and biases to ensure that the output (prediction) of the network resembles the actual observations (training data). The network computes the weight sum of inputs to get the hidden and output layers by
\begin{equation}
\begin{split}
    h_{W,b}(x) & = f(Wx+b) \\
    & = f(\sum_{i=1}^nW_ix_i + b) 
\end{split}
\end{equation}
where $x$ is the input item, $f(\cdot)$ is the activation function, $b$ is the bias, $n$ is the number of input units and $w$ is the weight.\\

\subsection{Deep learning models}
\subsubsection{Variants of LSTM networks}

\begin{figure}[h]
    \centering
    \includegraphics[width=3in,height=3in]{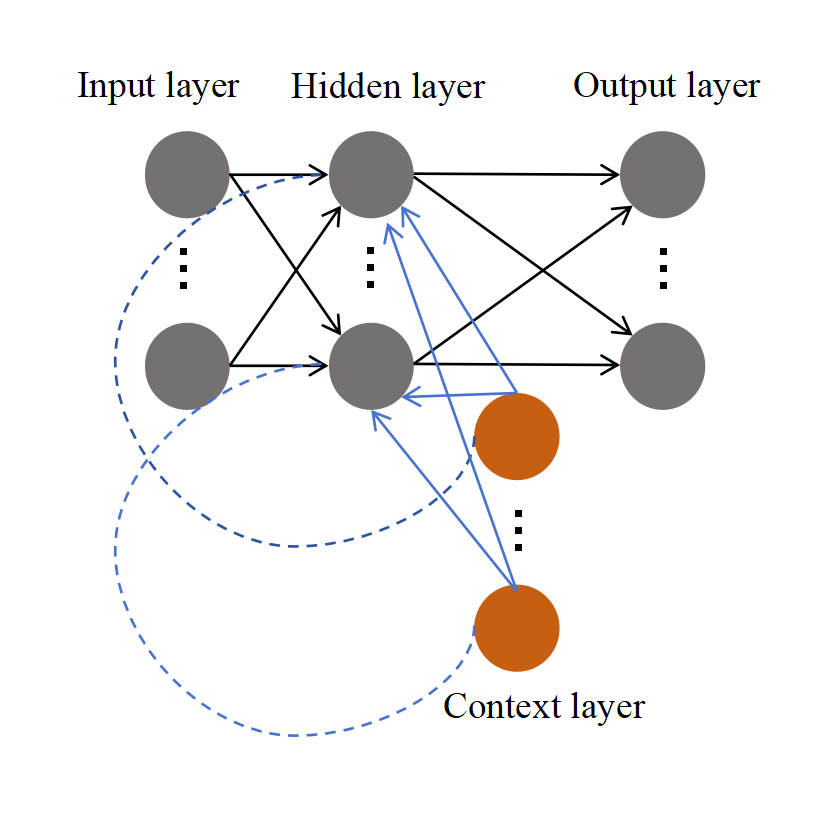}
    \caption{Architecture of Elman RNN that consists of input, hidden, context (state),  and output layers}
    \label{rnnsimple}
\end{figure}

RNNs are well-known for modelling temporal sequences, which are distinguished by their context layers as they memory information from prior input to influence the future results. There are several simple RNN architectures, such as the Elman RNN \cite{elman1990finding} (also known as simple RNN) which was one of the earliest attempts for effectively modelling temporal sequences. Figure \ref{rnnsimple} gives architecture of the Elman RNN. There are trainable weights connecting each two adjacent layers. A context (state or memory)  layer is used to store the output of state neurons resulting from the computation of previous time steps, making them appropriate for capturing time-varying patterns in data.
 
\begin{figure}[h]
    \centering
    \includegraphics[width=4in,height=2.3in]{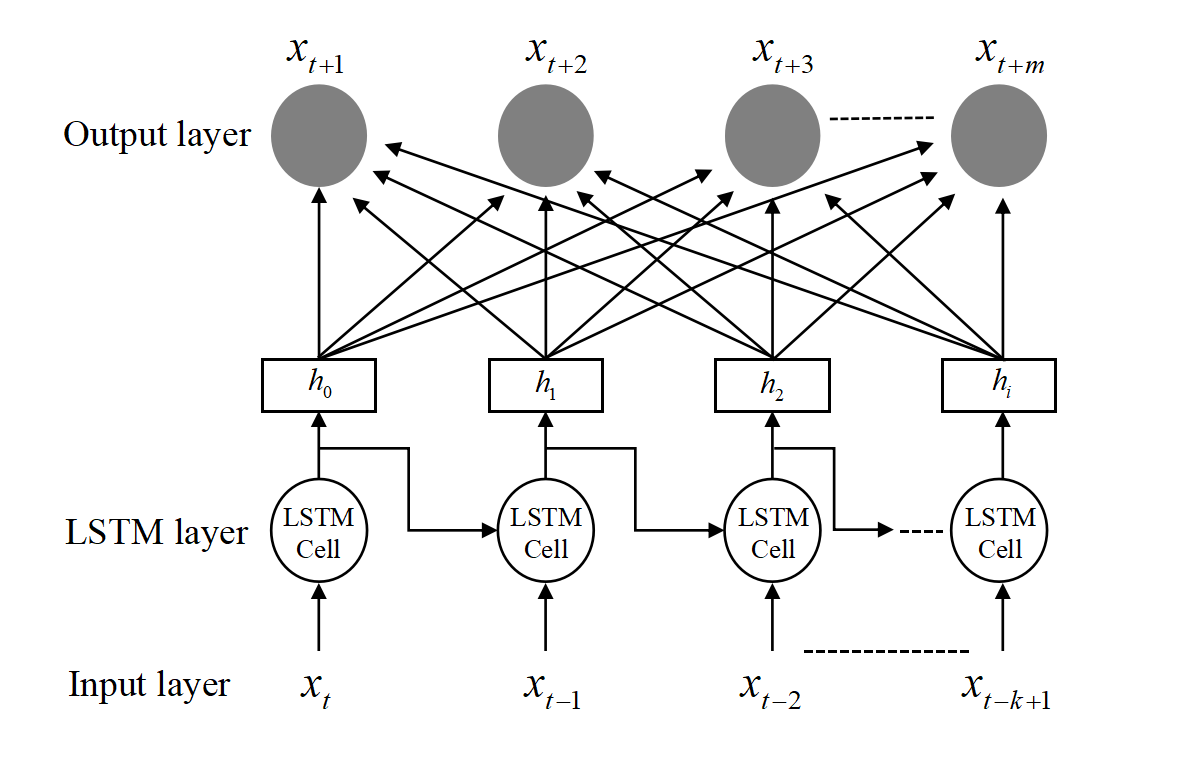}
    \caption{ LSTM network showing the input, hidden (LSTM cell), and output layers. LSTM cell extract information from input feature $x$ in over all time span.}
    \label{flstm}
\end{figure}

However, simple RNNs faced problems in training due to the vanishing gradient problem  \cite{hochreiter1998vanishing} arising when handling long-term dependencies in sequence data. The LSTM algorithm is considered to be an enhanced version of the RNN \cite{hochreiter1997long}. The LSTM overcame the vanishing gradient constraint by enhancing its ability to retain long-term dependencies through memory cells in the hidden layer. We present the architecture of LSTM network in Figure \ref{flstm} showing  how the information is passed through LSTM memory cells in the hidden layer. The LSTM cell is designed as a unit that memorises each input information for a long time, where previous information can still be retained, and hence addressing the problem of learning long-term dependencies in sequence data. The LSTM cell calculates a hidden state output $h_t$ by
\begin{equation}
\begin{split}
f_t &= \sigma(W_f[h_{t-1},x_t]+b_f)\\
    i_t &= \sigma(W_i[h_{t-1},x_t]+b_i)\\
    o_t &= \sigma(W_o[h_{t-1},x_t]+b_o)\\
    z &= tanh(W_z[h_{t-1},x_t]+b_z)\\
    C_t &= f*C_{t-1}+i_t*z\\
    h_t &= 0_t*tanh(C_t)\\
\end{split}
\label{eq1}
\end{equation}
where $f_t$ ,$i_t$ and $o_t$ refer to the forget gate, input gate and output gate respectively. $W$ is weight matrices adjusted learning along with $b$, which is the bias. $x_t$ is the number of input features, and $h_t$ is the number of hidden units. $z$ express as intermediate cell state, and $C_t$ is the current cell memory. 

\begin{figure}[htbp!]
    \centering
    \includegraphics[width=4in,height=3.5in]{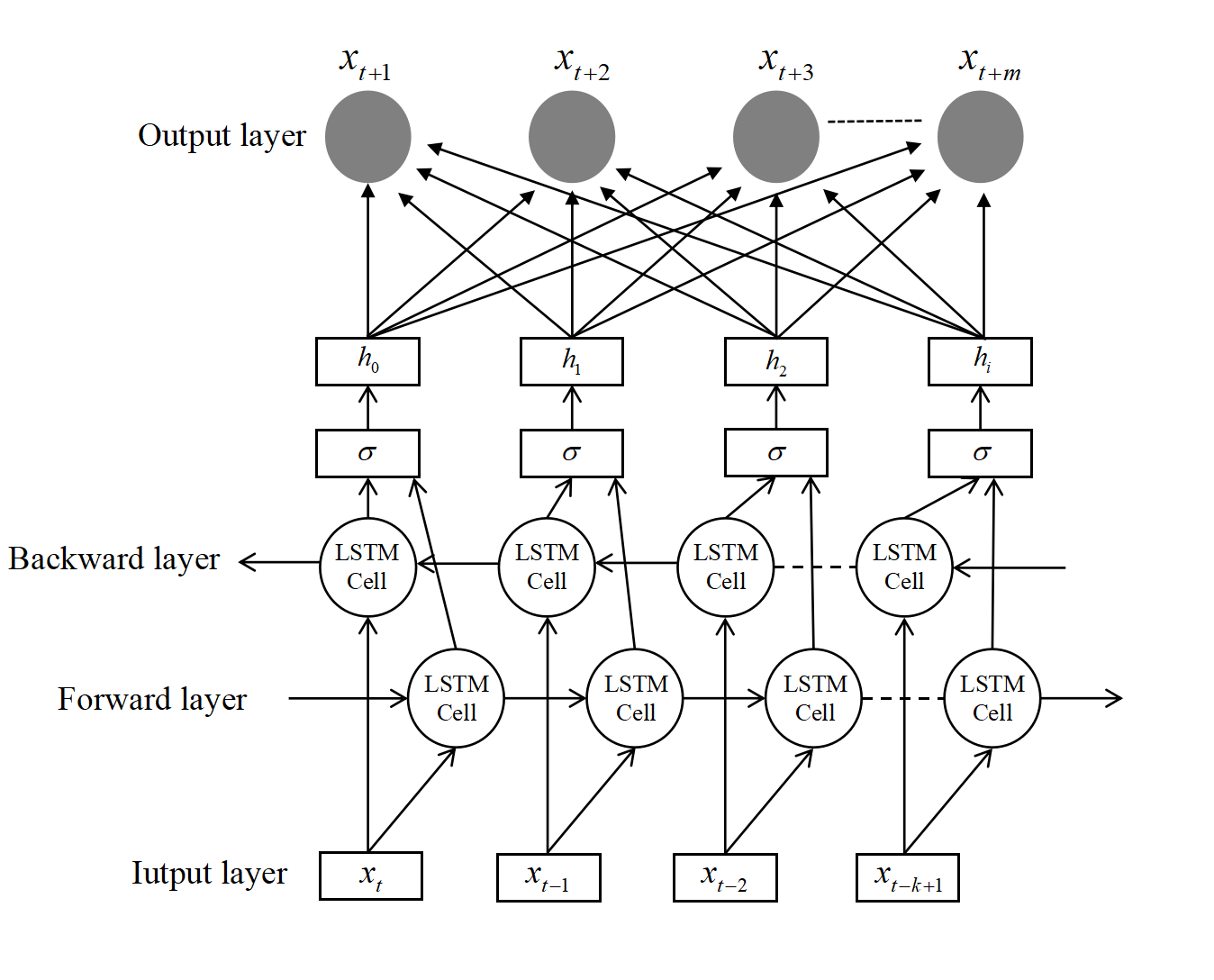}
    \caption{BD-LSTM Network showing the input, output and backward\&forward layer with the connection between them. These two hidden layer combine forward and backward information flow to enhance the ability of model to acquire information.}
    \label{fbdlstm}
\end{figure}

The bi-directional LSTM (BD-LSTM) is an advanced algorithm based on LSTM, which process information in both directions with two independent hidden layers \cite{graves2005framewise}. The basic idea is each input sequence passes through the RNN once in both the forward and reverse directions. This bidirectional architecture provides the output layer with complete past and future context information for each node in the input sequence. The structure of BD-LSTM is shown in Figure \ref{fbdlstm}. In contrast to LSTM, BD-LSTM exhibits greater efficacy in addressing problems that require the acquisition of context from both temporal directions. This is particularly evident in certain applications within the domains of natural language processing \cite{liu2016learning} and speech recognition \cite{chen2018end}.

\begin{figure*}[h]
    \centering
    \includegraphics[width=6.5in,height=2.5in]{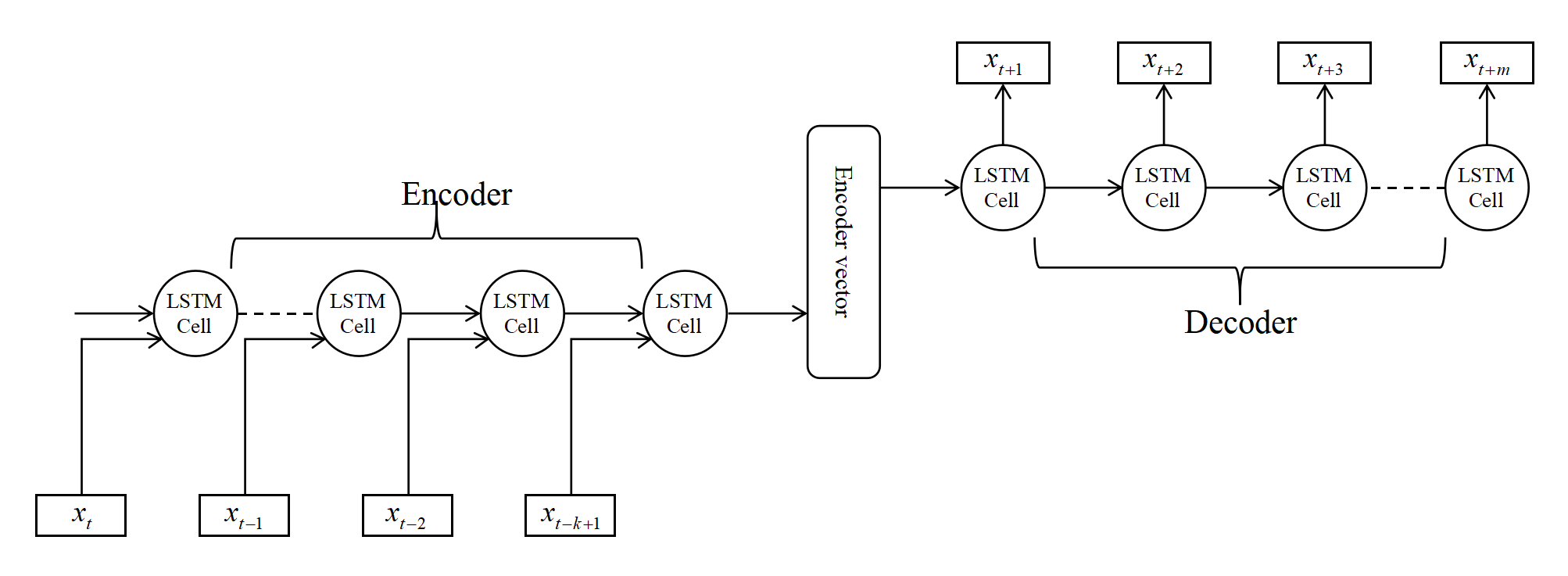}
    \caption{Encoder-Decoder LSTM Network showing encoder and decoder transfer sequence with encoder vector. }
    \label{fedlstm}
\end{figure*}

The encoder-decoder LSTM (ED-LSTM) can output a required sequence based on an input sequence (the length of the sequence can be different) \cite{sutskever2014sequence}. ED-LSTM makes specific architectural changes to the original LSTM to better handle a series of problems known as \textit{sequence to sequence}. ED-LSTM is very suitable for translating a certain language into different languages \cite{cho2014learning}. We present the ED-LSTM architecture in Figure  \ref{fedlstm}.

\subsubsection{Convolutional neural networks }

CNNs are one of the most prominent deep learning models initially designed for computer vision and image processing tasks \cite{gunduz2017intraday,di2016artificial,hoseinzade2019cnnpred,siripurapu2014convolutional}. Their application spans diverse areas, notably in detection \cite{jiang2017face} and segmentation tasks \cite{garcia2017review}, where they have shown superior efficacy accuracy when compared to traditional machine learning models. A CNN typically comprises several layers, including convolutional, pooling, and fully connected layers.

Subsequently, the fully connected layer, akin to conventional neural networks, ensures a dense interconnection between the nodes of consecutive layers. CNNs identifies hierarchical patterns (features) in the data through iterative convolution and pooling, culminating in a fully connected layer that consolidates these features for the final task output. This structural design has been pivotal for their proficiency in handling tasks related to image processing.


Given that our dataset consists of univariate time series for stock price, it's crucial to modify conventional two-diremntional CNN to suit our problem. Consequently, we've integrated a specialized function into our model that processes a set of stock data inputs, along with specifications like filter count, filter width, and stride length, while the kernel's height remains irrelevant. This function initializes filter values using a Gaussian distribution and sets biases to zero. It generates several matrices as outputs, where their quantity corresponds to the filter count. These matrices are crucial as they contribute to feature extraction within the CNN model, ultimately serving as inputs for the subsequent pooling layer following the activation function's execution. We note that in the case of multivariate time series data, the conventional 2D-CNN can be utilised. 



\begin{figure}[h]
    \centering
    \includegraphics[width=3.5in]{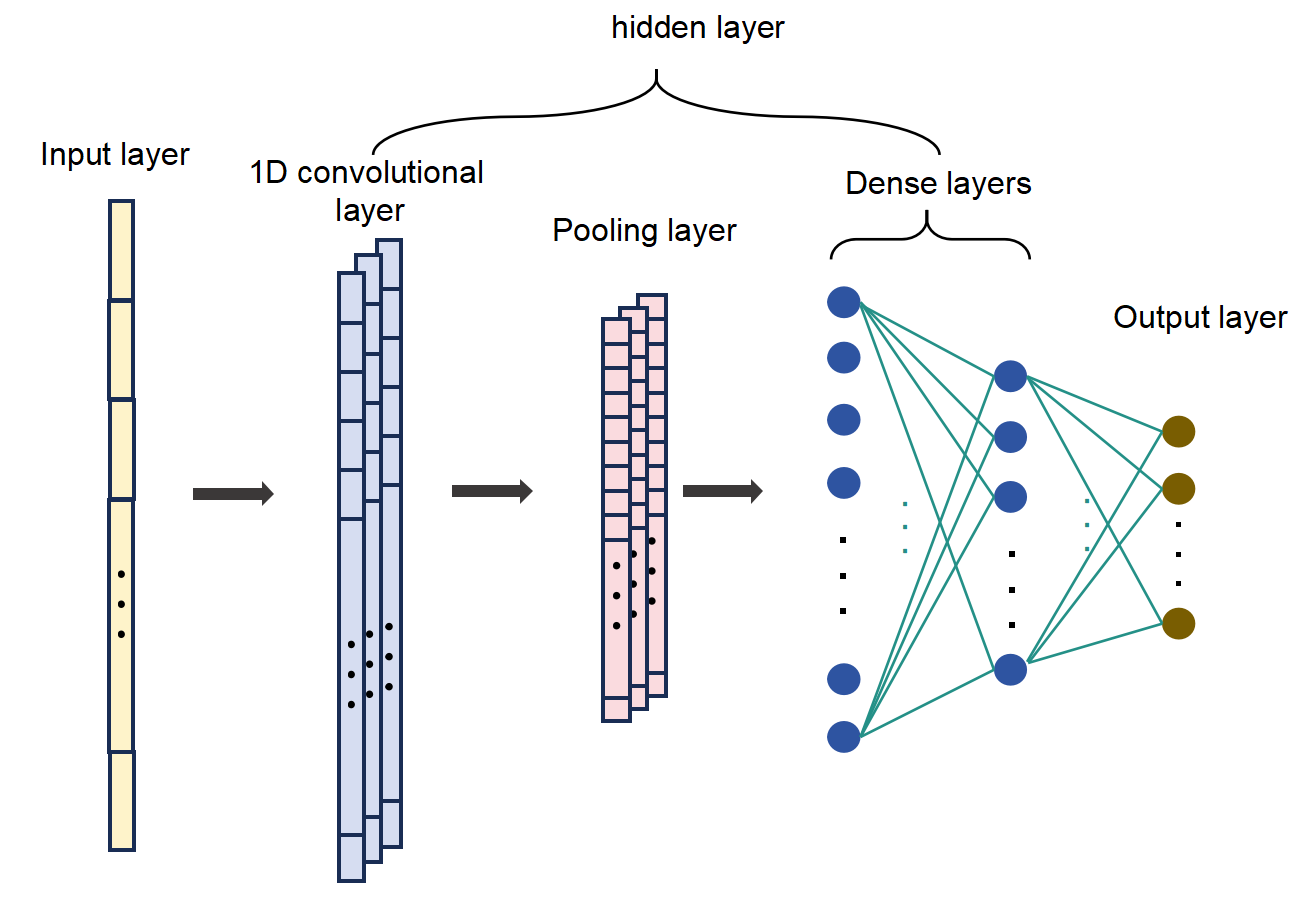}
    \caption{Architecture of 1D-CNN, featuring input, 1D-convolutional, pooling, dense and output layers.}
    \label{fcnn}
\end{figure}

The activation function  is essential to optimise models performance. The activation functions such as hyperbolic tangent (Tanh), rectifier linear units (ReLU), Sigmoid, and leaky ReLU are typically employed in CNNs. We opt for ReLU and Leaky ReLU which are prominent in the literature and also have ability to avoid vanishing gradients as given below. 

\begin{equation}
\begin{aligned}
\text{ReLU}(z) &= \begin{cases} 0 & \text{if } z \leq 0, \\ z & \text{if } z > 0, \end{cases} \\
\text{Leaky-ReLU}(z_i) &= \begin{cases} \alpha_i z_i & \text{if } z \leq 0, \\ z_i & \text{if } z > 0, \end{cases}
\end{aligned}
\end{equation}

where $z$ and $z_i$ are convolution outcomes, and $\alpha_i$  is user-defined hyperparameter for convolutional layer $i$, typically  starting at 0.01. We select  ReLU  for the initial convolutional layer to address the vanishing gradient issue, followed by Leaky-ReLU in subsequent layers as shown in Figure \ref{fcnn}.
We train the CNN model by minimising the   error defined by the loss function using the Adam optimiser \cite{kingma2014adam} with user-defined learning rate $\lambda=0.0001$. 





\subsubsection{Convolutional LSTM networks}
Convolutional LSTM (Conv-LSTM) network \cite{shi2015convolutional} was initially introduced for weather forecasting problems. This network extends the original fully connected LSTM and changes the matrix multiplication of the LSTM cell to convolution. We use $*$ to present convolution operation. And $\circ$ recognised as the Hadamard product. The key equations in the Conv-LSTM cell are expressed as :
\begin{equation}
\begin{split}
    f_t &= \sigma(W_{xf}*x_t + W_{hf}*h_{t-1} + W_{cf}\circ c_{t-1} + b_f)\\
    i_t &= \sigma(W_{xi}*x_t + W_{hi}*h_{t-1} + W_{ci}\circ c_{t-1} + b_i)\\
    o_t &= \sigma(W_{xo}*x_t + W_{ho}*h_{t-1} + W_{co}\circ c_t + b_o)\\
    c_t &= f_t \circ c_{t-1} + i_t \circ \tanh(W_{xc}*x_t+W_{hc}*h_{t-1}+b_c)\\
    h_t &= 0_t \circ \tanh(C_t)\\
\end{split}
\label{eq2}
\end{equation}
where $f_t$ ,$i_t$, $o_t$ and $h_t$ refer to the forget gate, input gate, output gate and hidden state respectively. $W$ is weight matrices adjusted learning along with $b$, which is the bias. Also, the past status $c_{t-1}$ can be regarded as “forgotten” in the process, and $c_t$ is the current cell memory. These equations are similar to \ref{eq1}. The Conv-LSTM model has the ability to capture both the spatial and temporal relationships in the data at the same time, resulting in more precise predictions. In our implementation, for the case of univariate time series, we utilise the 1D-convolutions in Conv-LSTM and 2D convolutional for multivariate time series forecasting.




\subsubsection{Transformer Networks}

\begin{figure*}[h]
    \centering
    \includegraphics[width=5.0in]{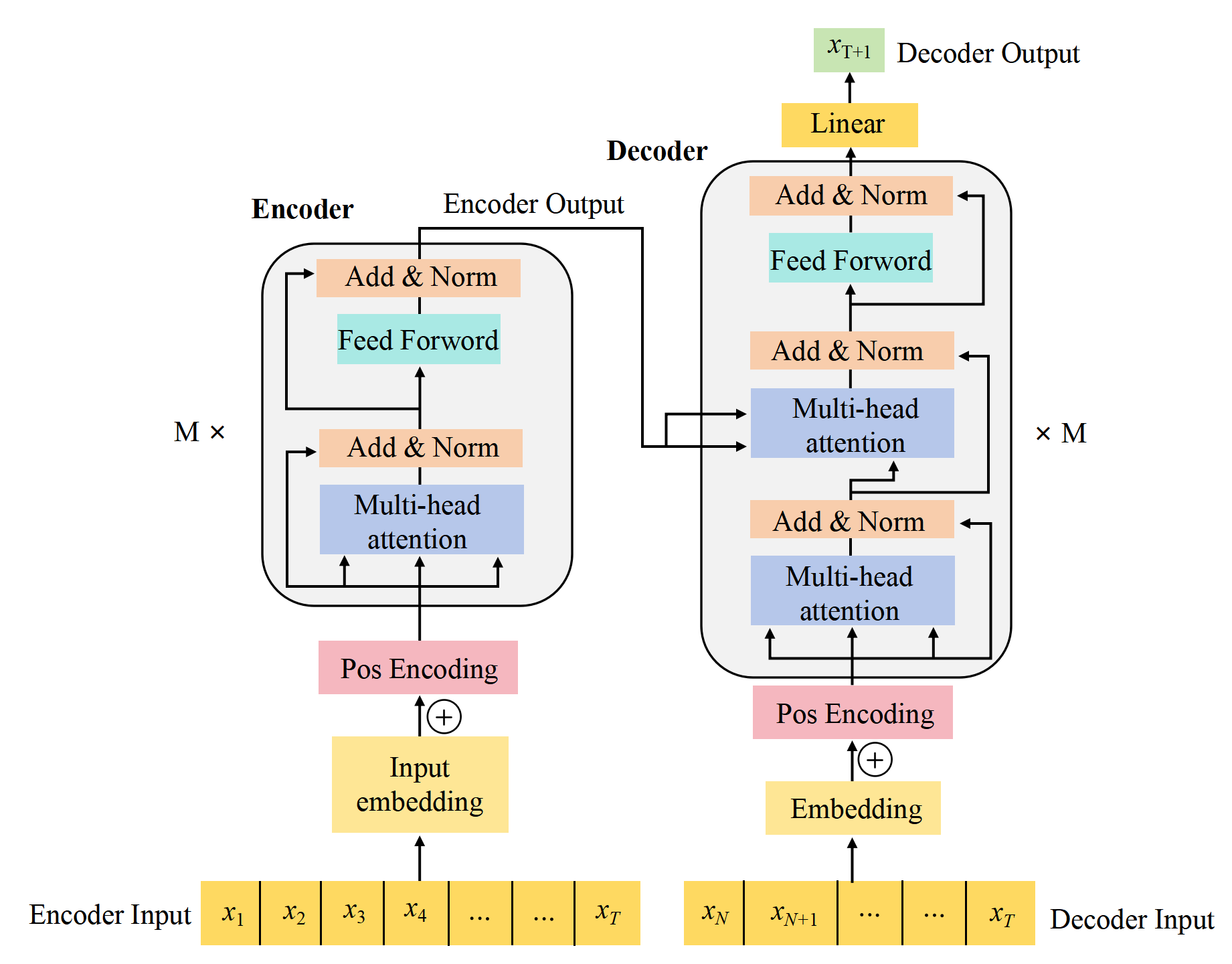}
    \caption{Architecture of Transformer model showing ...   }
    \label{ftransformer}
\end{figure*}
The Transformer model is an extension of the encoder-decoder LSTM architecture which has been widely used in machine translation problems \cite{cho2014properties}. 
The encoder condenses the essential data of the input sequence into a vector of fixed length, which is subsequently transformed into an output by the decoder \cite{sutskever2014sequence}. The design of the decoder offers a method for managing lengthy sequential data \cite{bahdanau2014neural}.

Analogously, we input the sequential data to a vector representation layer. Given the input sequence $X = \{x_i : i = 1, \ldots, N\} \in \mathbb{R}^N$, the $m$-dimensional embedding layer yields a matrix $B \in \mathbb{R}^{N \times m}$ through a dense network. 

We need to incorporate temporal encoding with the vectorised input to encapsulate the temporal structure of the time series. Hence,  employing sine and cosine functions at distinct frequencies to represent temporal information, we define:
\begin{align*}
\text{TE}_{(i, 2k)} &= \sin\left(i / 10000^{2k / m}\right), \\
\text{TE}_{(i, 2k+1)} &= \cos\left(i / 10000^{2k / m}\right),
\end{align*}
where $1 \leq 2k \leq m$. The temporal encoding, hence, is TE $\in \mathbb{R}^{N \times m}$. The vector representations alongside the temporal encodings are then concatenated and provided to the encoder layers.


A concise overview of the complete framework of our Transformer model is delineated in Figure \ref{ftransformer}. The encoder depicted in Figure \ref{ftransformer} consists of $M$ identically structured layers. Each layer is equipped with two sub-layers: a multihead self-attention mechanism and a fully connected feed-forward network. Both sub-layers incorporate residual connections and normalization to enhance their functionality. The decoder, also shown in Figure \ref{ftransformer}, mirrors the encoder's structure with a notable distinction: it features an additional multi-head self-attention layer. Unlike the original decoder described in \cite{vaswani2017attention}, this version omits the masked attention mechanism because it processes only observed historical data, which does not include future information.

The emergence of attention mechanisms marks a pivotal innovation in deep learning, focusing computational efforts to capture attention mechanism in cognition. Vaswani et al. \cite{vaswani2017attention} revolutionized this approach by introducing the Transformer architecture, predicated on the exclusive use of self-attention mechanisms. The self-attention mechanism, as defined, follows:

\begin{equation}
\text{Attention}(P, R, S) = \text{softmax}\left(\frac{P R^\top}{\sqrt{m}}\right) S,
\end{equation}

where $P, R, S \in \mathbb{R}^{N \times m}$ correspond to the query, key, and value matrices derived from three separate linear transformations of the same input. The architecture of the self-attention mechanism is illustrated in \ref{ftransformer}.

 The self-attention mechanism has transformed the strategy of focusing on vital local content within the data. Vaswani et al. \cite{vaswani2017attention} expanded this idea by proposing multi-head attention, whereby several self-attention processes, or "heads," are executed in parallel, each assessing different projected versions of the queries, keys, and values. The combined outcomes of these heads are then linearly transformed to obtain the final output as shown in Figure \ref{ftransformer}. 



\subsubsection{Model training with Adam Optimiser}
We utilise the modified Adam (adaptive moment estimation) optimiser \cite{kingma2014adam} which is  an extension to the \textit{stochastic gradient descent} \cite{amari1993backpropagation,ruder2016overview} and further extends adaptive gradient methods (AdaGrad \cite{duchi2011adaptive}, AdaDelta \cite{zeiler2012adadelta}, and RMSProp \cite{hinton2012neural}). Adam is an adaptive gradient-based optimisation algorithm that computes individual adaptive learning rates for different parameters from the history of the first and second moments (mean and variance) of the gradients. Let $g_k \circ g_k$ signify the element-wise square of $g_k$. 

\textbf{Input:}
\begin{itemize}
  \item Step size, $\alpha$
  \item Exponential decay rates for the moment estimates, $\beta_a, \beta_b \in [0,1)$
  \item Stochastic objective function with parameters, $f(\xi)$
  \item Initial parameter vector, $\xi_0$
\end{itemize}

Initialise:
\begin{align*}
m_0 & \leftarrow 0 \quad \text{(Initialize 1st moment vector)} \\
v_0 & \leftarrow 0 \quad \text{(Initialize 2nd moment vector)} \\
k & \leftarrow 0 \quad \text{(Initialize timestep)}
\end{align*}

\textbf{Algorithm:}
\begin{align*}
&\text{while } \xi_k \text{ not converged do} \\
&\quad k \leftarrow k + 1 \\
&\quad g_k \leftarrow \nabla_\xi f(\xi_{k-1}) \\
&\quad m_k \leftarrow \beta_a \cdot m_{k-1} + (1 - \beta_a) \cdot g_k \\
&\quad v_k \leftarrow \beta_b \cdot v_{k-1} + (1 - \beta_b) \cdot g_k \circ g_k \\
&\quad \hat{m}_k \leftarrow \frac{m_k}{(1 - \beta_a^k)} \\
&\quad \hat{v}_k \leftarrow \frac{v_k}{(1 - \beta_b^k)} \\
&\quad \xi_k \leftarrow \xi_{k-1} - \alpha \cdot \frac{\hat{m}_k}{(\sqrt{\hat{v}_k} + \epsilon')} \\
&\text{end while}
\end{align*}
\textbf{Output:} Resulting parameters $\xi_k$

Adam optimisation aims to minimise the expected value of a differentiable function, such as a neural network model $f(\xi)$ with a set of parameters given $\xi$ representing the weights and biases. The algorithm updates the exponential moving averages of the gradient $(m_k)$ and its square $(v_k)$, with hyperparameters $\beta_a, \beta_b$ controlling their decay rates. Adjustments in the algorithm improve efficiency, employing an updated computation for parameter adjustments with $\alpha_k = \alpha \sqrt{1 - \beta_b^k} / (1 - \beta_a^k)$. We note that vector operations are performed element-wise.

The key to Adam's update mechanism is the adaptive step size, influenced by the signal-to-noise ratio $\widehat{m}_k / \sqrt{\widehat{v}_k}$, dictating the magnitude of parameter updates. This feature allows for effective scaling of steps in parameter space, contributing to the robustness and versatility of the algorithm in various optimisation contexts.
 
\subsection{Data}

We choose four different cryptocurrencies to evaluate the performance of the respective statistical and deep learning models. The cryptocurrencies include Bitcoin, Ethereum, Dogecoin and Litecoin. We focus on multi-step ahead stock price forecasting, where a step is defined by a day. Bitcoin is the first and most prominent cryptocurrency, which was launched in 2009 by   Satoshi Nakamoto \cite{nakamoto2008bitcoin}.   Ethereum was designed in 2013 by Vitalik Buterin and Gavin Wood \cite{buterin2014next}. Ethereum is not just a cryptocurrency but also a platform for building decentralized applications using smart contracts. After Bitcoin, Ethereum is the cryptocurrency with the second-largest market capitalisation. Dogecoin, another open-source cryptocurrency based on the popular "doge" internet meme, grew in popularity and price in 2021 after billionaire Elon Musk publicly backed it. Litecoin created by Charlie Lee in 2011, Litecoin is based on Bitcoin's protocol but differs in terms of the algorithm used. Litecoin uses the scrypt encryption, proposed by Colin Percival \cite{percival2016scrypt}.   

Due to the incompleteness of data sources, we combine data sources from two websites, including Yahoo Finance\footnote{\url{https://finance.yahoo.com/lookup}} and Kaggle \cite{web:kaggle:cryptoprice} with fundamental details  summarised in Table \ref{t411}. The datasets feature the historical price information of the four cryptocurrencies with begin and end date and the  number of data points shown in Table \ref{t412}. We  forecast the closing price of each cryptocurrency using  univariate and multivariate deep learning models. According to Wang et al. \cite{wang2023machine}, in the multivariate model, it is feasible to incorporate the features of the cryptocurrency price such as the $open$, $high$, $low$ and $volume$ to enhance forecasting accuracy. Hence, we used these features in our multivariate models as shown in Table \ref{t412}.  We also add gold prices as an additional feature to the multivariate model, as noted by Huynh et al. \cite{huynh2020small} who found a  strong correlation between cryptocurrencies and the gold market. We obtained the Gold price data from London bullion market (LBMA) during 31 December 2012 to 28 February 2022 collected from Factset\footnote{\url{https://www.factset.com/}}.

 \begin{table*}[h]
   \centering
    \small
 \begin{tabular}{c c c c c}
     \hline \hline
     
   Cryptocurrency &  Period (Day/Month/Year) &   Size &  Mean: Close price (USD) &  Variance: Close price  \\
   \hline
   Bitcoin & 29/04/2013-01/04/2024 & 3991 & 13692 & $2.856 \times 10^{8}$ \\
    Ethereum & 08/08/2015-01/04/2024 & 3160 & 983.53 & $1.274 \times 10^{6}$ \\ 
    Dogecoin & 16/12/2013-01/04/2024 & 3760 & 0.0406 & $5.947 \times 10^{-3}$ \\
    Litecoin & 29/04/2013-01/04/2024 & 3991 & 60.208 & $3.884 \times 10^{3}$ \\ 
   \hline \hline
    \end{tabular}
   \caption{Statistical description of the dataset, where the size refers to the number of data points (days).}
   \label{t411}
 \end{table*}

\begin{table}[h]
    \centering
    \small
    \begin{tabular}{l l l}
    \hline \hline
    Variable& Variable Description & Data type \\
    \hline
    SNo & the order of the data & Number\\
    Name & Name of cryptocurrency & Letter \\
    Symbol & Abbreviation of cryptocurrency & Letter \\
    Date & Date of observation & Date \\
    High & Highest price on  given day & Number \\
    Low & Lowest price on  given day & Number \\
    Open & Opening price on given day & Number \\
    Close & Closing price on  given day & Number \\
    Volume & Volume of transactions on given day & Number \\
    \hline \hline
    \end{tabular}
    \caption{Dataset for multivariate models. }
    \label{t412}
\end{table}

\subsection{Data processing }

\begin{figure*}[htbp]
    \centering
    \includegraphics[width=5.0in]{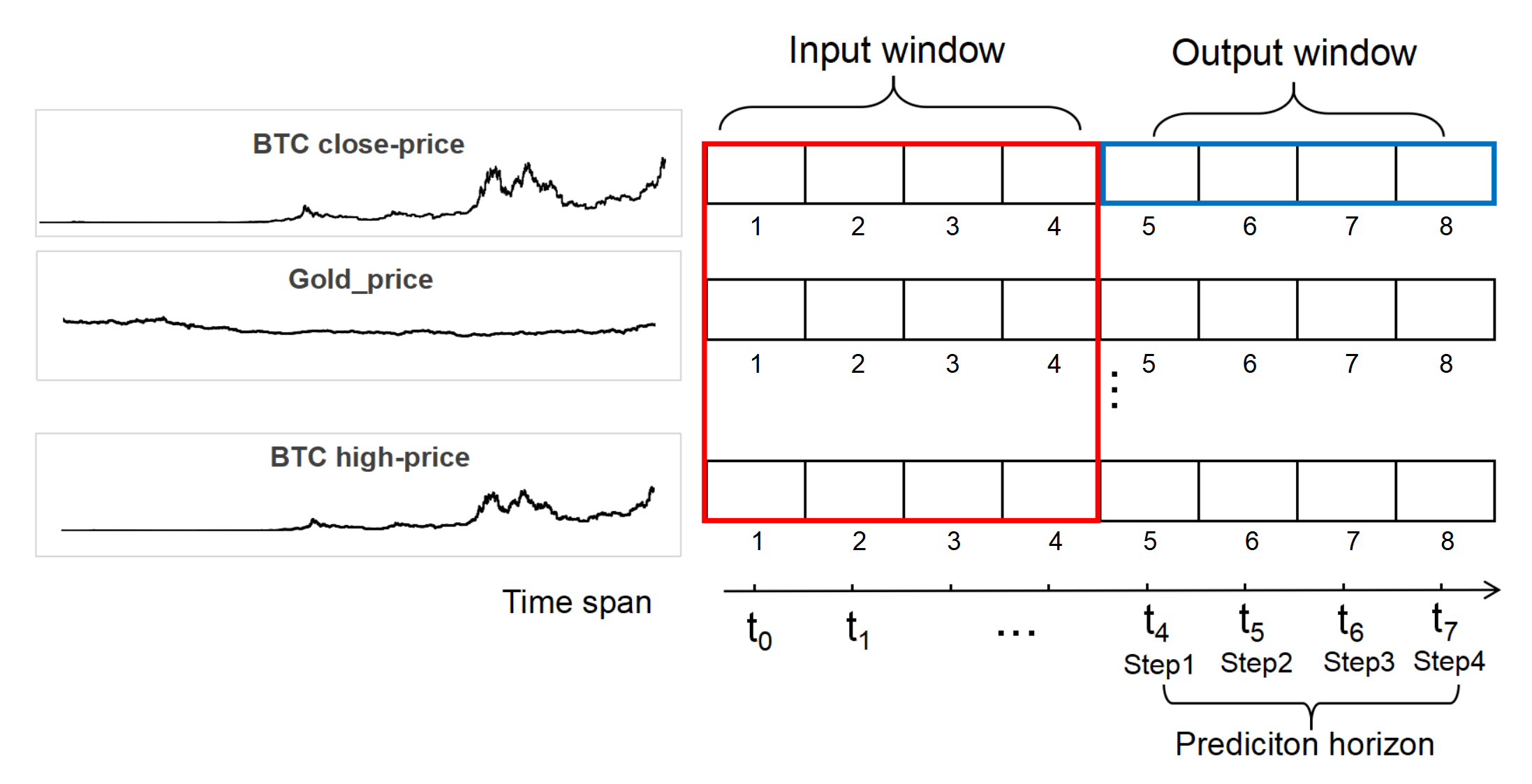}
\caption{Data windowing in multivariate approach with Bitcoin close-price forecasting, showing the size of input window (time series window) and output window (prediction horizon). The input data consists of multiple time series, such as the Bitcoin (BTC) high-price and close price  and the stock price of Gold, to provide multistep ahead prediction of BTC close price.}
    \label{fmulti}
\end{figure*}

 We need to reconstruct the original time series data for multi-step-ahead prediction using deep learning models. The embedding theorem of Taken's states that the reconstruction process can replicate significant characteristics of the initial time series \cite{takens2006detecting}. Given an observed time series $x(t)$, we can generate embedded phase space $Y(t)=[x(t),x(t-T),...,X(t-(D-1)T)]$ ; where $T$ is the time delay, $D$ is the embedding dimension with $t=0,1,2,...,N-D-1$ , and $N$ is the original length of the time series. Takens' theorem demonstrates that if the original attractor had a dimension of $d$, then an embedding dimension of $D=2d+1$ would be enough.
In univariate method prediction, we divide a single time series data set up into several sections. The input characteristics for each segment are data from $N$ subsequent time points, and the output label(s) is the time point(s) that comes afterwards. Therefore, we can have single-step prediction or multi-step ahead prediction.
In the multivariate strategy, as input vectors, we will utilise a window that holds multiple time series data at sequential multiple time points as shown for the case of Bitcoin price prediction in Figure \ref{fmulti}. The input data consists of multiple time series, such as the high-price and close-price of Bitcoin, and the stock price of Gold to provide a multi-step prediction of Bitcoin close price. 

\subsection{Framework} 
\begin{figure*}[htbp!]
    \centering
    \includegraphics[width=6in,height=4.5in]{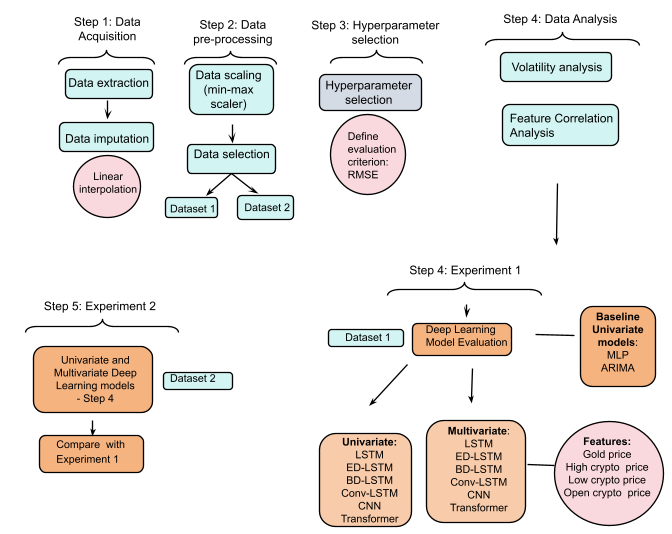}
    \caption{Framework of evaluation and details include data collection, pre-experiment analysis, data pre-processing, hyperparameter selection and design of experiments. }
    \label{fframework}
\end{figure*}

We present our framework in Figure \ref{fframework} that highlights major components of the entire process. In Step 1, we extract data and pre-process data for analysis. Since the Gold price data only has its value on trading days (Monday to Friday), we use interpolation methods to fill in prices on non-trading days (Saturday, Sunday and Public Holidays). The interpolation we used is a linear method\footnote{\url{https://pandas.pydata.org/docs/reference/api/pandas.DataFrame.interpolate.html\#pandas.DataFrame.interpolate}} which fills in missing values with the average of both sides. This is commonly used to handle missing values in time series data \cite{gnauck2004interpolation}. 

The data pre-processing in Step 2 features data-scaling to ensure the model stability, which limits the price within the range defined by \textit{min-max} scalar\footnote{\url{https://scikit-learn.org/stable/modules/generated/sklearn.preprocessing.MinMaxScaler.html}}. Furthermore, we separate the data into 2 sets for comparing the influence of COVID-19 and also determine the train-test data split, which is based on a given timeline (not shuffled).   Our goal is to predict the closing price of the respective cryptocurrencies, therefore we have univariate data with close price and multivariate data with close price, gold price, high price, low price and opening price as shown in Table 1.  We split the  dataset  into a 70:30 ratio.  We use the data for the selected cryptocurrency from the opening to June 2021 as Dataset 1, where the last 30\% of the data includes the period following the beginning of COVID-19 (March 2020 to June 2021), during which high volatility in crypto price was evident. We know that cryptocurrency is a financial asset with highly volatile prices \cite{bhosale2018volatility}, hence we explore which model is more effective in predicting the rising trend following the breakout of COVID-19. Dataset 2 features  the COVID-19 period both in the train and test dataset (March 2020 to April 2024) to ensure that high volatility crypto price data is part of the training dataset.  Table \ref{tab:data-descr} presents the dates for the train and test  datasets, for both experiments. 

\begin{table}[h]
    \centering
    \small
    \begin{tabular}{c c c c}
    \hline \hline
    Crypto & Experiment & Split& Period (Day/Month/Year) \\
    \hline
    Bitcoin & Dataset 1 & Train & 29/04/2013-27/12/2018 \\
      &  & Test & 28/12/2018-01/06/2021 \\
      & Dataset 2 & Train & 01/03/2020-09/01/2023 \\
  &   & Test & 10/01/2023-01/04/2024 \\
    \hline
    Ethereum & Dataset1 & Train& 08/08/2015-02/09/2019 \\
  &   & Test& 03/09/2019-01/06/2021 \\
   & Dataset 2 & training & 01/03/2020-09/01/2023 \\
   &   & testing & 10/01/2023-01/04/2024 \\
    \hline
    Dogecoin & Dataset 1 & Train & 16/12/2013-06/03/2019 \\
    &  & Test & 07/03/2019-01/06/2021 \\
     & Dataset 2 & Train & 01/03/2020-09/01/2023 \\
    &   & Test & 10/01/2023-01/04/2024 \\
    \hline
    Litecoin & Dataset 1 & Train & 29/04/2013-27/12/2018 \\
     & & Test & 28/12/2018-01/06/2021 \\
     & Dataset 2 & Train & 01/03/2020-09/01/2023 \\
 & & Test & 10/01/2023-01/04/2024 \\
    \hline \hline
    \end{tabular}
    \caption{Time span of training and test dataset in different experiments for each cryptocurrency. }
    \label{tab:data-descr}
\end{table}
In Step 3, we select the optimal hyperparameters for each model using trial runs, knowledge from the same model runs in literature (e.g. \cite{chandra2021evaluation}), and the default values in the library implementation (e.g. PyTorch\cite{paszke2019pytorch}). We note that we use RMSE as the accuracy measure for all the models in our framework. Once the best parameters have been determined, we can then continue with our investigations that compare univariate and multivariate deep learning models. 

In Step 4, we provide data analysis by first implementing a volatility analysis of the close price of selected cryptocurrencies, which can reveal fluctuations and patterns throughout the selected period. Historically, cryptocurrencies featured a wide price range, with high major fluctuation across the COVID-19 period \cite{mandaci2022herding,naeem2021asymmetric}. We use the volatility analysis to review the fluctuations since they can lead to instability during the model training process, causing slower convergence and poor generalisation ability on the test dataset. In Step 4, taking into account our multivariate models, we provide feature correlation analysis to find out how the different features affect each other.

We next compare the respective models in Step 5 with  Experiment 1  (pre-COVID-19 training data) and select the two best-performing models for the next step. We develop and compare the multivariate model and univariate deep learning model including LSTM, BD-LSTM, ED-LSTM, CNN, Conv-LSTM, and Transformer.  models  predict the close price of the selected cryptocurrencies.  We use MLP and ARIMA models as baseline models for Bitcoin dataset.

In Step 6, we use Dataset 2 for Experiment 2 to  predict the close price during  COVID-19 using training data that features COVID-19 effect on the cryptocurrencies. We  do this to determine whether the prediction accuracy has been improved and hence compare the results with Experiment 1.  We also incorporate the \textit{shuffle} data splitting strategy to enhance the efficacy of model performance. The initial 70\% of the data is randomly rearranged using the \textit{shuffle} strategy.



 \subsection{Technical details}

In order to distinguish the model performance, we use  RMSE  as the criterion for the different prediction horizons. The smaller the RMSE values, the better the prediction accuracy:

 \begin{equation}
     RMSE = \sqrt{\frac{1}{N}\sum_{i=1}^N(y_i-\hat{y_i})^2}
 \end{equation} 
 
 where $y_i$ and $\hat{y_i}$ are the observed data and predicted data, respectively. $N$ is the length of observed data.

As indicated earlier, we use the Adam optimiser for all the deep learning models, where we use default values for the hyperparameters, i.e. $\alpha = 0.001, \beta_a = 0.9, \beta_b = 0.999$, and $\epsilon' = 1e-8$.

In the case of the Transformer and ARIMA model, we reviewed the literature \cite{sridhar2021multi,tanwar2022prediction,roy2018bitcoin} to obtain the hyperparameters.   Table \ref{t423} describes the details of  model hyperparameters, including the number of input layers, output layers, hidden layers and other hyperparameters. We use the ReLu activation function in the respective deep learning models with a maximum training time of 200 epochs via the Adam optimiser \cite{kingma2014adam}.

We implemented specific experiments to determine the appropriate hyperparameters for each model. Based on related models in the literature \cite{chandra2021evaluation}, we used model architectures: CNN and LSTM variants feature one hidden layer with selected hidden units. We refer to previous research on MLP model for time series prediction \cite{wang2006forecasting} and evaluate performance  for selected hidden neurons, as shown in Table \ref{t421}. We use the first 70\% of the Bitcoin close-price in dataset1 for training and the remaining for testing. We repeated model training with different initial parameters for each hyperparameter configuration 5 times and reported the average. Table \ref{t421} presents the performance (RMSE) of each model in the test dataset for the hyperparameters, with the best values in bold.  

\begin{table}[htbp!]
\small
    \centering
    \begin{tabular}{c c c c}
    \hline \hline
    Model & Hidden & Train & Test \\
    \hline
    LSTM & 20 & 0.0194 & 0.0696 \\
    & 50 & 0.0176 & 0.0490 \\
    & 100 & 0.0165 & 0.0370 \\
    \hline
    BD-LSTM &20 &0.0195 & 0.0239 \\
    & 50 & 0.0177 & 0.0210 \\
    & 100 & 0.0170 & 0.0204 \\
    \hline
    ED-LSTM &20 &0.0192 & 0.0930 \\
    & 50 & 0.0169 & 0.0609 \\
    & 100 & 0.0164 & 0.0373 \\
    \hline
    Conv-LSTM &20 &0.0124 & 0.0176 \\
    & 50 & 0.0123 & 0.0181 \\
    & 100 & 0.0132 & 0.0194 \\
    \hline
    CNN &20 &0.0126 & 0.0206 \\
    & 50 & 0.0128 & 0.0209 \\
    & 100 & 0.0135 & 0.0235 \\
    \hline
    MLP &5 &0.0137 & 0.0278 \\
    & 10 & 0.0130 & 0.0268 \\
    & 20 & 0.0122 & 0.0195 \\
    \hline \hline
    \end{tabular}
    \caption{Hyperparameter selection for the number of hidden units for the selected models using RMSE for the train and test dataset (LSTM, BD-LSTM, ED-LSTM, Conv-LSTM, CNN, and MLP).}
    \label{t421}
\end{table}

 \begin{table*}[h]
   \centering
   \footnotesize
  \begin{tabular}{l c c c c}
   \hline \hline
        & \makecell[c]{Input\\layers} & \makecell[c]{Hidden\\Layers} & \makecell[c]{Output\\layers} & Comments \\
    \hline
    MLP & (6,1) & 3 & (1,5) & \makecell[l]{Include three hidden layers.} \\
    ARIMA & - & - & - & \makecell[l]{Construct ARIMA(1,0,1) model.} \\
    LSTM & (6,1) & 2 & (1,5) & \makecell[l]{Include two LSTM layers.}\\
    BD-LSTM & (6,1) & 2 & (1,5) & \makecell[l]{Include Forward\&Backward LSTM layer.} \\
    ED-LSTM & (6,1) & 4 & (1,5) & \makecell[l]{Two LSTM networks with a time distributed layer.}\\
    Conv-LSTM & (6,1) & 3 & (1,5) & \makecell[l]{Include Conv1D layer, LSTM network and dense layer.} \\
    CNN & (6,1) & 4 & (1,5) & \makecell[l]{Include Conv1D layer, pooling layer and two dense layers.} \\
    Transformer & (6,1) & 2 & (1,5) & \makecell[l]{Include a Multi-Head Attention mechanism and a \\position-wise fully connected feed-forward network.} \\
     \hline \hline
    \end{tabular}
    \caption{Hyperparameter configuration of all respective models.}
    \label{t423}
 \end{table*}

\section{Results} \label{ch4}
In this section, we provide comprehensive information about the datasets and present research design with computational results.

\subsection{Data analysis}

The \textit{coronavirus disease 2019} (COVID-19)  pandemic \cite{cucinotta2020declares}  originated 17th November 2019 in Wuhan, China, and extensively began spreading from March 2020 \cite{andersen2020proximal} worldwide. COVID-19 had a devastating effect on the world economy, and its impact included   finance, supply chain, politics, and mental health \cite{brodeur2021literature}, with further effects in the post-pandemic era \cite{leach2021post,miao2023immunogen,laskawiec2022post}. Therefore, it is necessary to analyze the price trends of the four cryptocurrencies in our study. 

\begin{figure*}[h]
\subfloat[Bitcoin]{
    \label{btc-price}
    \includegraphics[width=9cm]{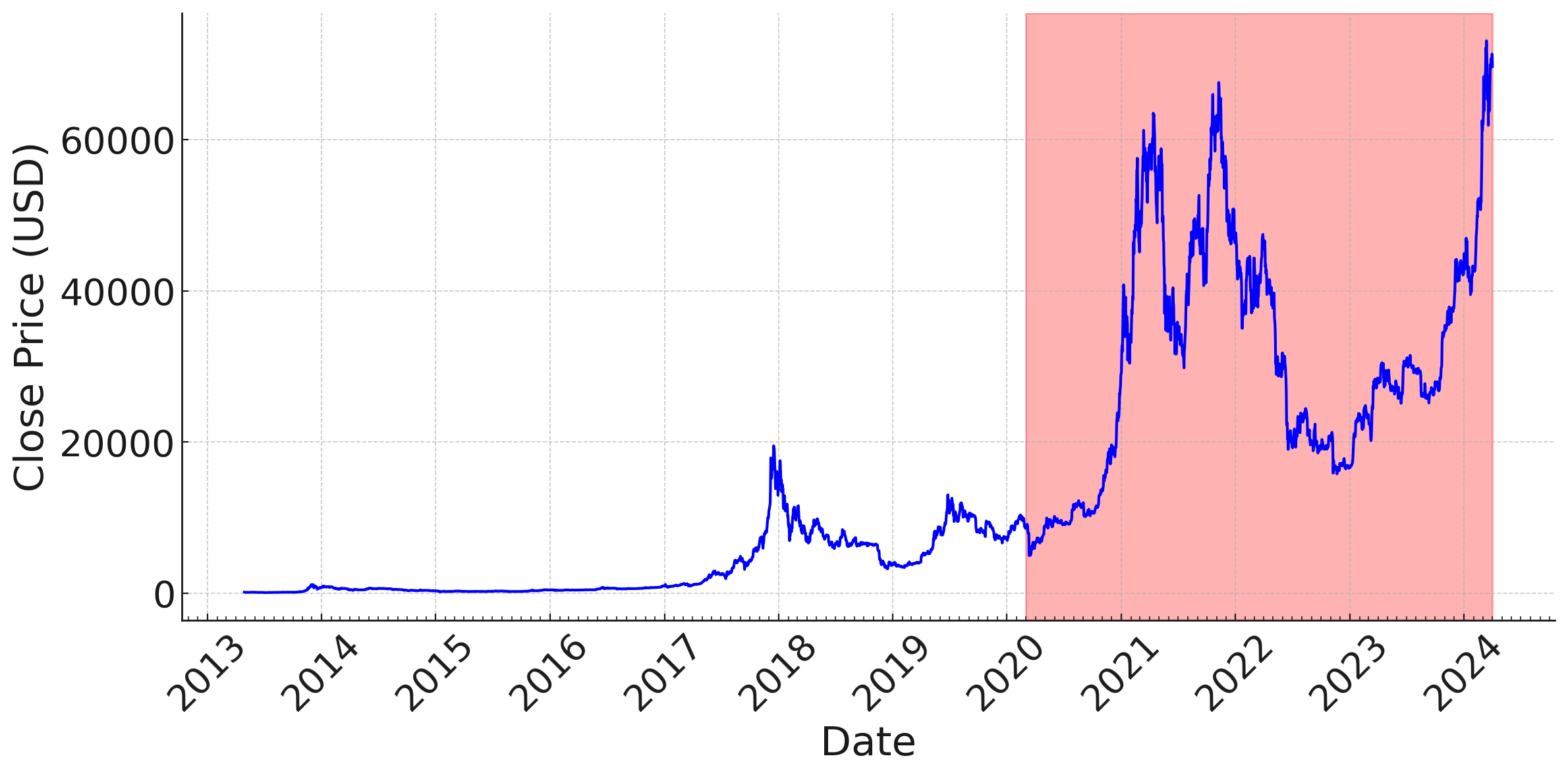}
}
\subfloat[Ethereum]{
    \label{eth-price}
    \includegraphics[width=9cm]{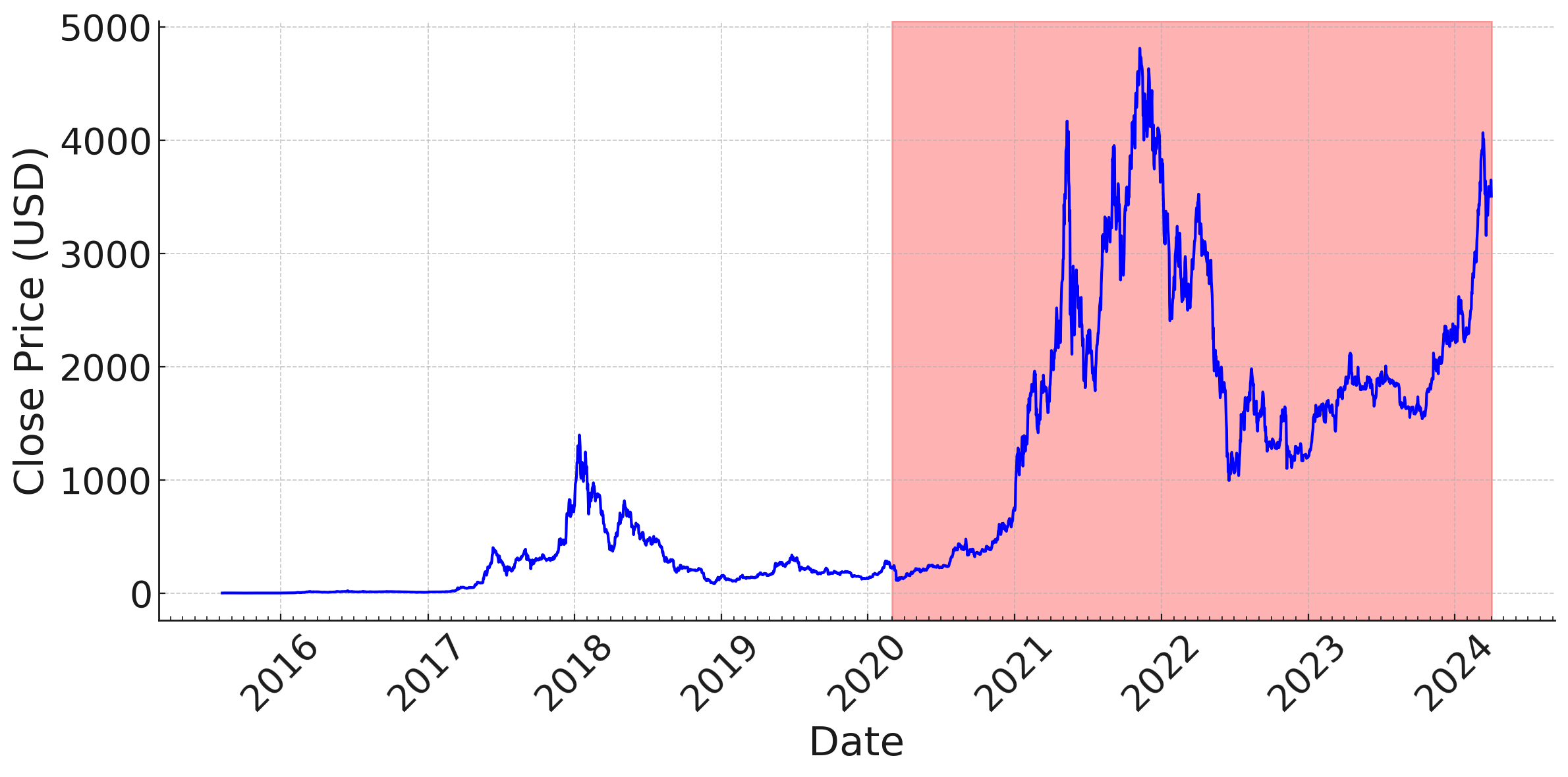}
}
\\
\subfloat[Dogecoin]{
    \label{doge-price}
    \includegraphics[width=9cm]{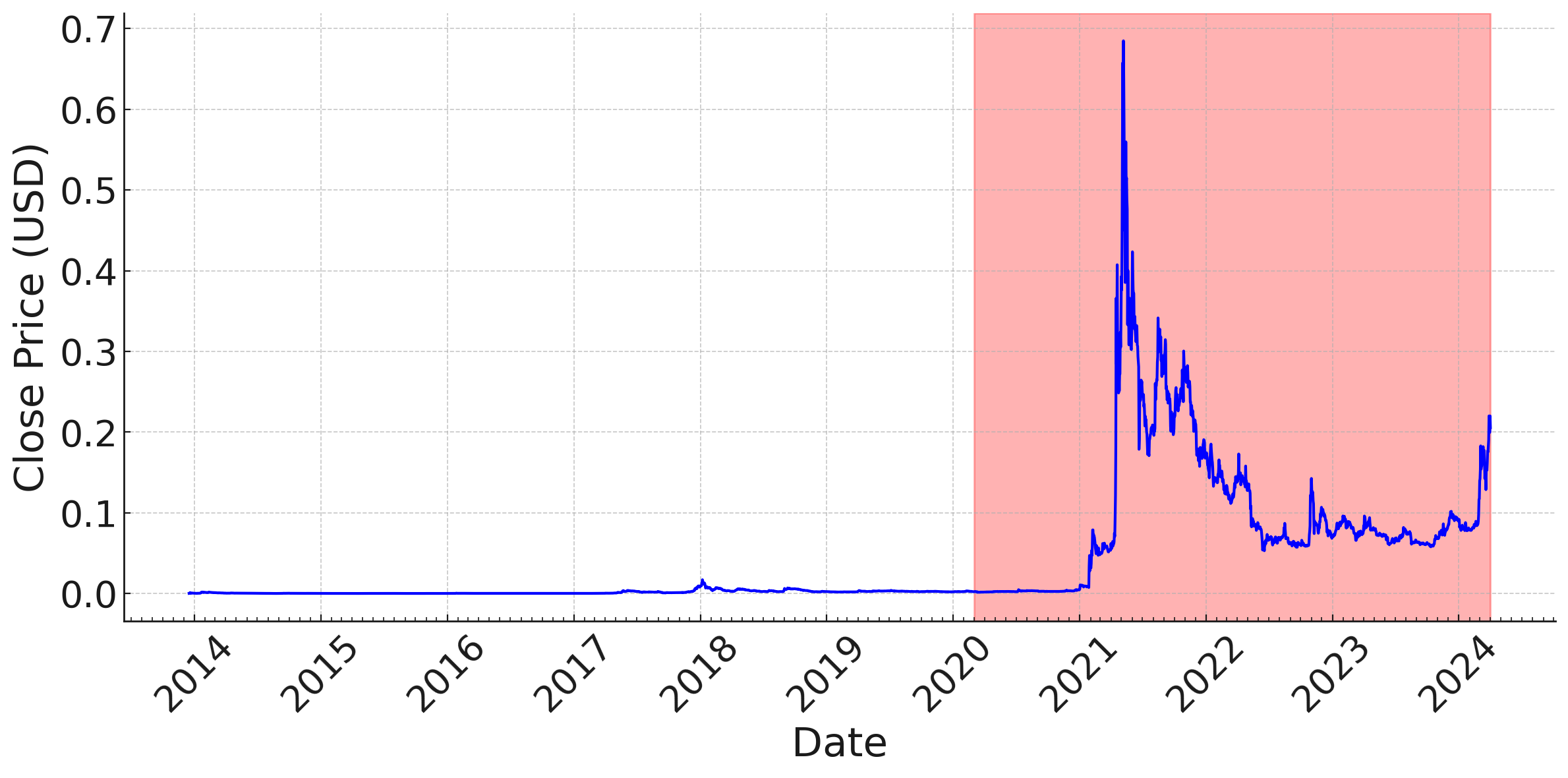}
}
\subfloat[Litecoin]{
    \label{ltc-price}
    \includegraphics[width=9cm]{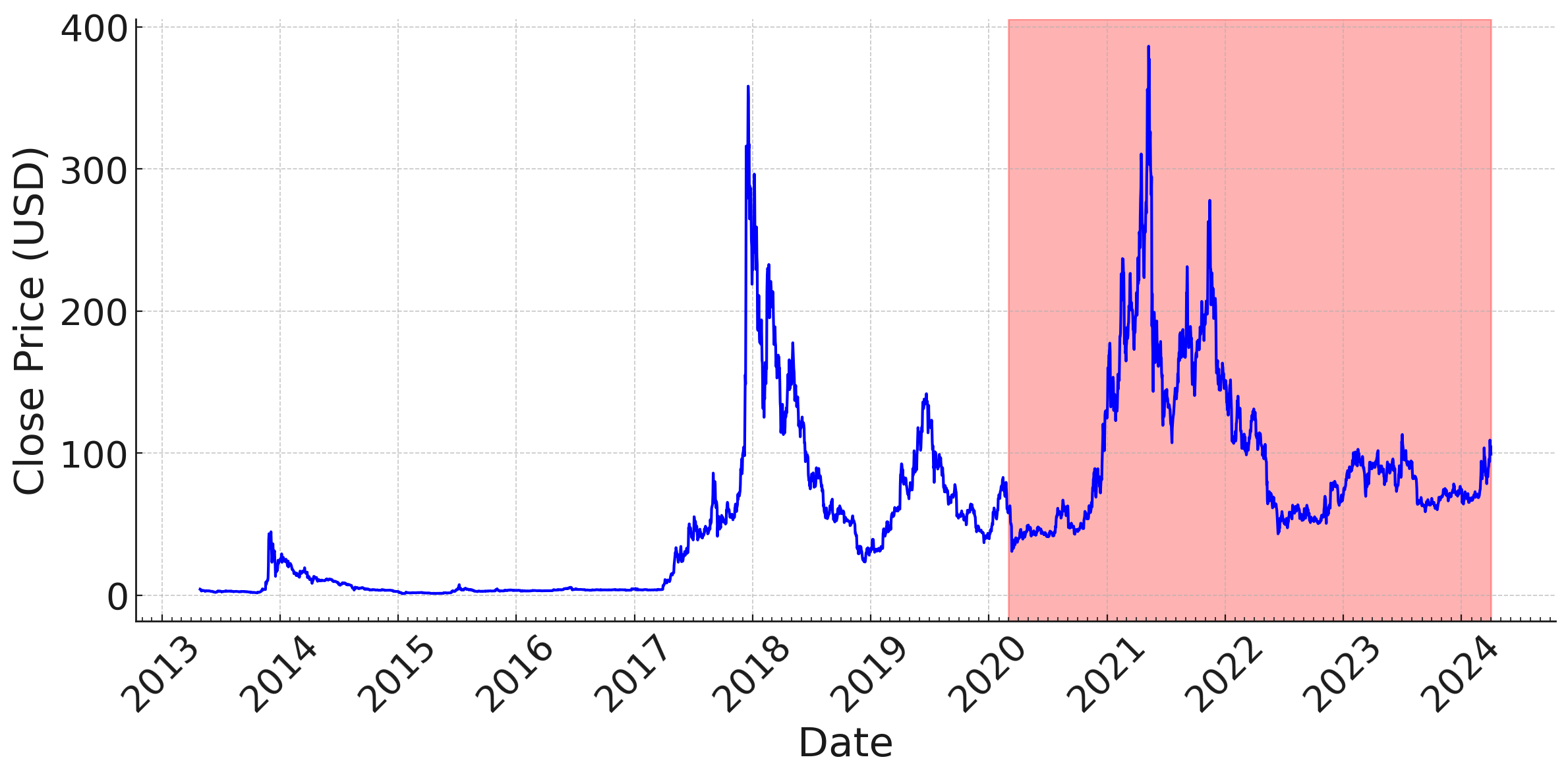}
}
\caption{Time-series of cryptocurrency close-price highlighting COVID-19 trend in pink.}
\label{fclose}
\end{figure*}

\begin{figure*}[h]
\subfloat[Bitcoin]{
    \label{btc-price}
    \includegraphics[width=9cm]{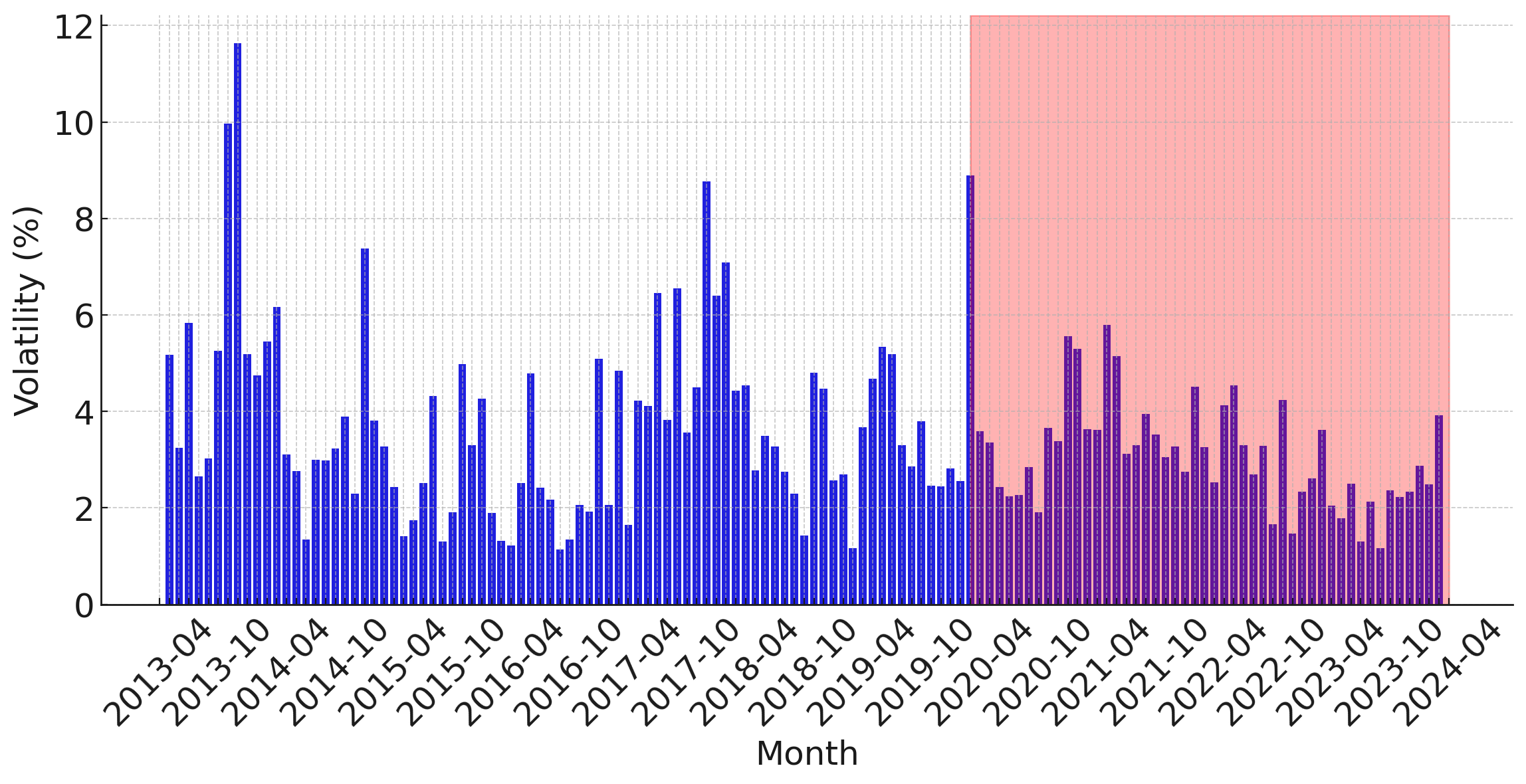}
}
\subfloat[Ethereum]{
    \label{eth-price}
    \includegraphics[width=9cm]{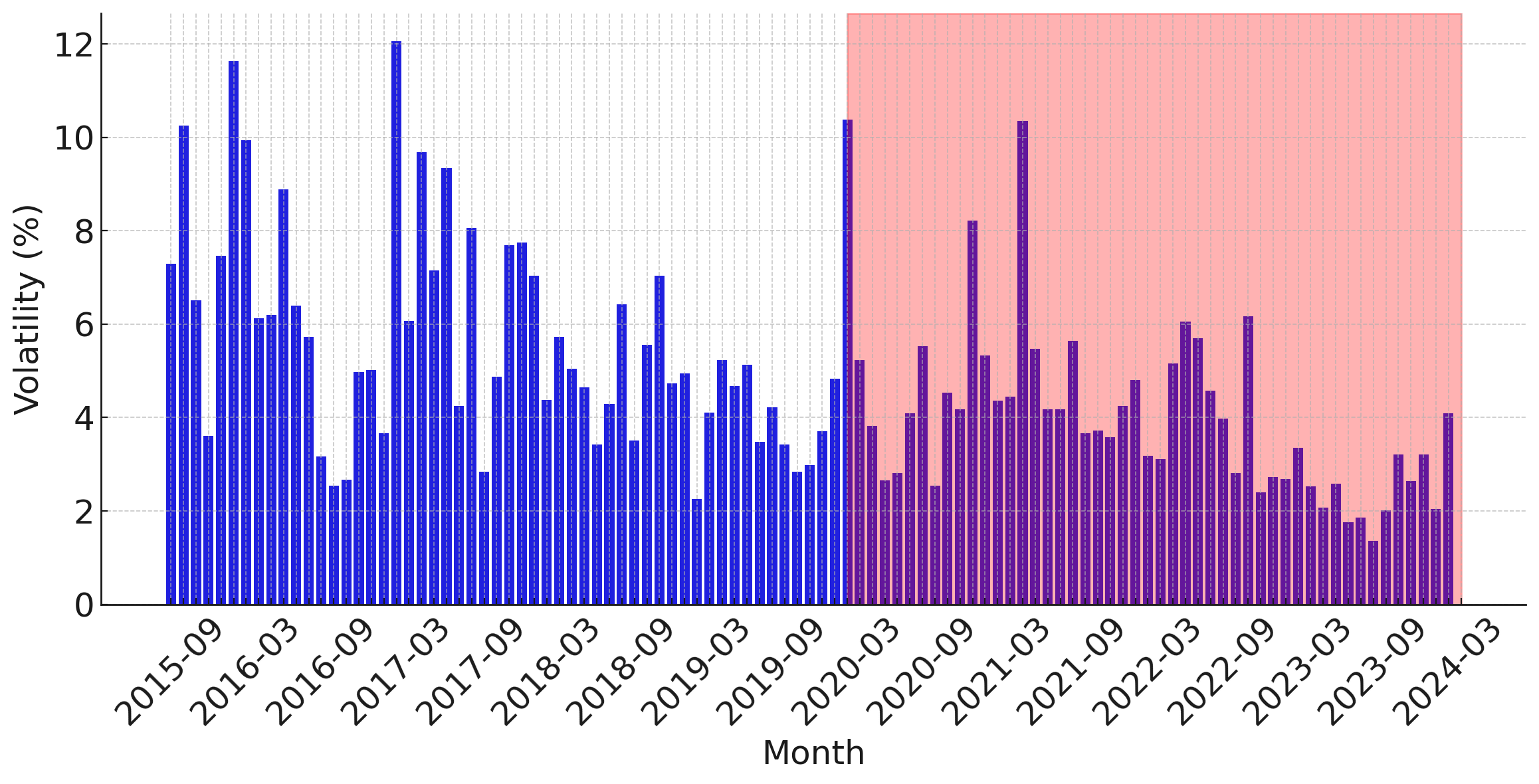}
}
\\
\subfloat[Dogecoin]{
    \label{doge-price}
    \includegraphics[width=9cm]{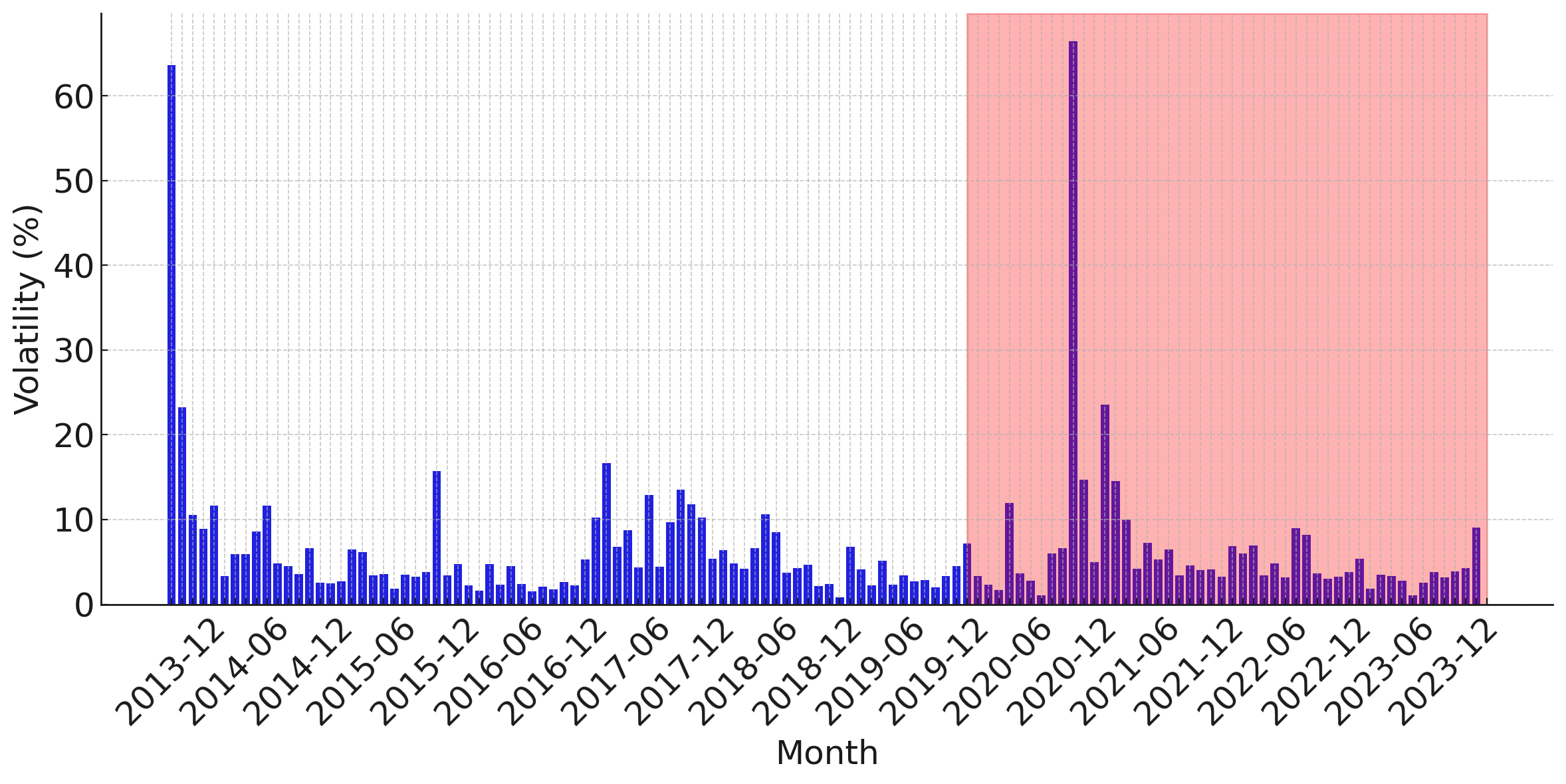}
}
\subfloat[Litecoin]{
    \label{ltc-price}
    \includegraphics[width=9cm]{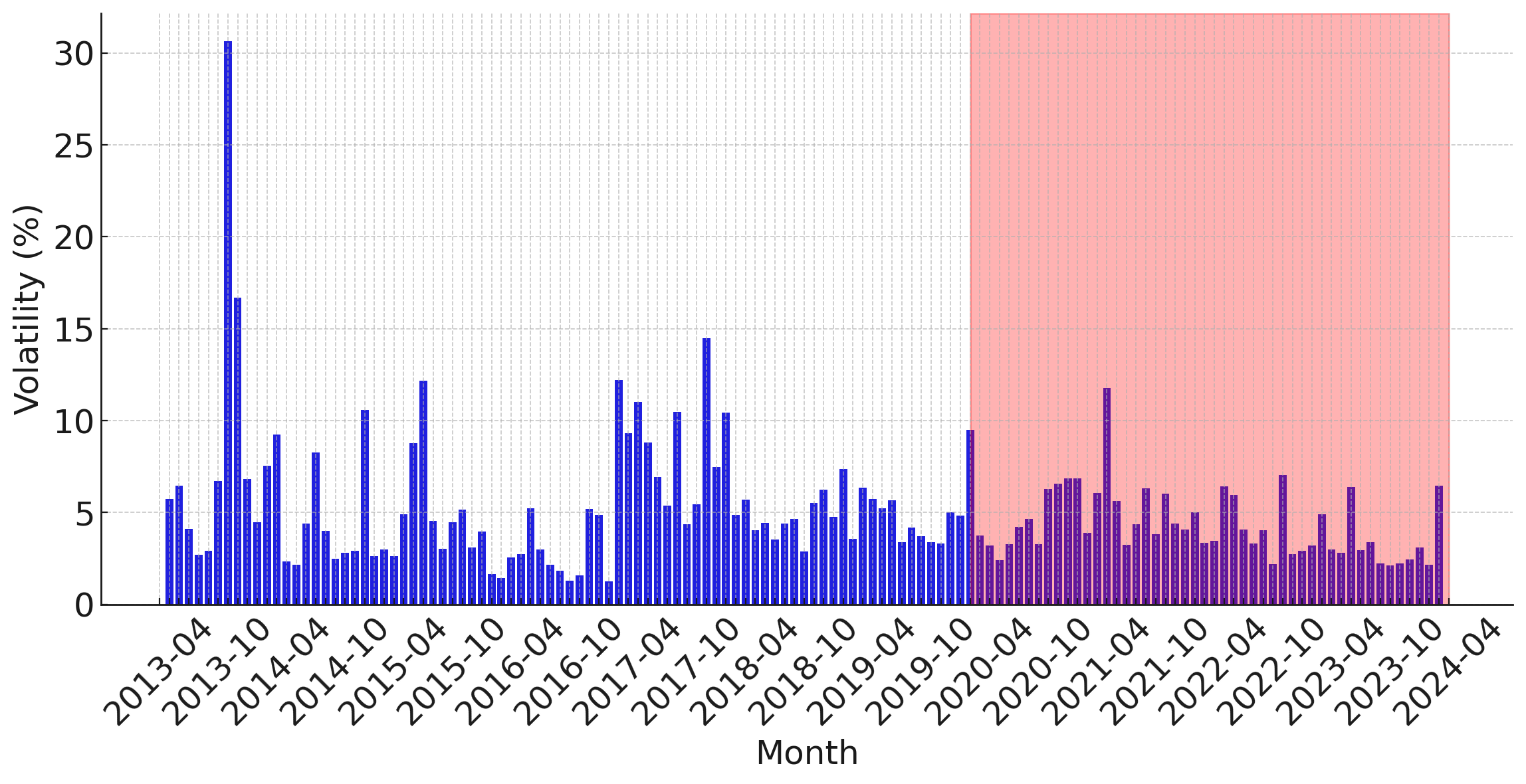}
}
\caption{Monthly volatility of cryptocurrency highlighting COVID-19 trend in pink.}
\label{fvolatility}
\end{figure*}

We investigate the trends for the four selected cryptocurrencies over the given period covering COVID-19. We cover all the phases of COVID-19, including its initiation, spread, and decline. Figure \ref{fclose} presents the close price of Bitcoin, Ethereum, Dogecoin and Litecoin across the selected period, with the  shaded region (pink) indicating COVID-19. We can observe  that the closing price of each cryptocurrency exhibited large fluctuation within the red area. Litecoin experienced significant volatility before the beginning of  COVID-19, while the price fluctuations of the other three cryptocurrencies (Bitcoin, Dogecoin and Ethereum) before COVID-19 were not significant. This demonstrates that after COVID-19, the price of cryptocurrency is more volatile than before. We observe that Ethereum trend is highly correlated to Bitcoin before and during COVID-19. There is a significant price increase from 2020 to 2022, which was subsequently followed by a decrease and another increase in recovering the price, in the case of Bitcoin and Ethereum. Next, we  present the monthly volatility plot in Figure \ref{fvolatility}, where we observe that  Ethereum and Litecoin generally lie below 10\% during COVID-19 and highlighted (pink). We also show the Bitcoin monthly volatility below 6\% during the same time; however, Dogecoin presents a different trend during COVID-19. The monthly volatility of the Dogecoin reached above 20\% in January  and  May, 2021. During other months, it remained consistently at a value of 15\%. Our analysis reveals that the volatility patterns of 4 cryptocurrencies indicate a significant decrease in volatility in the subsequent month after the periods of high volatility. The monthly volatility during COVID-19  is generally similar to the monthly volatility, prior to the pandemic (2018 onwards).   Although the monthly volatility does not change significantly, it fluctuates significantly when looking at the daily close price across the entire period.

\begin{figure}[htbp]
    \includegraphics[width=7.5cm]{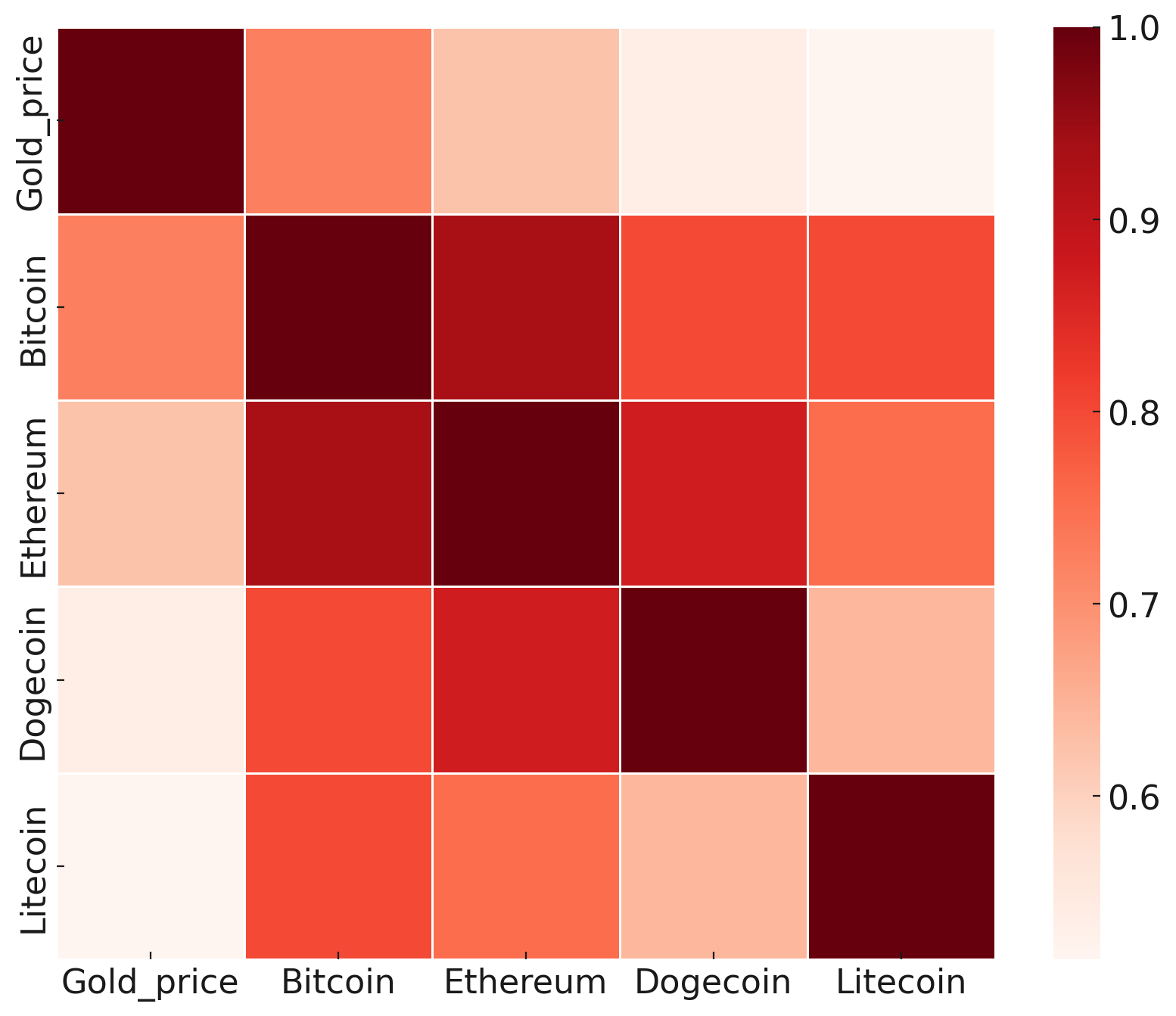}
   \caption{Correlation coefficients between Gold stock price and closing price of four cryptocurrencies.}
   \label{fig:correlation}

\end{figure}

\begin{figure*}[htbp!] 
\subfloat[Correlation coefficients between variables in multivariable models in Bitcoin]{
    \label{btc-corr}
    \includegraphics[width=7.5cm]{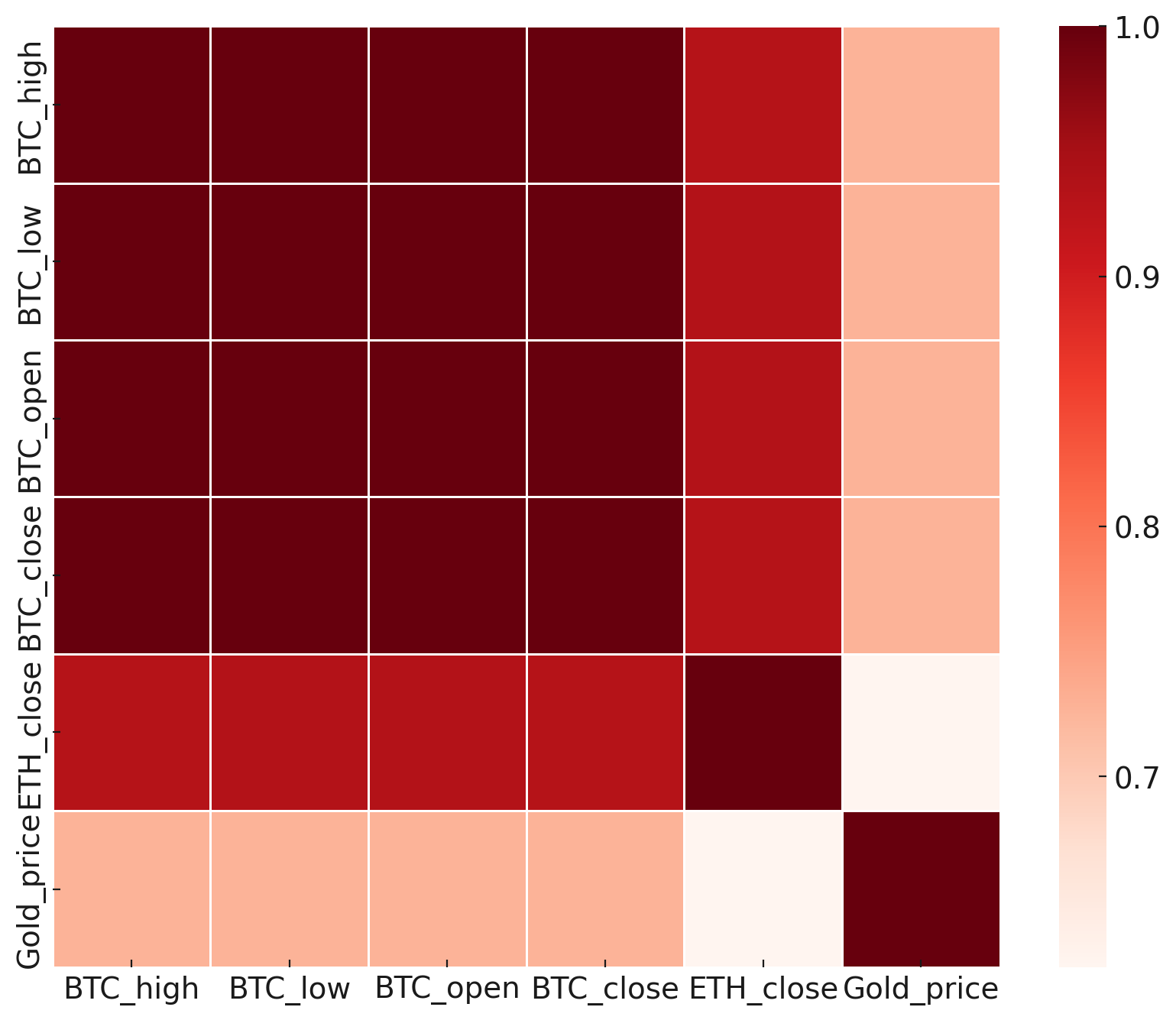}
}
\subfloat[Correlation coefficients between variables in multivariable models in Ethereum]{
    \label{doge-corr}
    \includegraphics[width=7.5cm]{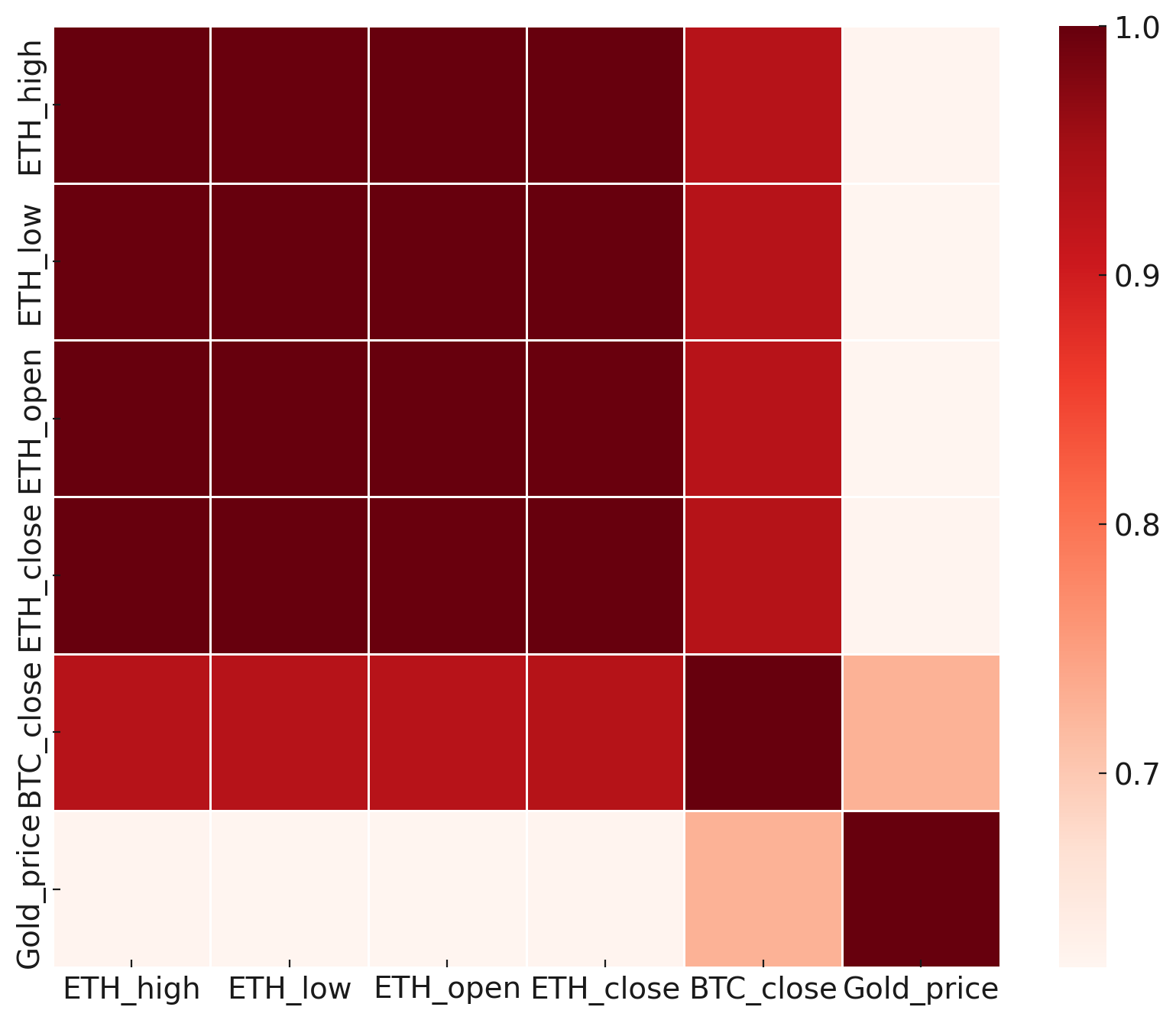}
}
\\
\subfloat[Correlation coefficients between variables in multivariable models in Dogecoin]{
    \label{eth-corr}
    \includegraphics[width=7.5cm]{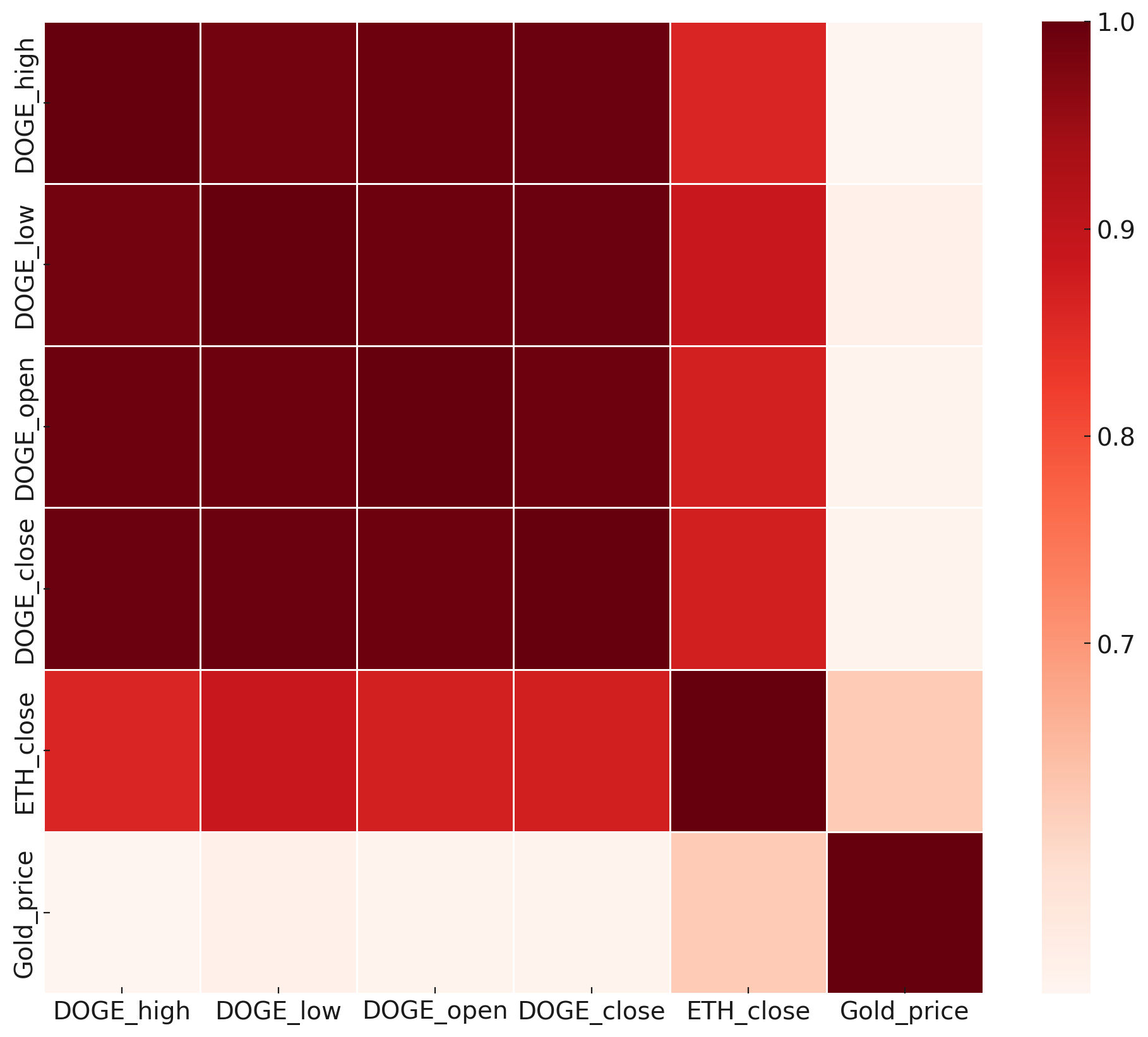}
}
\subfloat[Correlation coefficients between variables in multivariable models in Litecoin]{
    \label{ltc-corr}
    \includegraphics[width=7.5cm]{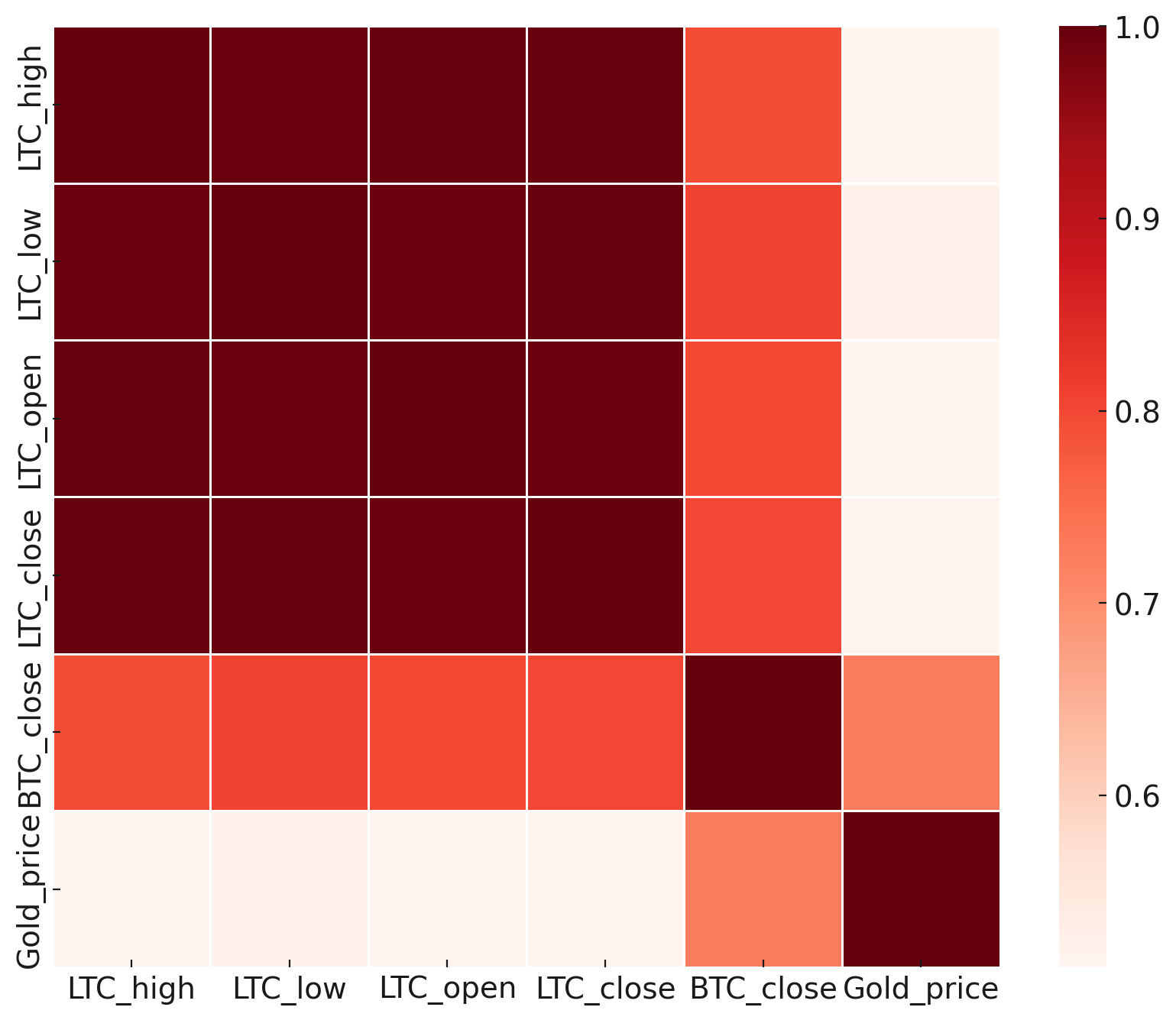}
}
\caption{Heatmap of Pearson correlation coefficients for the multivariate models.}
\label{fcorr}
\end{figure*}


Since we will develop a multivariate model, we also need to provide analyses of how different features of the cryptocurrency (low, high, open, and close price) are correlated with the Gold price.
Figure \ref{fig:correlation} shows the correlations between the features of the multivariate model in each cryptocurrency using Pearson correlation. We observe that close-price is highly correlated to the low-price, high-price and open price. We observe that there is a lower correlation between Gold and other features; however, we will use Gold in our multivariate model as data that is outside the crypto ecosystem, but linked to it. We also find that Gold price has the highest correlation with Bitcoin, followed by Ethereum and Litecoin, and the least with Dodgecoin. Figure \ref{fcorr} presents the Pearson correlation for the respective features including the close, high, low and opening price for a given cryptocurrency with Gold price and most correlated other cryptocurrency (using Figure \ref{fig:correlation}), which is Ethereum in the case of Bitcoin, i.e. Figure \ref{fcorr} -Panel (a). We will use this for multivariate prediction strategy using data processing as shown in Figure \ref{fmulti}.
 
\subsection{Results: pre-COVID-19} 
 We next implement the investigations outlined in Step 4 (Experiment 1) of Framework (Figure \ref{fframework}, where we compare the selected deep learning models and univariate and multivariate strategies using training dataset pre-COVID-19. Note that our test dataset includes the first phase of COVID-19 (Table \ref{tab:data-descr}).
 
We present the results for each prediction horizon (step) obtained from 30 independent experimental runs (mean RMSE  and 95\% confidence interval) that feature  model training  using different initial weights and biases. We note that \textit{robustness} is the degree of confidence in a forecast, which is indicated by a low confidence interval. Moreover, \textit{scalability} refers to the capacity to maintain constant performance as the prediction horizon expands. Our main focus is the performance (RMSE) on the test dataset, both in terms of the mean of 5 prediction horizons, and the individual prediction horizons.  Therefore, in the rest of the discussion, we focus on the test dataset.

\begin{figure*}[htbp!]
\subfloat[Univariate model train and test data (mean of 5 prediction horizons)]{
    \label{11}
    \includegraphics[width=8.4cm]{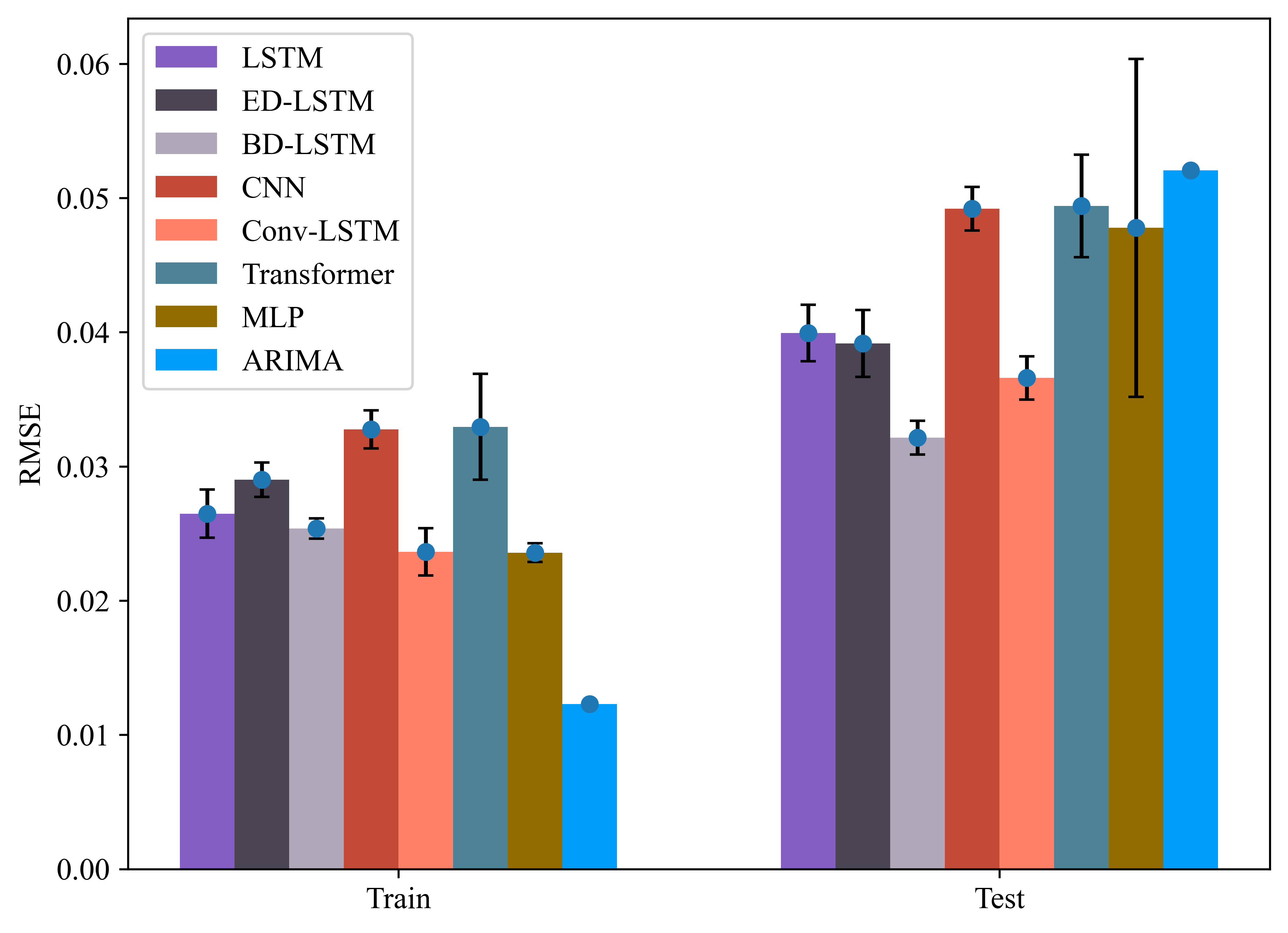}
}
\subfloat[5 step-ahead prediction for test dataset]{
    \label{12}
    \includegraphics[width=7.85cm]{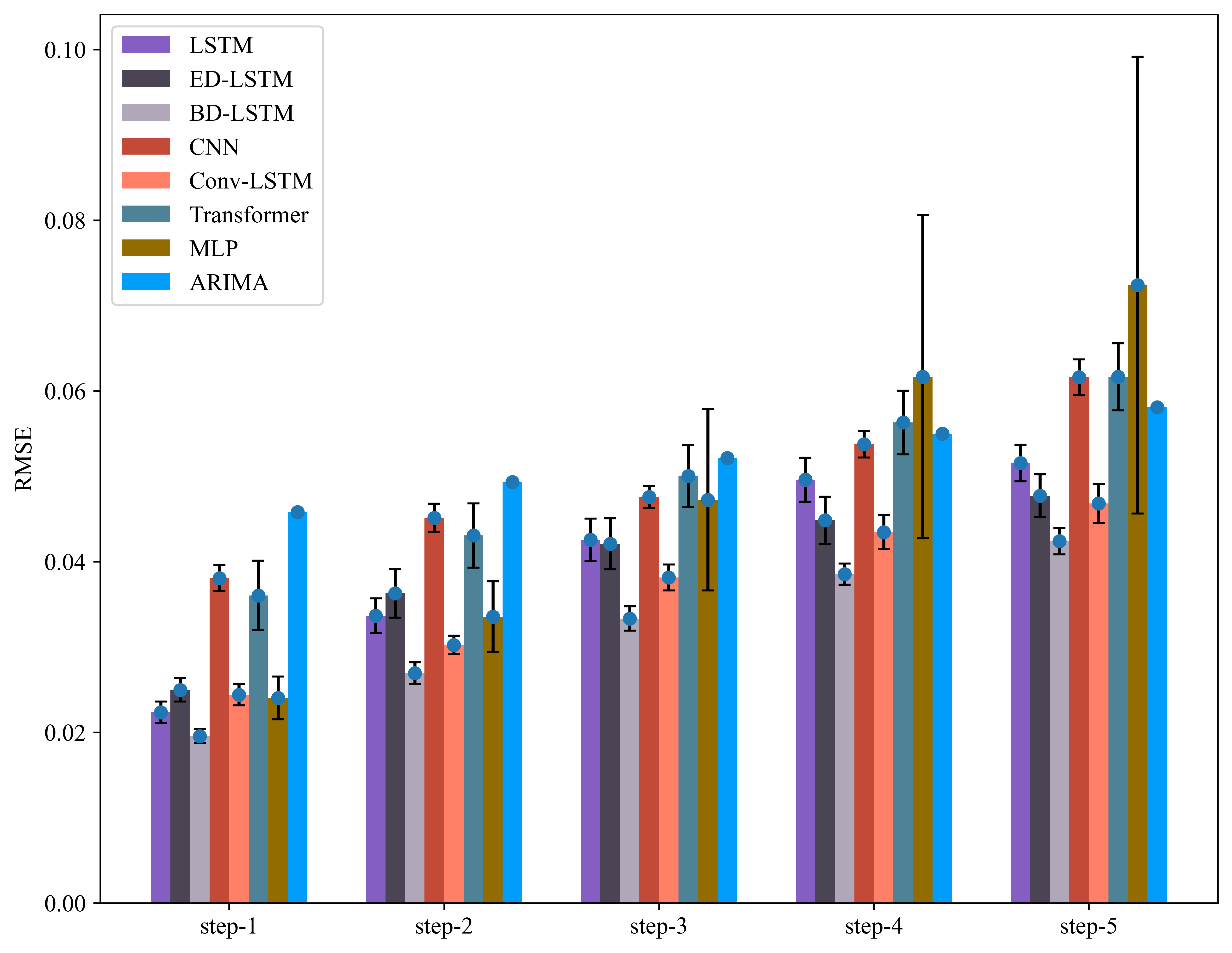}
} 
\caption{Bitcoin: performance evaluation of respective univariate methods (RMSE mean with 95\% confidence interval for 30 experimental runs).}
\label{f421}
\end{figure*}

\begin{figure*}[htbp!]
\subfloat[Multivariate model train and test data (mean of 5 prediction horizons)]{
    \label{15}
    \includegraphics[width=8.4cm]{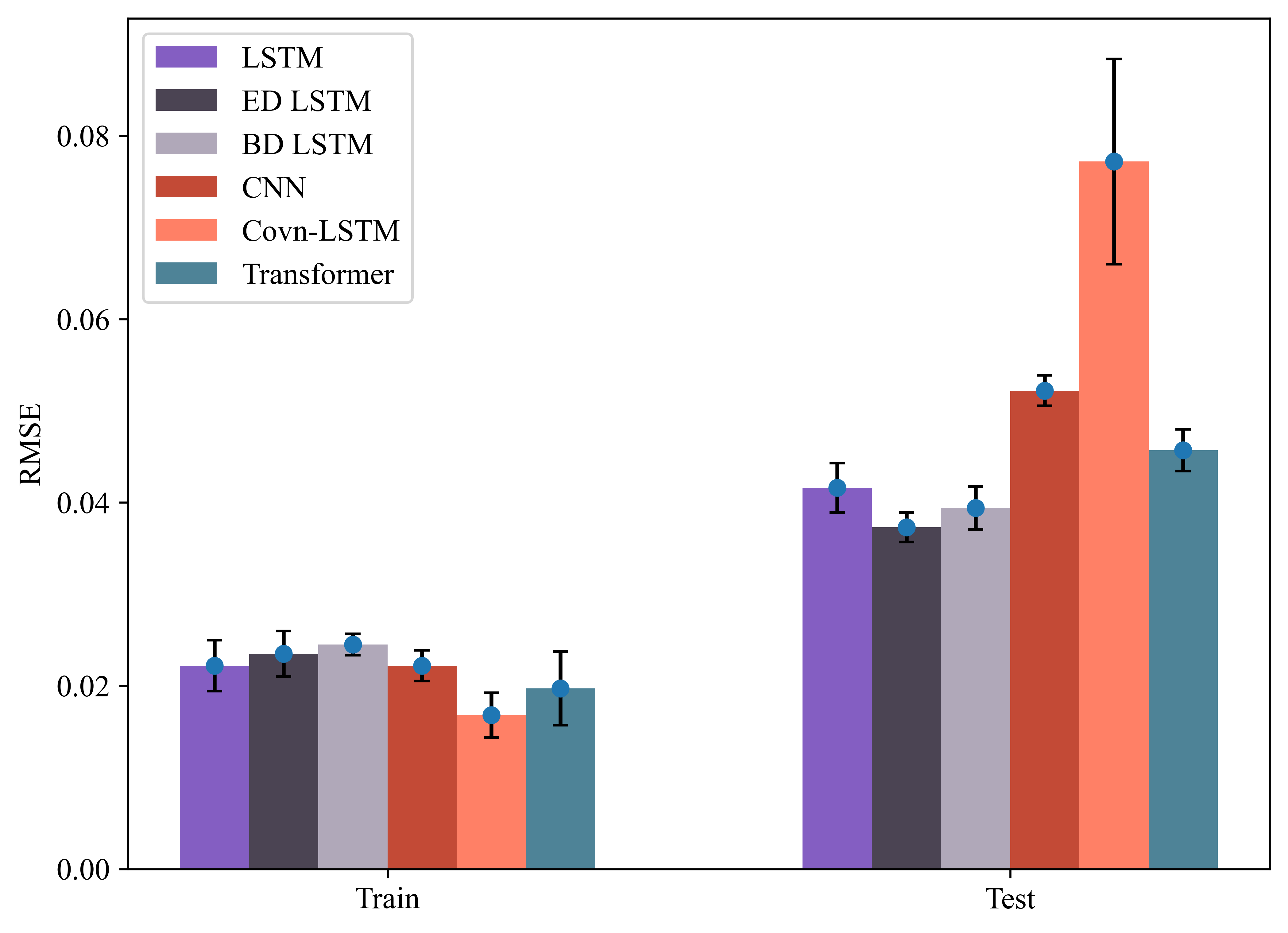}
}
\subfloat[5 step-ahead prediction  for the test dataset]{
    \label{16}
    \includegraphics[width=7.85cm]{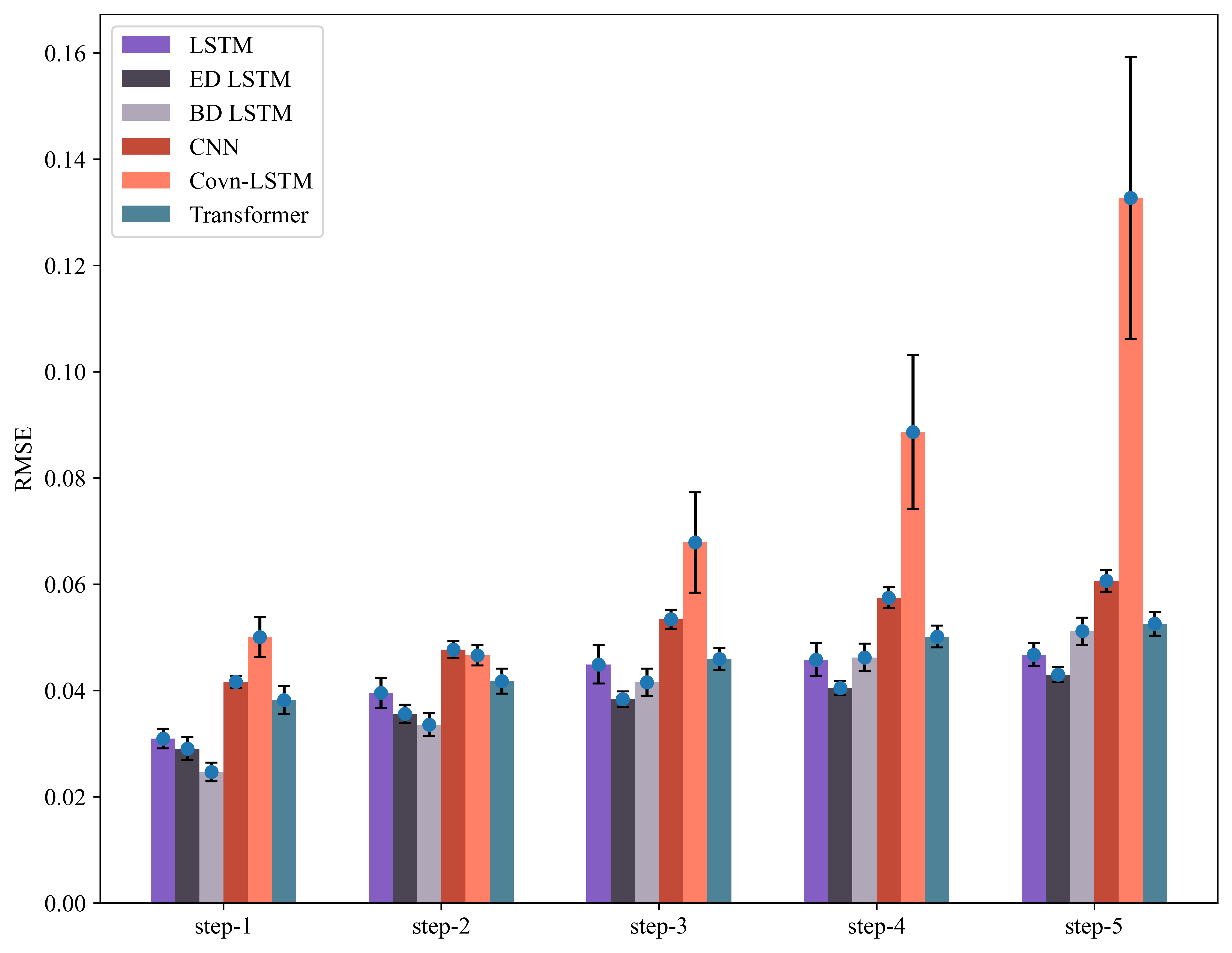}
}
 
\caption{BTC: performance evaluation of respective multivariate methods (RMSE mean  with 95\% confidence interval for 30 experimental runs).}
\label{f422}
\end{figure*}

We first use Bitcoin data to evaluate conventional models  (MLP and ARIMA) when compared to deep learning models (LSTM, ED-LSTM, BD-LSTM, CNN, Conv-LSTM, Transformer), for the univariate (Figure \ref{f421}) and multivariate strategies (\ref{f422}).  The  results show that   MLP and ARIMA perform worse than the deep learning models. MLP exhibits a lack of robustness, and ARIMA model struggles in test prediction accuracy when compared to the deep learning models. We note that ARIMA does the best on the train dataset due to over-training and struggles in generalisation ability.  The deep learning model results are consistent with the finding by Chandra et al. \cite{chandra2021evaluation} where the prediction accuracy of deep learning models is better than conventional machine learning models for  multistep ahead time-series forecasting.  The prediction performance of each model shows a trend where the best Multivariate strategy (ED-LSTM) provides consistent accuracy as the prediction horizon changes when compared to the Univariate strategy (BD-LSTM).    In Figure \ref{f422}, the Multivariate strategy shows  that Conv-LSTM provides the   lowest prediction accuracy, while ED-LSTM and BD-LSTM models provide the most accurate predictions.  In Figure \ref{f421}, contrary to the results of the Multivariate strategy, the most robust Univariate model for predicting Bitcoin is Conv-LSTM.

\begin{figure*}[htbp!]
\subfloat[Univariate model accuracy for train and test data (mean of 5 prediction horizons)]{
    \label{21}
    \includegraphics[width=8.4cm]{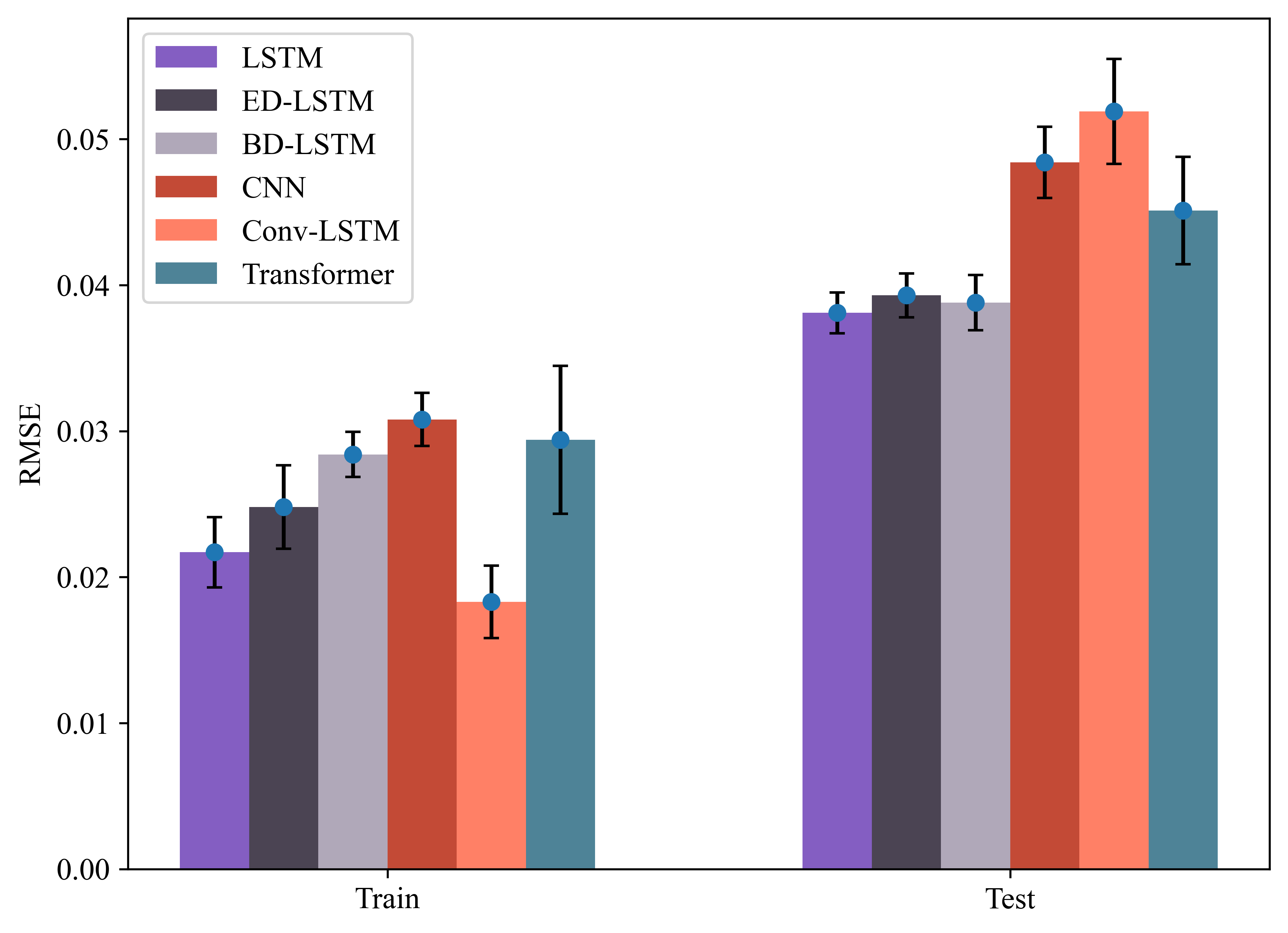}
}
\subfloat[5 step-ahead prediction  for the test dataset]{
    \label{22}
    \includegraphics[width=7.85cm]{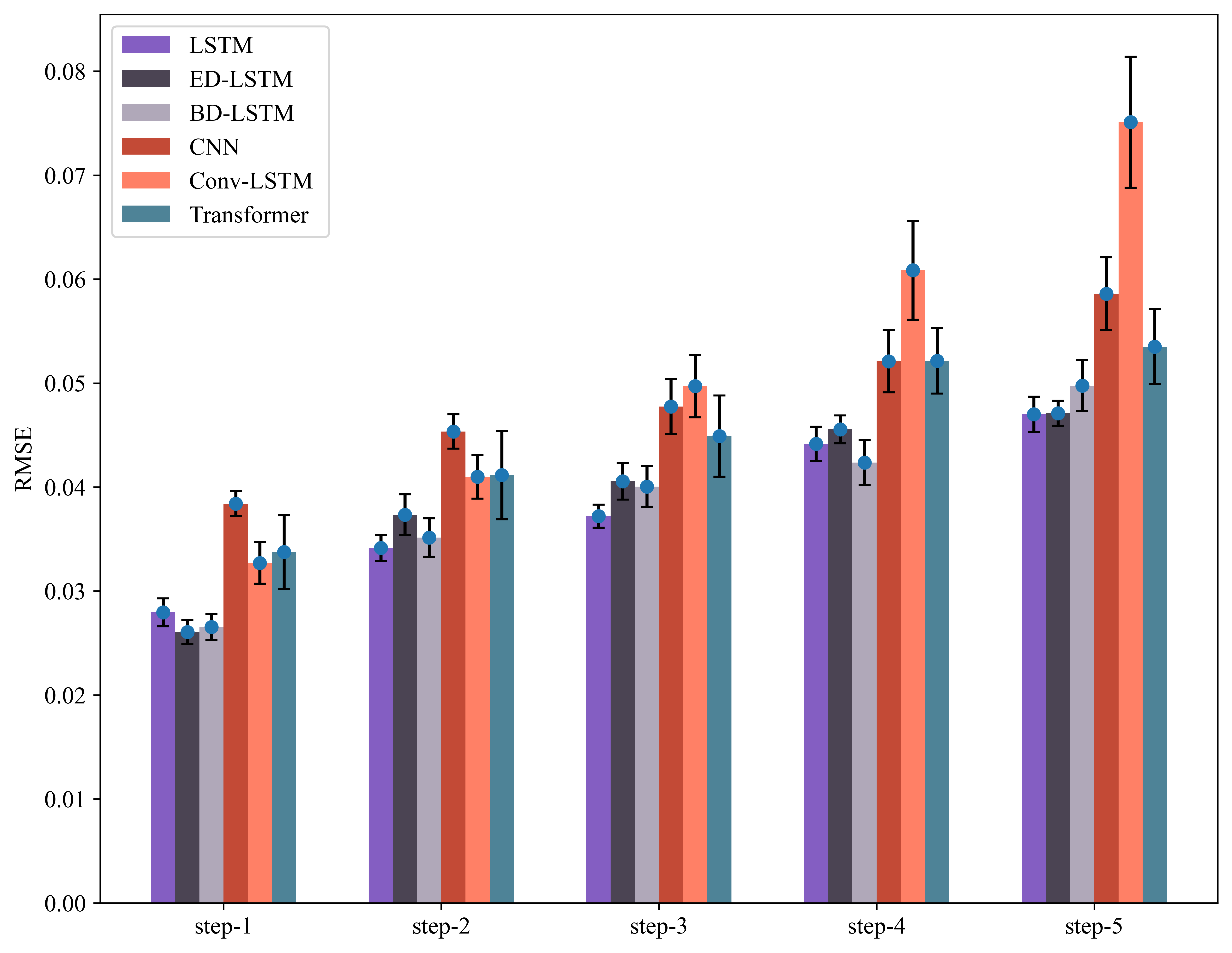}
} 
\caption{ETH: performance evaluation of respective univariate methods (RMSE mean with 95\% CI for 30 experimental runs.)}
\label{f423}
\end{figure*}

\begin{figure*}[htbp!]
\subfloat[Multivariate model accuracy for train and test data (mean of 5 prediction horizons)]{
    \label{25}
    \includegraphics[width=8.4cm]{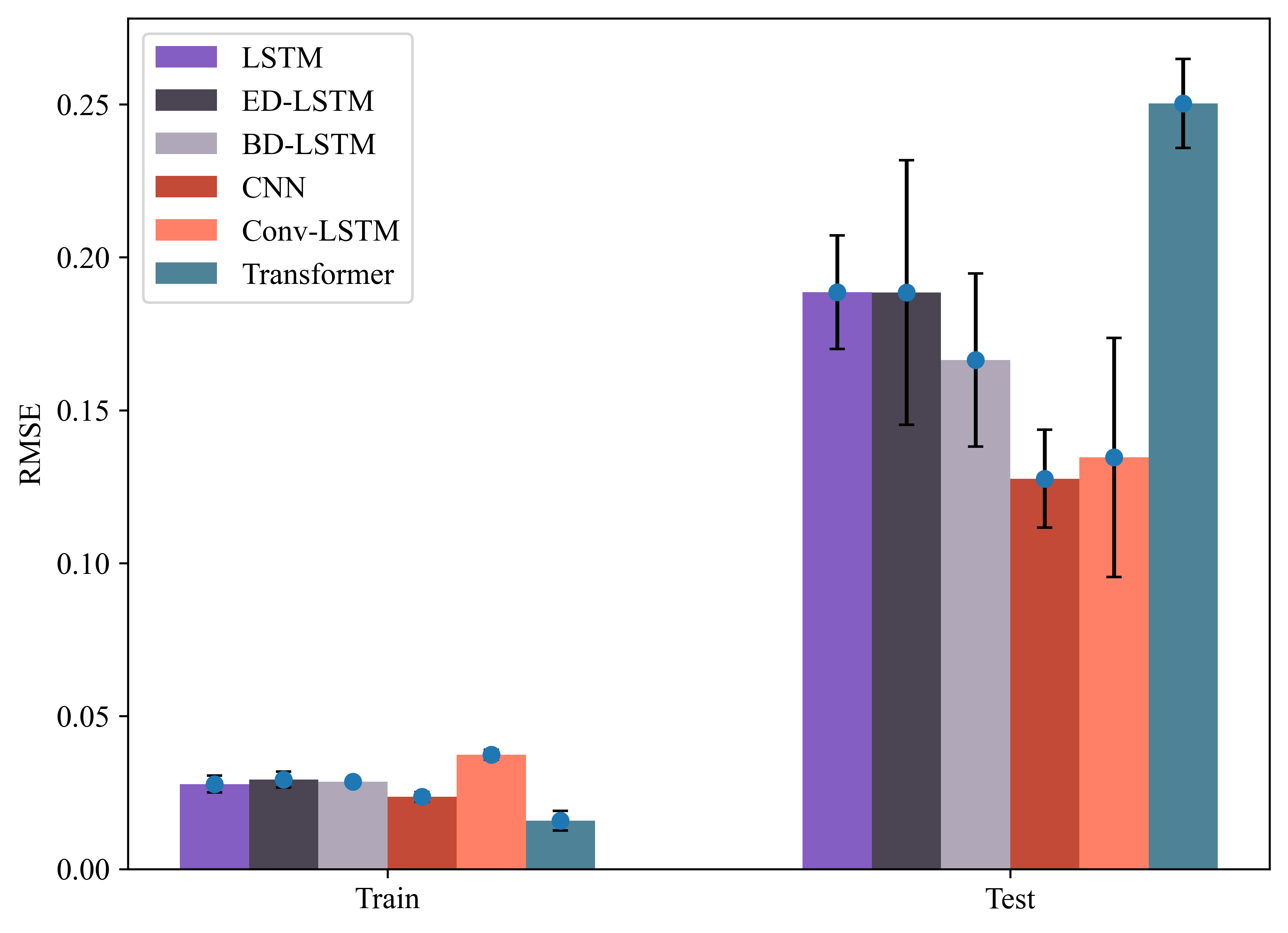}
}
\subfloat[5 step-ahead prediction for the test dataset]{
    \label{26}
    \includegraphics[width=7.85cm]{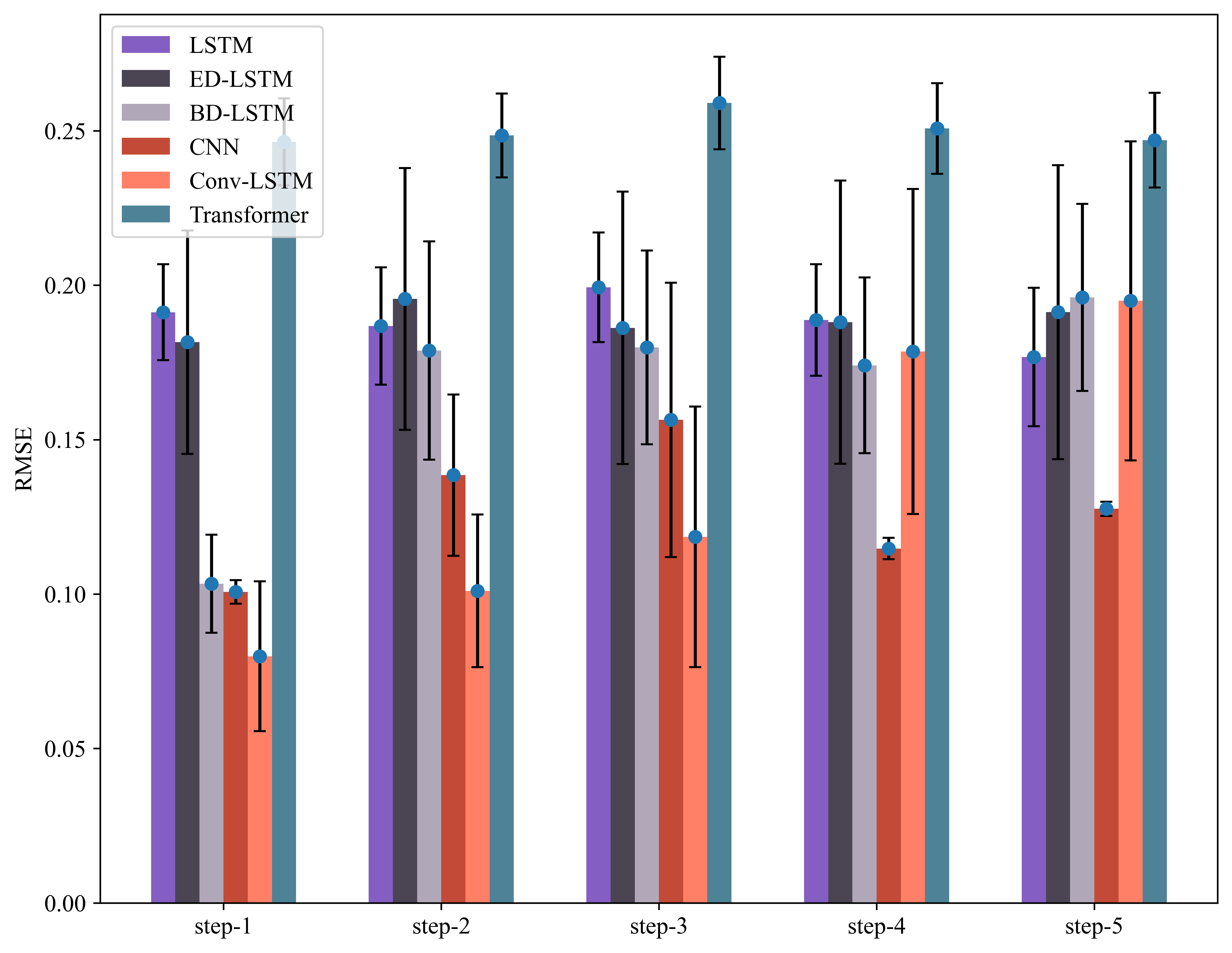}
} 
\caption{ETH: performance evaluation of respective multivariate methods (RMSE mean   with 95\% CI for 30 experimental runs.)}
\label{f424}
\end{figure*}

Figure \ref{f423} presents the results for Ethereum using the Univariate strategy, where we observe that LSTM provides the best test performance, followed by BD-LSTM. Figure \ref{f424}  provides the results for the Multivariate strategy, where CNN provides the best performance which is followed by Conv-LSTM. Notably, the Transformer model provides the best performance. In comparison to the Univariate strategy, we notice that the Multivariate strategy provides a much better test accuracy, which is also more robust and scalable, i.e. higher prediction horizons maintain better accuracy. Furthermore, we note that the Conv-LSTM provides the worst performance in the Univariate case, but one of the best in the Multivariate strategy.

\begin{figure*}[htbp!]
\subfloat[Univariate model accuracy for train and test data (mean of 5 prediction horizons)]{
    \label{31}
    \includegraphics[width=8.4cm]{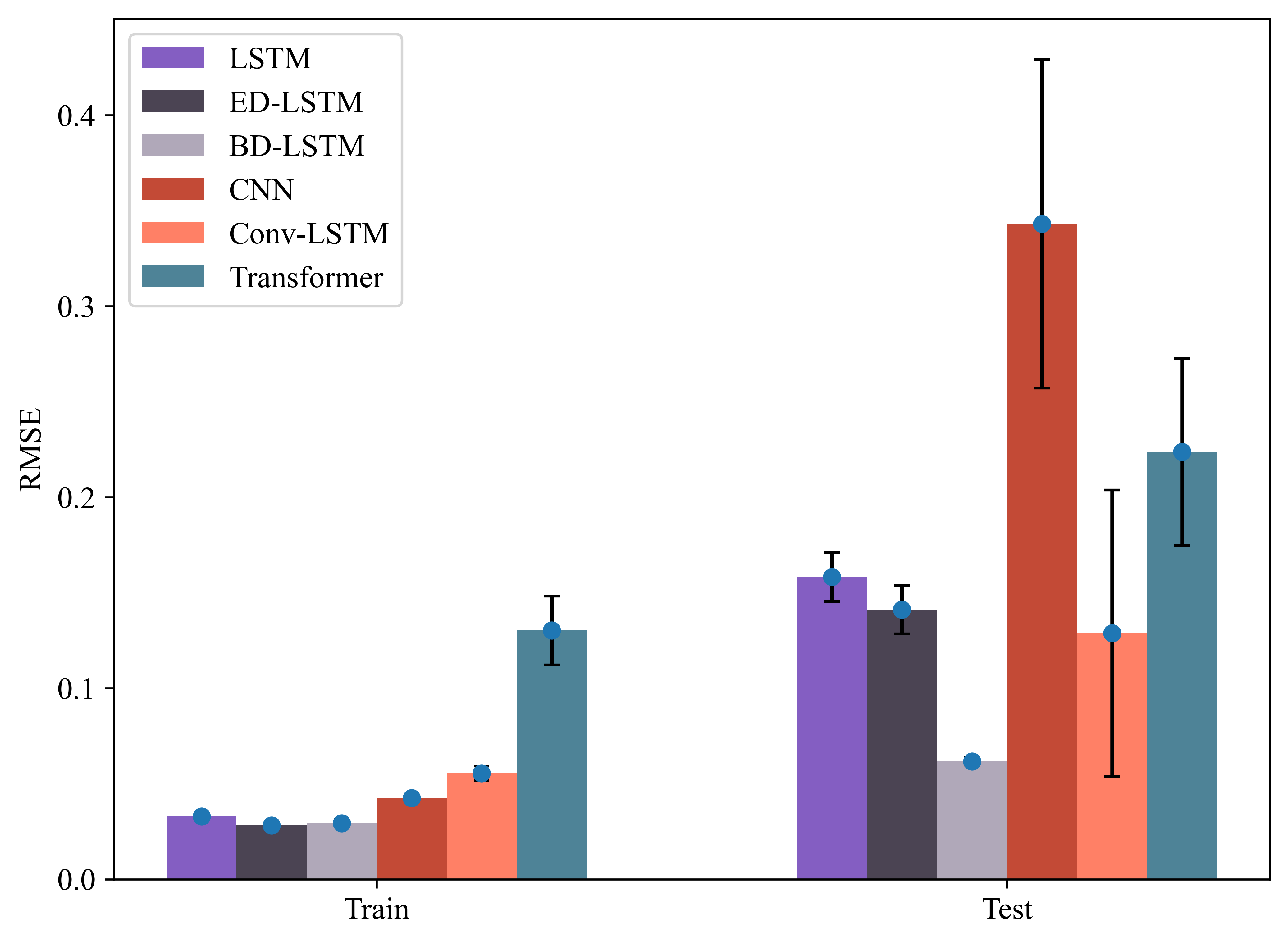}
}
\subfloat[5 step-ahead prediction for the test dataset]{
    \label{32}
    \includegraphics[width=7.85cm]{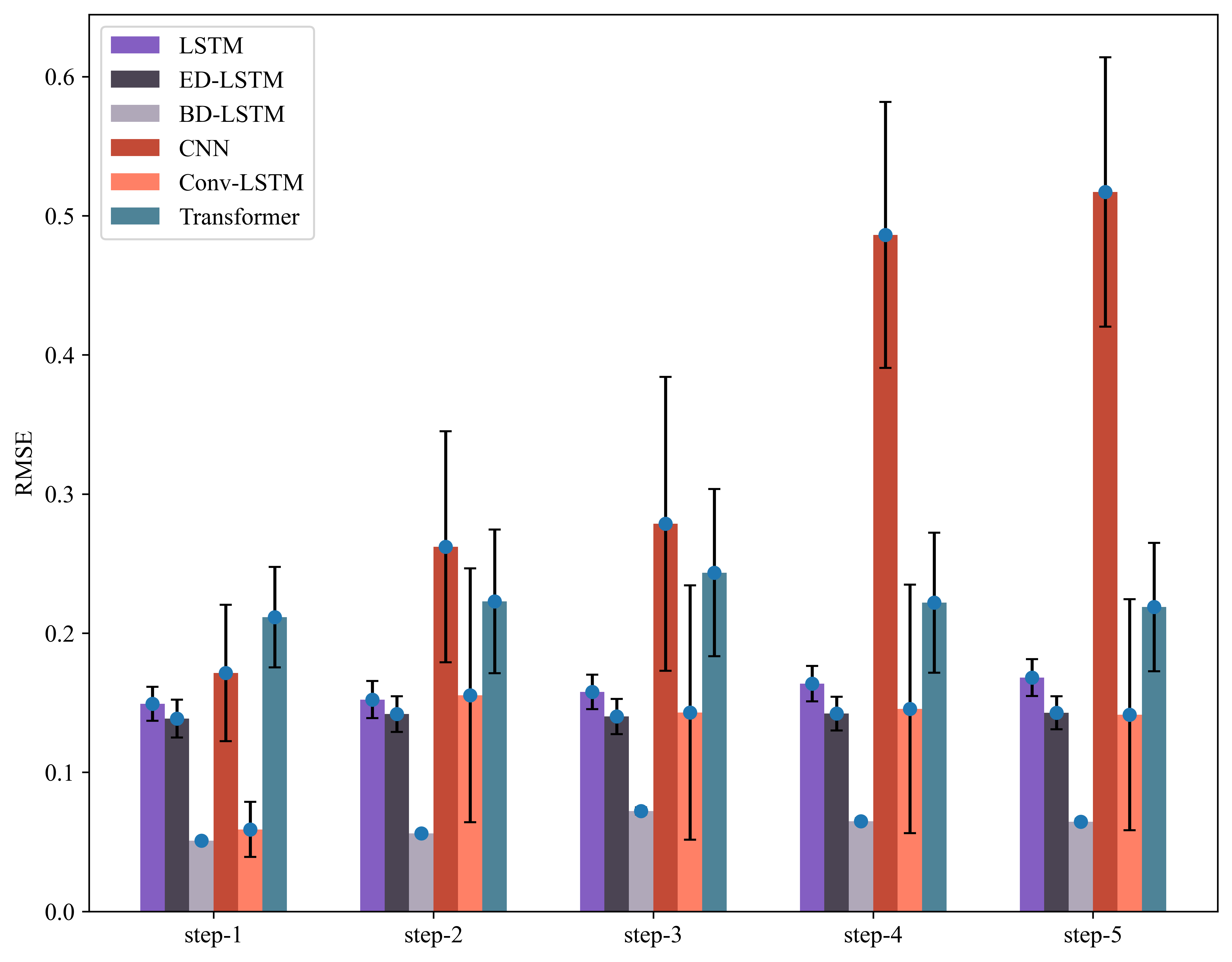}
} 
\caption{DOGE: performance evaluation of respective univariate methods (RMSE mean   with 95\% CI for 30 experimental runs.)}
\label{f425}
\end{figure*}

\begin{figure*}[htbp!]
\subfloat[Multivariate model accuracy for train and test data (mean of 5 prediction horizons)]{
    \label{35}
    \includegraphics[width=8.4cm]{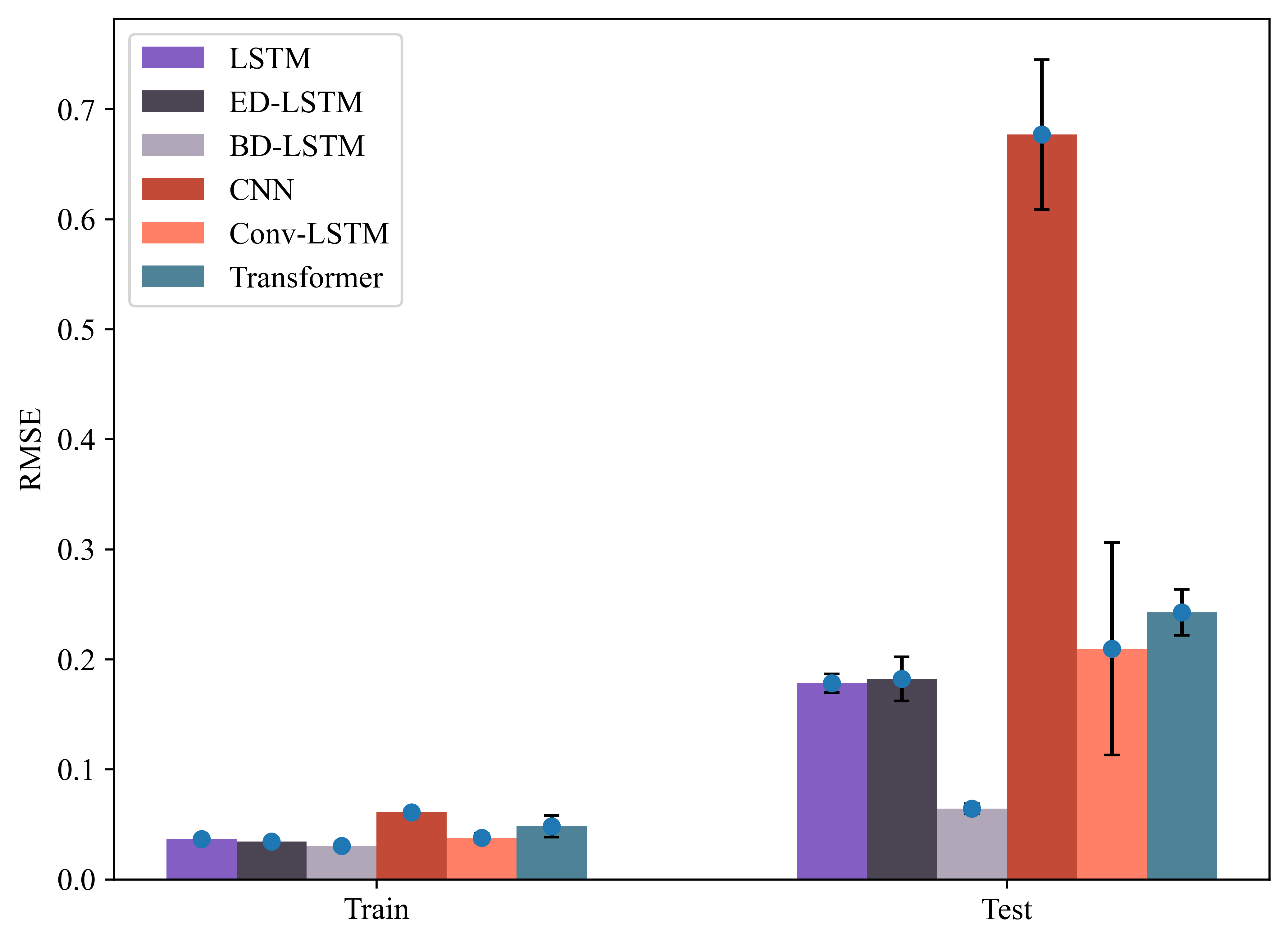}
}
\subfloat[5 step-ahead prediction for the test dataset]{
    \label{36}
    \includegraphics[width=7.85cm]{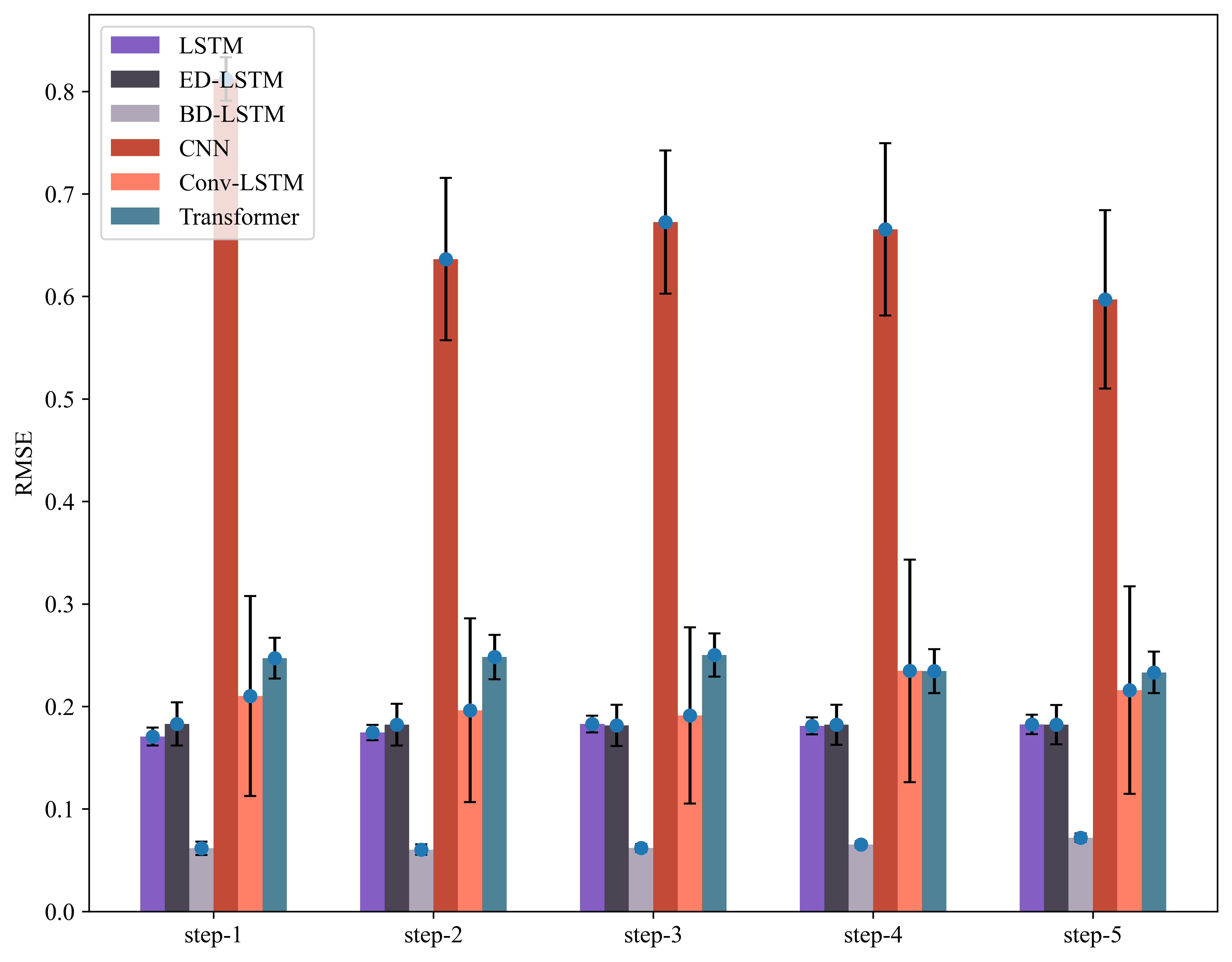}
} 
\caption{DOGE: performance evaluation of respective multivariate methods (RMSE mean with 95\% CI for 30 experimental runs.)}
\label{f426}
\end{figure*}

In the case of Dodgecoin,   Figures \ref{f425} and \ref{f426} reveal that BD-LSTM exhibits the best accuracy, both for the Univariate and the Multivariate strategies, and also provides similar stability for higher prediction horizons. This could be due to the price and vitality  trends in Figures \ref{fclose} and \ref{fvolatility} (Panels c), where we notice that Dodgecoin has a similar trend pre-COVID-19 and during the first phase of COVID-19 which makes Dataset 1 used for these experiments. Furthermore, we also note that in Figure \ref{fcorr} (Panel c), Dodgecoin is least correlated with the Gold stock price,  which is the major factor making a difference in the multivariate model. We notice that CNN provides the worst accuracy formed by the Transformer model in both strategies.

\begin{figure*}[htbp!]
\subfloat[Univariate model accuracy for train and test data (mean of 5 prediction horizons)]{
    \label{41}
    \includegraphics[width=8.4cm]{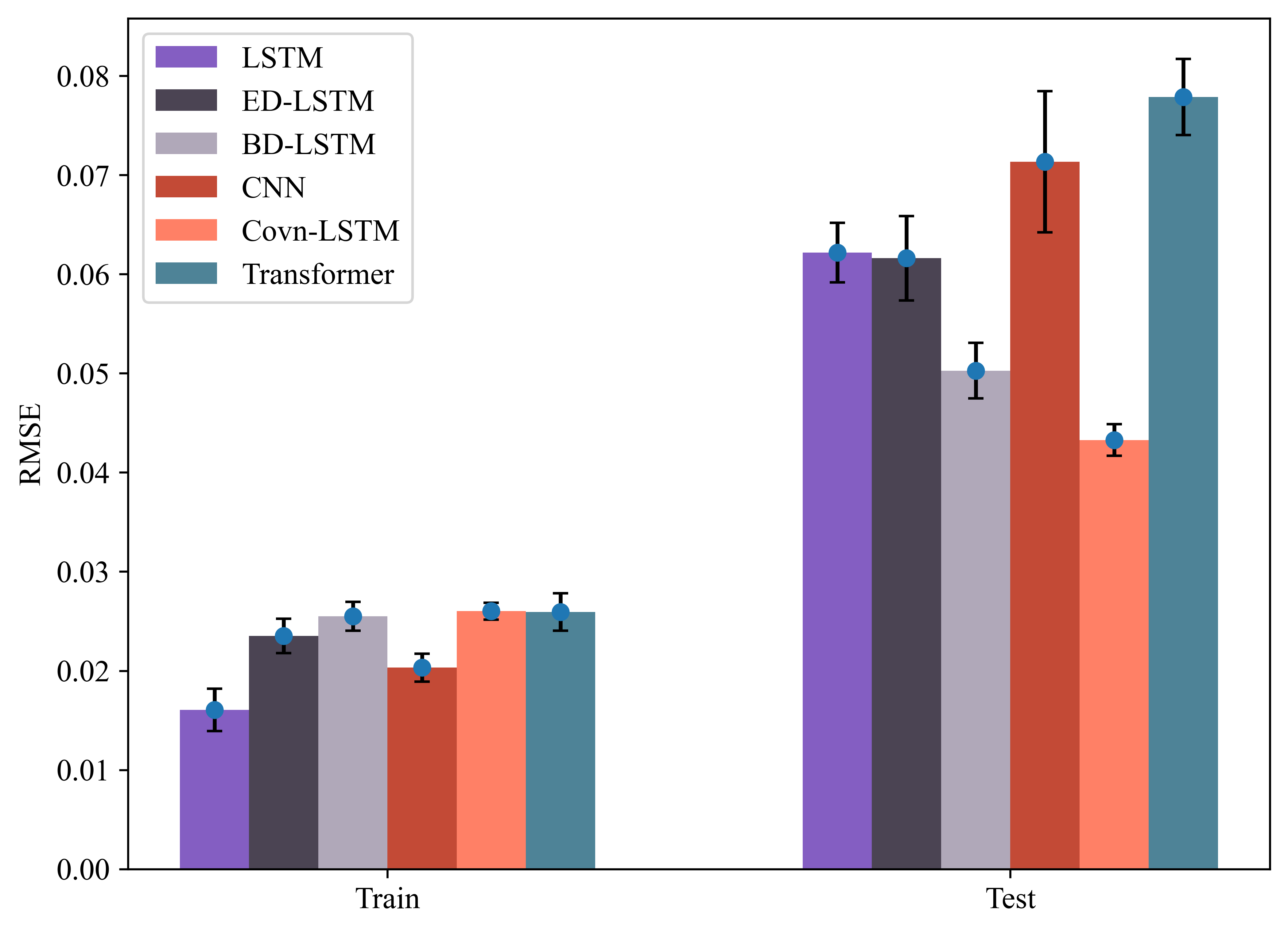}
}
\subfloat[5 step-ahead prediction for the test dataset]{
    \label{42}
    \includegraphics[width=7.85cm]{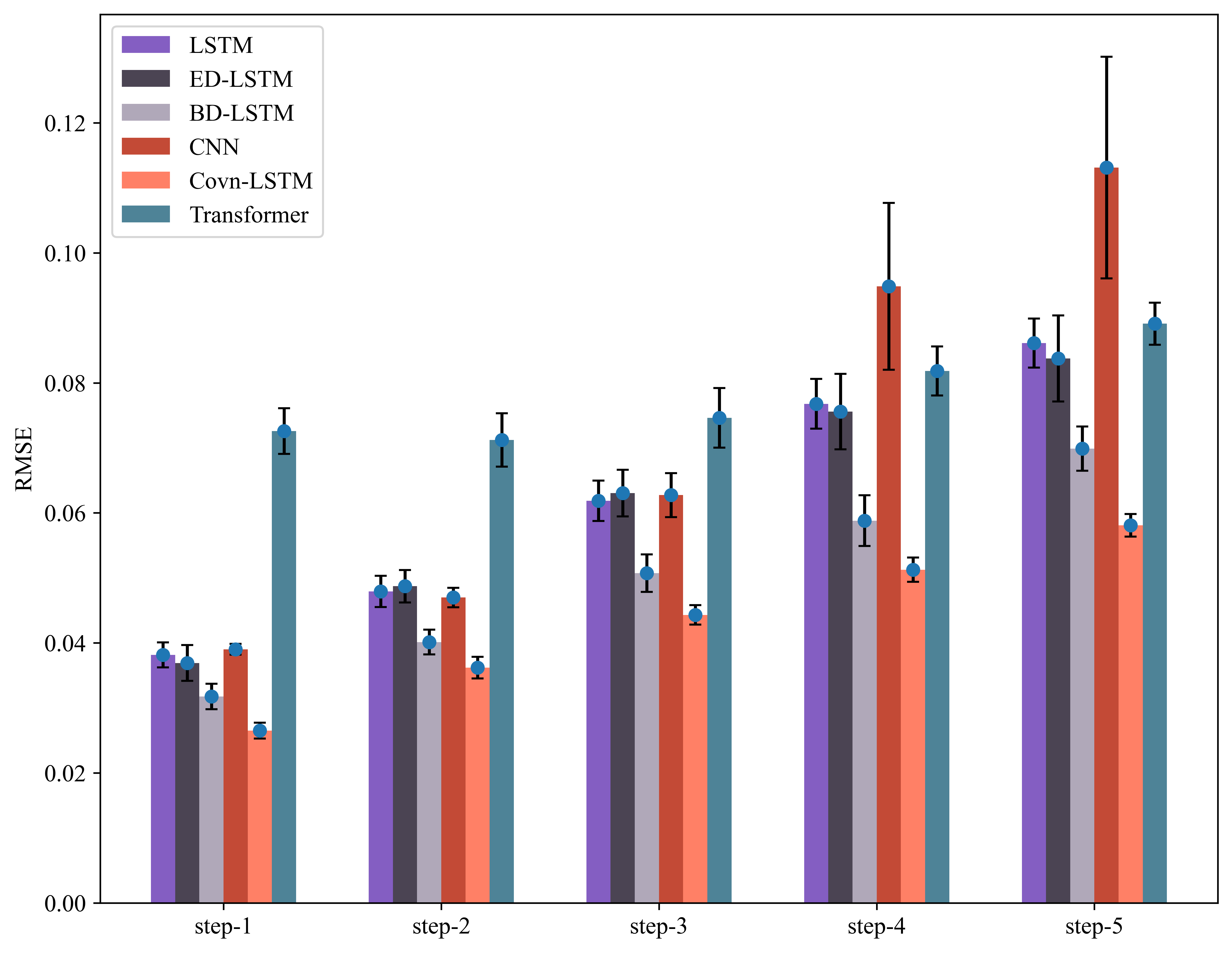}
} 
\caption{LTC: performance evaluation of respective univariate methods (RMSE mean   with 95\% CI for 30 experimental runs.)}
\label{f427}
\end{figure*}

\begin{figure*}[htbp!]
\subfloat[Multivariate model  accuracy for train and test data (mean of 5 prediction horizons)]{
    \label{45}
    \includegraphics[width=8.4cm]{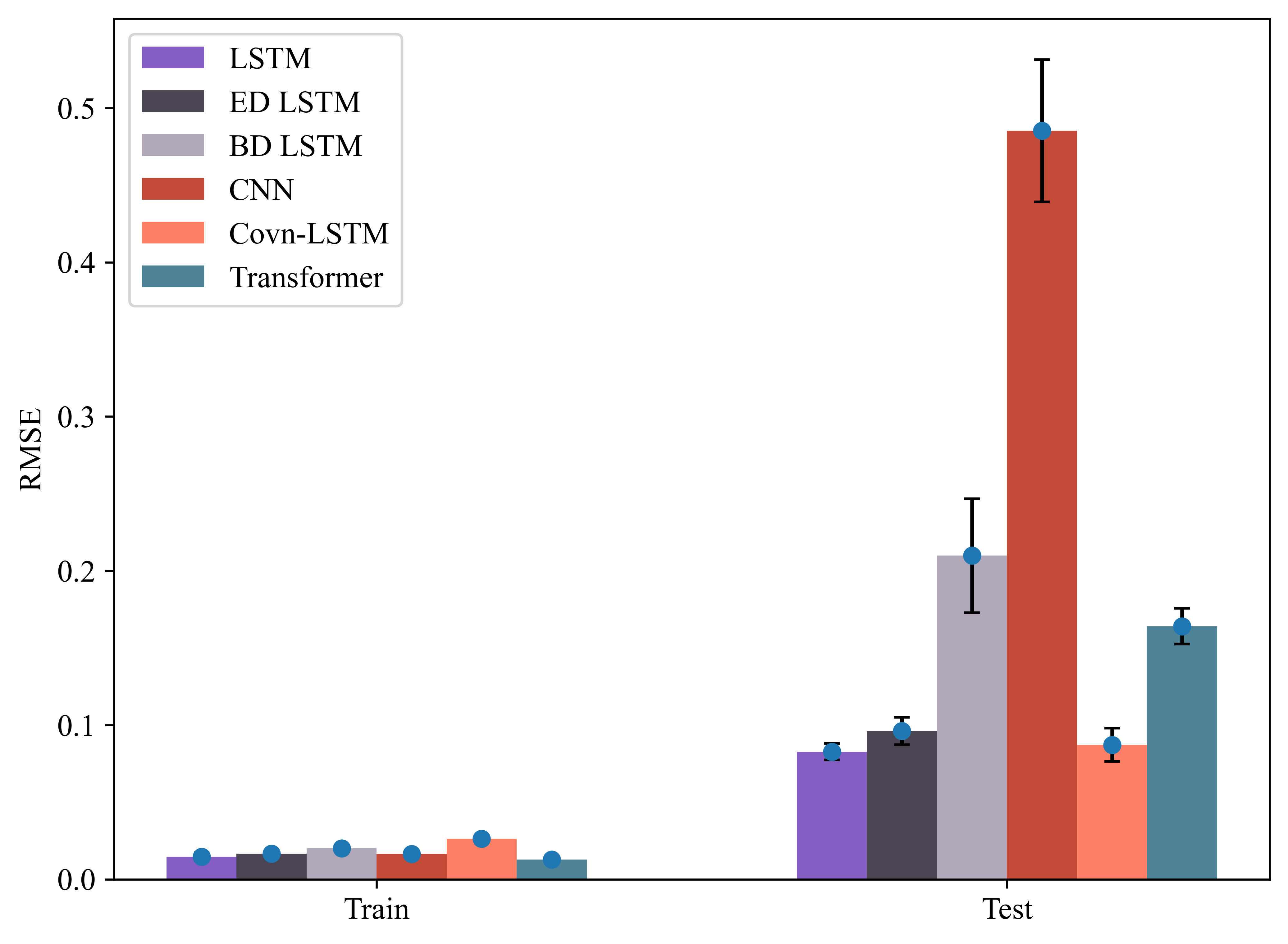}
}
\subfloat[5 step-ahead prediction for the test dataset]{
    \label{46}
    \includegraphics[width=7.85cm]{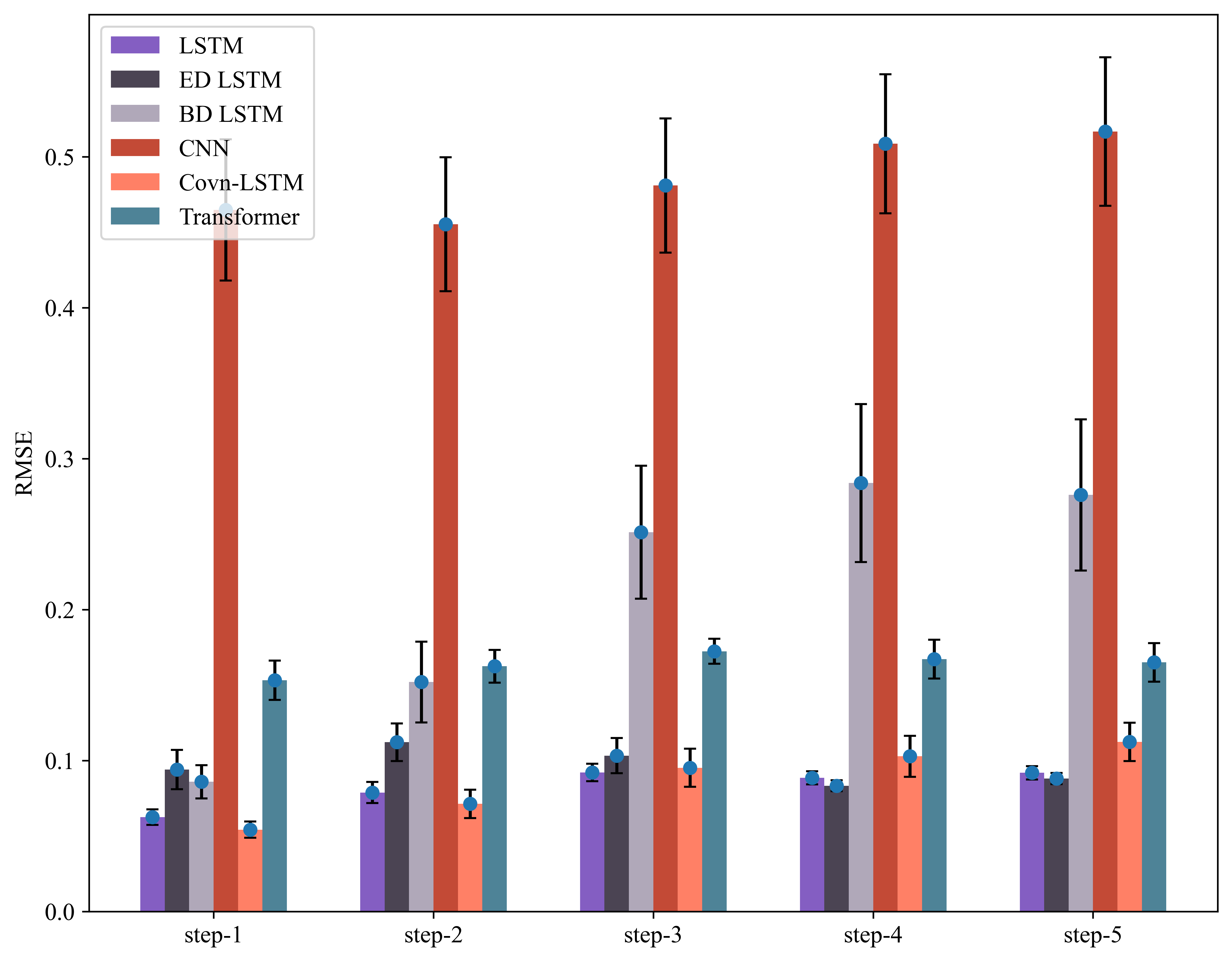}
} 
\caption{LTC: performance evaluation of respective multivariate methods (RMSE mean with 95\% CI for 30 experimental runs.)}
\label{f428}
\end{figure*}

Finally, we present the results for the Litcoin for both Univariate and Multivariate strategies.  Figures \ref{f427}  and  \ref{f428} show all the results of the Univariate models, where we find that the Conv-LSTM and BD-LSTM show the best performance, whereas LSTM, ED-LSTM and Conv-LSTM provide the best performances for the Multivariate strategies. On the contrary, the CNN provides the worst performance for the Multivariate strategy, much higher in magnitude when compared to the rest of the models. We also notice that the Multivariate strategy provides better stability as the prediction horizon increases, when compared to the Univariate strategy. The Univariate strategy provides much better accuracy of the best models when compared to the Multivariate strategy. 




\begin{table*}[htbp!]
    \centering
    \small
    \begin{tabular}{ p{0.322in} p{0.48in} c c c c c c c }
    \hline 
    \hline
 Data &  Strategy  &Model&Step 1&Step 2&Step 3&Step 4&Step 5 & Test Mean \\
        \hline
 BTC  &  Univariate & CNN & 0.0380$\pm$0.0015 & 0.0451$\pm$0.0017 & 0.0476$\pm$0.0013 & 0.0537$\pm$0.0015 & 0.0616$\pm$0.0021 & 0.0492$\pm$0.0016 \\
   &  & LSTM & 0.0223$\pm$0.0013 & 0.0337$\pm$0.0020 & 0.0425$\pm$0.0025 & 0.0496$\pm$0.0026 & 0.0515$\pm$0.0022 & 0.0399$\pm$0.0021 \\
   &  & ED-LSTM & 0.0250$\pm$0.0014 & 0.0363$\pm$0.0029 & 0.0421$\pm$0.0030 & 0.0448$\pm$0.0028 & 0.0477$\pm$0.0025 & 0.0392$\pm$0.0025 \\
   &  & BD-LSTM & \textbf{0.0196$\pm$0.0008} & \textbf{0.0296$\pm$0.0013} & \textbf{0.0333$\pm$0.0014} & \textbf{0.0385$\pm$0.0012} & \textbf{0.0424$\pm$0.0015} & \textbf{0.0327$\pm$0.0012} \\
  &  & Conv-LSTM & 0.0244$\pm$0.0012 & 0.0302$\pm$0.0011 & 0.0381$\pm$0.0015 & 0.0434$\pm$0.0020 & 0.0468$\pm$0.0023 & 0.0366$\pm$0.0016 \\
   &  & Transformer & 0.0360$\pm$0.0041 & 0.0431$\pm$0.0038 & 0.0500$\pm$0.0036 & 0.0563$\pm$0.0037 & 0.0617$\pm$0.0039 & 0.0494$\pm$0.0038 \\
   \hline 
   & Multivariate & CNN & 0.0416$\pm$0.0011 & 0.0477$\pm$0.0016 & 0.0534$\pm$0.0018 & 0.0575$\pm$0.0019 & 0.0606$\pm$0.0021 & 0.0522$\pm$0.0017 \\
    &  & LSTM & 0.0310$\pm$0.0018 & 0.0395$\pm$0.0029 & 0.0449$\pm$0.0036 & 0.0458$\pm$0.0031 & 0.0467$\pm$0.0022 & 0.0416$\pm$0.0027 \\
 &  & ED-LSTM & 0.0290$\pm$0.0022 & 0.0356$\pm$0.0017 & \textbf{0.0384$\pm$0.0015} & \textbf{0.0405$\pm$0.0013} & \textbf{0.0430$\pm$0.0014} & \textbf{0.0373$\pm$0.0016} \\
    & & BD-LSTM & \textbf{0.0247$\pm$0.0018} & \textbf{0.0336$\pm$0.0022} & 0.0415$\pm$0.0026 & 0.0462$\pm$0.0026 & 0.0511$\pm$0.0026 & 0.0394$\pm$0.0024 \\
    &  & Conv-LSTM & 0.0500$\pm$0.0037 & 0.0466$\pm$0.0019 & 0.0678$\pm$0.0095 & 0.0886$\pm$0.0144 & 0.1327$\pm$0.0266 & 0.0771$\pm$0.0112 \\
   & & Transformer & 0.0382$\pm$0.0026 & 0.0418$\pm$0.0024 & 0.0459$\pm$0.0021 & 0.0501$\pm$0.0020 & 0.0526$\pm$0.0022 & 0.0457$\pm$0.0023 \\

 \hline
    & Multivariate* & CNN & 0.0202$\pm$0.0007 & 0.0237$\pm$0.0008 & 0.0259$\pm$0.0007 & 0.0286$\pm$0.0008 & 0.0309$\pm$0.0008 & 0.0259$\pm$0.0008 \\
   & & LSTM & 0.0178$\pm$0.0018 & 0.0213$\pm$0.0017 & 0.0241$\pm$0.0012 & 0.0269$\pm$0.0011 & 0.0294$\pm$0.0014 & 0.0237$\pm$0.0014 \\
   & & ED-LSTM & 0.0171$\pm$0.0012 & 0.0214$\pm$0.0015 & 0.0240$\pm$0.0014 & 0.0269$\pm$0.0012 & 0.0294$\pm$0.0011 & 0.02378$\pm$0.0013 \\
   & & BD-LSTM & 0.0175$\pm$0.0010 & 0.0210$\pm$0.0010 & 0.0239$\pm$0.0009 & 0.0263$\pm$0.0007 & 0.0291$\pm$0.0010 & 0.0236$\pm$0.0009 \\
   & & Conv-LSTM & \textbf{0.0147$\pm$0.0004} & \textbf{0.0176$\pm$0.0004} & \textbf{0.0205$\pm$0.0004} & \textbf{0.0224$\pm$0.0006} & \textbf{0.0250$\pm$0.0006} & \textbf{0.0200$\pm$0.0005} \\
   & & Transformer & 0.0186$\pm$0.0008 & 0.0210$\pm$0.0008 & 0.0231$\pm$0.0008 & 0.0257$\pm$0.0008 & 0.0279$\pm$0.0008 & 0.0233$\pm$0.0008\\
        \hline
        \hline
 ETH &    Univariate & CNN & 0.0384$\pm$0.0012 & 0.0453$\pm$0.0017 & 0.0478$\pm$0.0026 & 0.0521$\pm$0.0030 & 0.0586$\pm$0.0035 & 0.0484$\pm$0.0024 \\
    &  & LSTM & 0.0280$\pm$0.0013 & \textbf{0.0341$\pm$0.0012} & \textbf{0.0372$\pm$0.0011} & 0.0441$\pm$0.0016 & \textbf{0.0470$\pm$0.0017} & \textbf{0.0381$\pm$0.0014} \\
  &  & ED-LSTM & \textbf{0.0261$\pm$0.0012} & 0.0373$\pm$0.0019 & 0.0405$\pm$0.0018 & 0.0456$\pm$0.0013 & 0.0471$\pm$0.0012 & 0.0393$\pm$0.0015 \\
 &  & BD-LSTM & 0.0265$\pm$0.0012 & 0.0351$\pm$0.0019 & 0.0401$\pm$0.0019 & \textbf{0.0423$\pm$0.0022} & 0.0498$\pm$0.0025 & 0.0388$\pm$0.0019 \\
  &    & Conv-LSTM & 0.0327$\pm$0.0020 & 0.0410$\pm$0.0021 & 0.0497$\pm$0.0030 & 0.0609$\pm$0.0048 & 0.0751$\pm$0.0063 & 0.0519$\pm$0.0036 \\
  &  & Transformer & 0.0337$\pm$0.0035 & 0.0412$\pm$0.0042 & 0.0449$\pm$0.0039 & 0.0521$\pm$0.0032 & 0.0535$\pm$0.0036 & 0.0451$\pm$0.0037 \\
  \hline
   & Multivariate & CNN & \textbf{0.1007$\pm$0.0038} & \textbf{0.1385$\pm$0.0261} & \textbf{0.1564$\pm$0.0444} & \textbf{0.1147$\pm$0.0034} & \textbf{0.1276$\pm$0.0023} & \textbf{0.1276$\pm$0.0160} \\
   &  & LSTM & 0.1913$\pm$0.0156 & 0.1868$\pm$0.0190 & 0.1993$\pm$0.0178 & 0.1887$\pm$0.0180 & 0.1767$\pm$0.0224 & 0.1886$\pm$0.0186 \\
    &  & ED-LSTM & 0.1815$\pm$0.0362 & 0.1955$\pm$0.0423 & 0.1862$\pm$0.0441 & 0.1881$\pm$0.0459 & 0.1913$\pm$0.0476 & 0.1885$\pm$0.0432 \\
    &  & BD-LSTM & 0.1033$\pm$0.0159 & 0.1788$\pm$0.0354 & 0.1799$\pm$0.0314 & 0.1741$\pm$0.0285 & 0.1960$\pm$0.0302 & 0.1664$\pm$0.0283 \\
   &  & Conv-LSTM & 0.0799$\pm$0.0243 & 0.1011$\pm$0.0247 & 0.1185$\pm$0.0422 & 0.1786$\pm$0.0526 & 0.1950$\pm$0.0517 & 0.1346$\pm$0.0391 \\
    &  & Transformer & 0.2464$\pm$0.0141 & 0.2485$\pm$0.0136 & 0.2590$\pm$0.0150 & 0.2508$\pm$0.0147 & 0.2470$\pm$0.0153 & 0.2503$\pm$0.0145 \\
    \hline
    & Multivariate* & CNN & 0.0202$\pm$0.0005 & 0.0228$\pm$0.0007 & 0.0243$\pm$0.0006 & 0.0266$\pm$0.0008 & 0.0282$\pm$0.0010 & 0.0244$\pm$0.0007 \\
   & & LSTM & 0.0168$\pm$0.0009 & 0.0217$\pm$0.0014 & 0.0241$\pm$0.0007 & 0.0274$\pm$0.0010 & 0.0311$\pm$0.0014 & 0.0242$\pm$0.0011 \\
   & & ED-LSTM & 0.0179$\pm$0.0010 & 0.0223$\pm$0.0013 & 0.0257$\pm$0.0011 & 0.0278$\pm$0.0013 & 0.0308$\pm$0.0013 & 0.0249$\pm$0.0012 \\
   & & BD-LSTM & 0.0171$\pm$0.0008 & 0.0210$\pm$0.00006 & 0.0246$\pm$0.0009 & 0.0278$\pm$0.0010 & 0.0300$\pm$0.0011 & 0.0241$\pm$0.0009 \\
   & & Conv-LSTM & \textbf{0.0158$\pm$0.0006} & \textbf{0.0193$\pm$0.0006} & \textbf{0.0212$\pm$0.0008} & \textbf{0.0231$\pm$0.0007} & \textbf{0.0255$\pm$0.0009} & \textbf{0.0210$\pm$0.0007} \\
   & & Transformer & 0.0186$\pm$0.0009 & 0.0201$\pm$0.0009 & 0.0220$\pm$0.0008 & 0.0239$\pm$0.0010 & 0.0262$\pm$0.0009 & 0.0222$\pm$0.0009 \\
   \hline
    \hline

    \end{tabular}
    \caption{Prediction accuracy on the test dataset   (RMSE mean and 95\% confidence interval ($\pm$)) for Bitcoin and Ethereum. The best models for the different steps are highlighted in bold, and the test mean provides the average of the five steps. Note that the Multivariate strategy only features Gold price and features (high, opening, low) to predict the close price of the given cryptocurrency (e.g. Bitcoin). In the Multivariate* strategy, there is an addition of the close price of the most correlated cryptocurrency (e.g. Bitcoin uses Ethereum as shown in Figure 13)   }
    \label{t511a}
\end{table*}

\begin{table*}[htbp!]
    \centering
    \small
    \begin{tabular}{ p{0.322in} p{0.48in} c c c c c c c }
    \hline 
    \hline
 Data &  Strategy  &Model&Step 1&Step 2&Step 3&Step 4&Step 5 & Test Mean \\
        \hline
  
        \hline
DOGE  &  Univariate & CNN & 0.1714$\pm$0.0490 & 0.2620$\pm$0.0831 & 0.2786$\pm$0.1056 & 0.4862$\pm$0.0956 & 0.5171$\pm$0.0968 & 0.3431$\pm$0.0860 \\
 &  & LSTM & 0.1492$\pm$0.0122 & 0.1523$\pm$0.0134 & 0.1577$\pm$0.0123 & 0.1637$\pm$0.0127 & 0.1681$\pm$0.0133 & 0.1582$\pm$0.0128 \\
    & & ED-LSTM & 0.1386$\pm$0.0136 & 0.1419$\pm$0.0129 & 0.1401$\pm$0.0127 & 0.1422$\pm$0.0121 & 0.1428$\pm$0.0118 & 0.1411$\pm$0.0126 \\
   & & BD-LSTM & \textbf{0.0509$\pm$0.0014} & \textbf{0.0562$\pm$0.0017} & \textbf{0.0722$\pm$0.0025} & \textbf{0.0648$\pm$0.0021} & \textbf{0.0645$\pm$0.0020} & \textbf{0.0617$\pm$0.0019} \\
  &    & Conv-LSTM & 0.0590$\pm$0.0198 & 0.1554$\pm$0.0913 & 0.1430$\pm$0.0914 & 0.1456$\pm$0.0893 & 0.1413$\pm$0.0830 & 0.1289$\pm$0.0750 \\
  &  & Transformer & 0.2116$\pm$0.0361 & 0.2228$\pm$0.0517 & 0.2435$\pm$0.0601 & 0.2219$\pm$0.0503 & 0.2188$\pm$0.0461 & 0.2237$\pm$0.0489 \\
  \hline
  &  Multivariate & CNN & 0.8122$\pm$0.0212 & 0.6364$\pm$0.0792 & 0.6725$\pm$0.0700 & 0.6656$\pm$0.0840 & 0.5972$\pm$0.0871 & 0.6768$\pm$0.0683 \\
    & & LSTM & 0.1706$\pm$0.0087 & 0.1746$\pm$0.0074 & 0.1828$\pm$0.0083 & 0.1810$\pm$0.0083 & 0.1825$\pm$0.0095 & 0.1783$\pm$0.0084 \\
    &  & ED-LSTM & 0.1829$\pm$0.0211 & 0.1823$\pm$0.0205 & 0.1816$\pm$0.0200 & 0.1821$\pm$0.0196 & 0.1823$\pm$0.0192 & 0.1822$\pm$0.0201 \\
  & & BD-LSTM & \textbf{0.0616$\pm$0.0065} & \textbf{0.0603$\pm$0.0051} & \textbf{0.0619$\pm$0.0039} & \textbf{0.0653$\pm$0.0032} & \textbf{0.0720$\pm$0.0042} & \textbf{0.0642$\pm$0.0046} \\
   &  & Conv-LSTM & 0.2103$\pm$0.0975 & 0.1962$\pm$0.0897 & 0.1912$\pm$0.0859 & 0.2347$\pm$0.1085 & 0.2160$\pm$0.1012 & 0.2097$\pm$0.0966 \\
    & & Transformer & 0.2472$\pm$0.0200 & 0.2482$\pm$0.0217 & 0.2501$\pm$0.0211 & 0.2345$\pm$0.0215 & 0.2332$\pm$0.0203 & 0.2426$\pm$0.0209 \\
    \hline

    & Multivariate* & CNN & 0.0220$\pm$0.0017 & 0.0257$\pm$0.0017 & 0.0301$\pm$0.0019 & 0.0326$\pm$0.0022 & 0.0330$\pm$0.0022 & 0.0287$\pm$0.0019 \\
   & & LSTM & 0.0194$\pm$0.0013 & 0.0238$\pm$0.0015 & 0.0260$\pm$0.0011 & 0.0307$\pm$0.0017 & 0.0323$\pm$0.0008 & 0.0264$\pm$0.0013 \\
   & & ED-LSTM & 0.0196$\pm$0.0020 & 0.0248$\pm$0.0022 & 0.0281$\pm$0.0023 & 0.0317$\pm$0.0025 & 0.0348$\pm$0.0021 & 0.0278$\pm$0.0022 \\
   & & BD-LSTM & 0.0183$\pm$0.0013 & 0.0234$\pm$0.0013 & 0.0272$\pm$0.0015 & 0.0317$\pm$0.0018 & 0.0350$\pm$0.0021 & 0.0271$\pm$0.0016 \\
   & & Conv-LSTM & \textbf{0.0168$\pm$0.0014} & \textbf{0.0173$\pm$0.0012} & \textbf{0.0179$\pm$0.0015} & \textbf{0.0200$\pm$0.0018} & \textbf{0.0230$\pm$0.0018} & \textbf{0.0190$\pm$0.0015} \\
   & & Transformer & 0.0208$\pm$0.0016 & 0.0212$\pm$0.0015 & 0.0237$\pm$0.0016 & 0.0271$\pm$0.0019 & 0.0304$\pm$0.0022 & 0.0246$\pm$0.0018 \\
        \hline
        \hline 
 LTC &    Univariate & CNN & 0.0390$\pm$0.0008 & 0.0470$\pm$0.0015 & 0.0627$\pm$0.0033 & 0.0948$\pm$0.0128 & 0.1131$\pm$0.0170 & 0.0713$\pm$0.0071 \\
    & & LSTM & 0.0382$\pm$0.0019 & 0.0479$\pm$0.0024 & 0.0619$\pm$0.0031 & 0.0768$\pm$0.0038 & 0.0861$\pm$0.0038 & 0.0622$\pm$0.0030 \\
   &  & ED-LSTM & 0.0369$\pm$0.0028 & 0.0487$\pm$0.0025 & 0.0631$\pm$0.0036 & 0.0756$\pm$0.0058 & 0.0838$\pm$0.0066 & 0.0616$\pm$0.0043 \\
    & & BD-LSTM & 0.0318$\pm$0.0019 & 0.0401$\pm$0.0019 & 0.0507$\pm$0.0029 & 0.0588$\pm$0.0039 & 0.0699$\pm$0.0034 & 0.0503$\pm$0.0028 \\
    & & Conv-LSTM & \textbf{0.0265$\pm$0.0012} & \textbf{0.0362$\pm$0.0017} & \textbf{0.0443$\pm$0.0015} & \textbf{0.0513$\pm$0.0019} & \textbf{0.0581$\pm$0.0017} & \textbf{0.0433$\pm$0.0016} \\
  &  & Transformer & 0.0726$\pm$0.0035 & 0.0712$\pm$0.0041 & 0.0746$\pm$0.0046 & 0.0818$\pm$0.0038 & 0.0891$\pm$0.0032 & 0.0779$\pm$0.0038 \\
  \hline
    & Multivariate & CNN & 0.4648$\pm$0.0468 & 0.4553$\pm$0.0444 & 0.4810$\pm$0.0445 & 0.5086$\pm$0.0460 & 0.5167$\pm$0.0492 & 0.4853$\pm$0.0462 \\
    & & LSTM & 0.0625$\pm$0.0052 & 0.0788$\pm$0.0070 & \textbf{0.0921$\pm$0.0058} & 0.0885$\pm$0.0044 & 0.0919$\pm$0.0044 & \textbf{0.0828$\pm$0.0054} \\
    &  & ED-LSTM & 0.0940$\pm$0.0130 & 0.1121$\pm$0.0125 & 0.1032$\pm$0.0116 & \textbf{0.0833$\pm$0.0036} & \textbf{0.0880$\pm$0.0037} & 0.0961$\pm$0.0089 \\
 &  & BD-LSTM & 0.0859$\pm$0.0111 & 0.1520$\pm$0.0268 & 0.2513$\pm$0.0441 & 0.2838$\pm$0.0523 & 0.2760$\pm$0.0501 & 0.2098$\pm$0.0369 \\
    &  & Conv-LSTM & \textbf{0.0542$\pm$0.0054} & \textbf{0.0713$\pm$0.0094} & 0.0952$\pm$0.0126 & 0.1028$\pm$0.0136 & 0.1123$\pm$0.0127 & 0.0872$\pm$0.0107 \\
   &  & Transformer & 0.1532$\pm$0.0131 & 0.1624$\pm$0.0108 & 0.1724$\pm$0.0083 & 0.1671$\pm$0.0129 & 0.1650$\pm$0.0128 & 0.1640$\pm$0.0116 \\
   \hline
     & Multivariate* & CNN & 0.0277$\pm$0.0011 & 0.0304$\pm$0.0012 & 0.0342$\pm$0.0012 & 0.0371$\pm$0.0011 & 0.0409$\pm$0.0011 & 0.0341$\pm$0.0011 \\
   & & LSTM & 0.0233$\pm$0.0015 & 0.0295$\pm$0.0016 & 0.0340$\pm$0.0016 & 0.0380$\pm$0.0017 & 0.0416$\pm$0.0017 & 0.0333$\pm$0.0016 \\
   & & ED-LSTM & 0.0230$\pm$0.0012 & 0.0300$\pm$0.0015 & 0.0350$\pm$0.0016 & 0.0395$\pm$0.0014 & 0.0433$\pm$0.0016 & 0.0342$\pm$0.0015 \\
   & & BD-LSTM & 0.0240$\pm$0.0013 & 0.0306$\pm$0.0010 & 0.0355$\pm$0.0010 & 0.0395$\pm$0.0013 & 0.0432$\pm$0.0013 & 0.0346$\pm$0.0012 \\
   & & Conv-LSTM & \textbf{0.0223$\pm$0.0008} & \textbf{0.0270$\pm$0.0009} & 0.0307$\pm$0.0012 & \textbf{0.0332$\pm$0.0013} & \textbf{0.0362$\pm$0.0011} & \textbf{0.0300$\pm$0.0011} \\
   & & Transformer & 0.0253$\pm$0.0013 & 0.0276$\pm$0.0012 & \textbf{0.0303$\pm$0.0008} & 0.0336$\pm$0.0010 & 0.0374$\pm$0.0014 & 0.0308$\pm$0.0011 \\
   
    \hline \hline
    \end{tabular}
    \caption{Prediction accuracy for the test dataset   (RMSE mean and 95\% confidence interval ($\pm$)) for Dodgecoin and Litecoin. The best models for the different steps are highlighted in bold, and the test mean provides the average of the five steps. Note that the Multivariate strategy only features Gold price and features (high, opening, low) to predict the close price of the given cryptocurrency (e.g. Dogecoin). In the Multivariate* strategy, there is an addition of the close price of the most correlated cryptocurrency (e.g. Dodgecoin uses Ethereum as shown in Figure 13).  }
    \label{t511b}
\end{table*}

\begin{figure*}[htbp!]
\subfloat[Bitcoin]{
    \label{fnew1}
    \includegraphics[width=8.4cm]{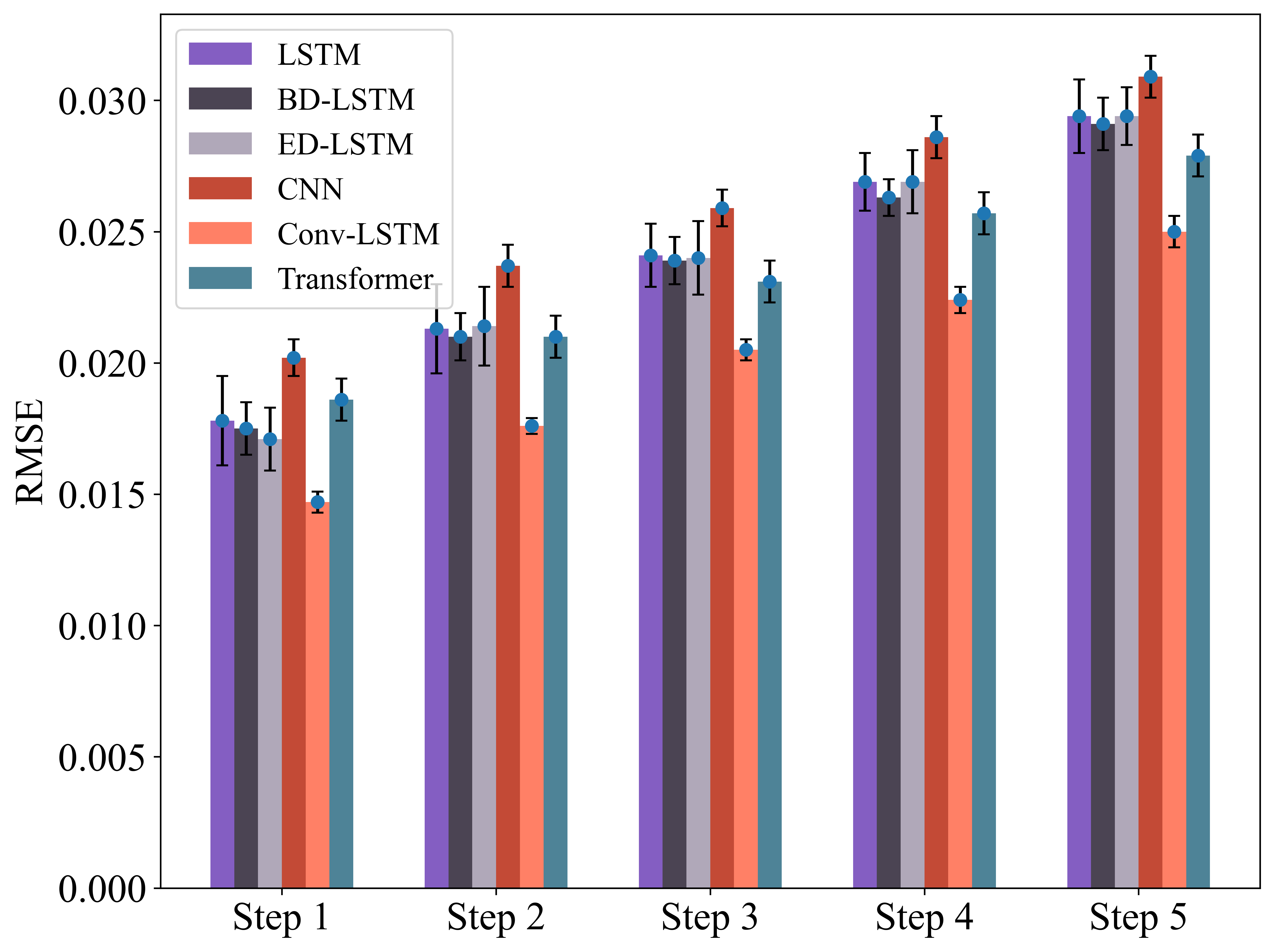}
}
\subfloat[Ethereum]{
    \label{fnew2}
    \includegraphics[width=8.4cm]{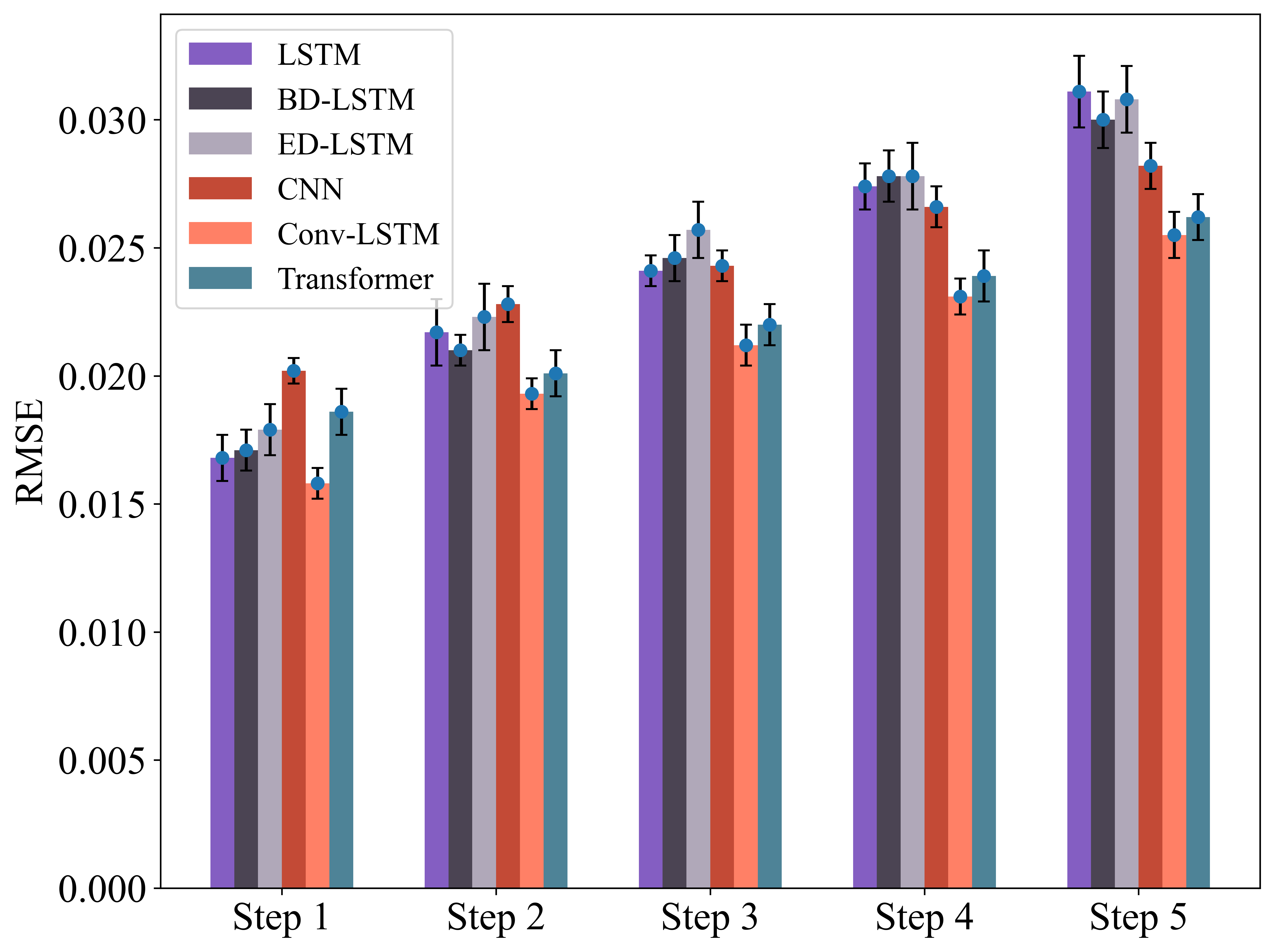}
} 
\\
\subfloat[Dogecoin]{
    \label{fnew3}
    \includegraphics[width=8.4cm]{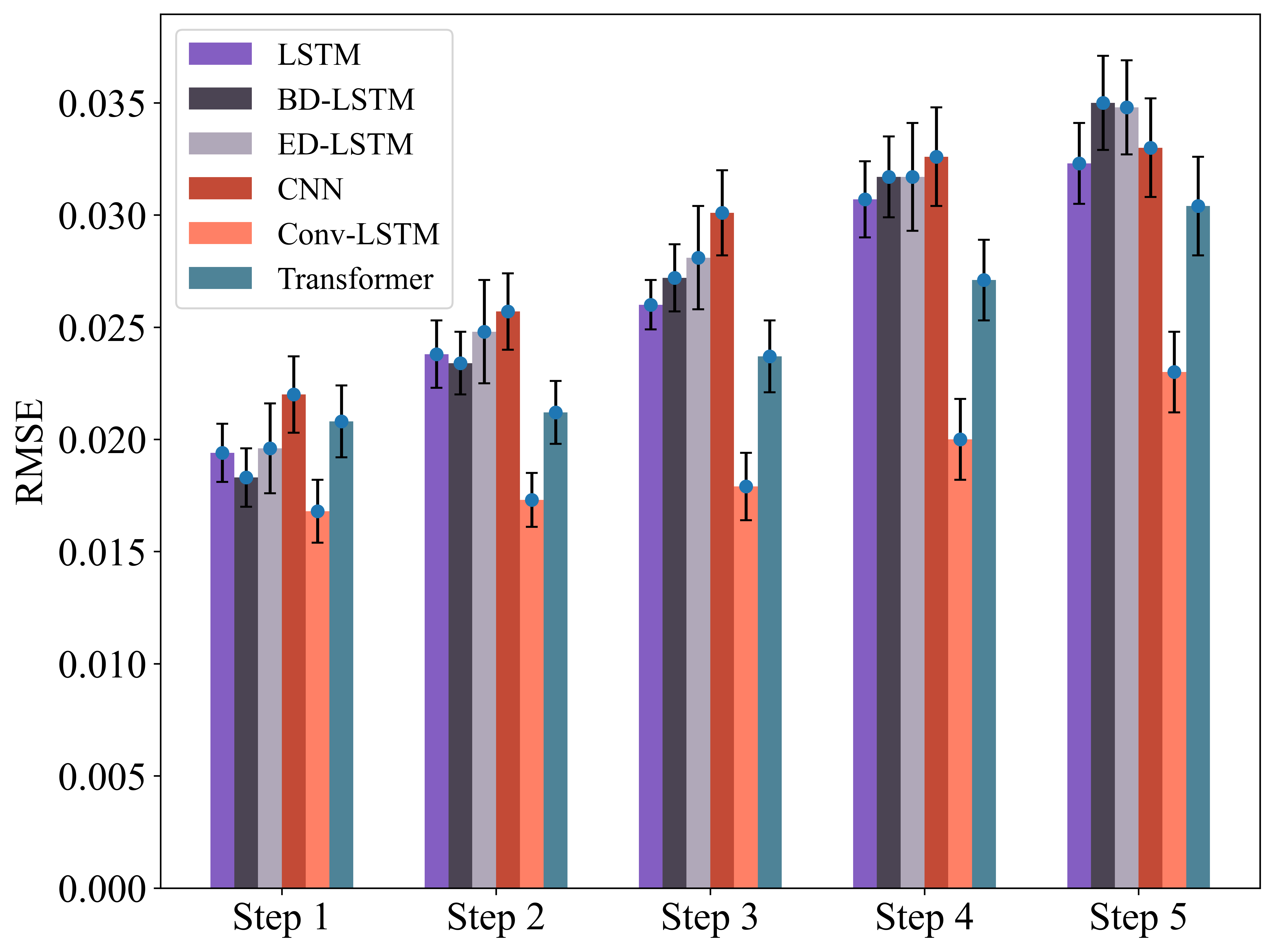}
}
\subfloat[Litecoin]{
    \label{fnew4}
    \includegraphics[width=8.4cm]{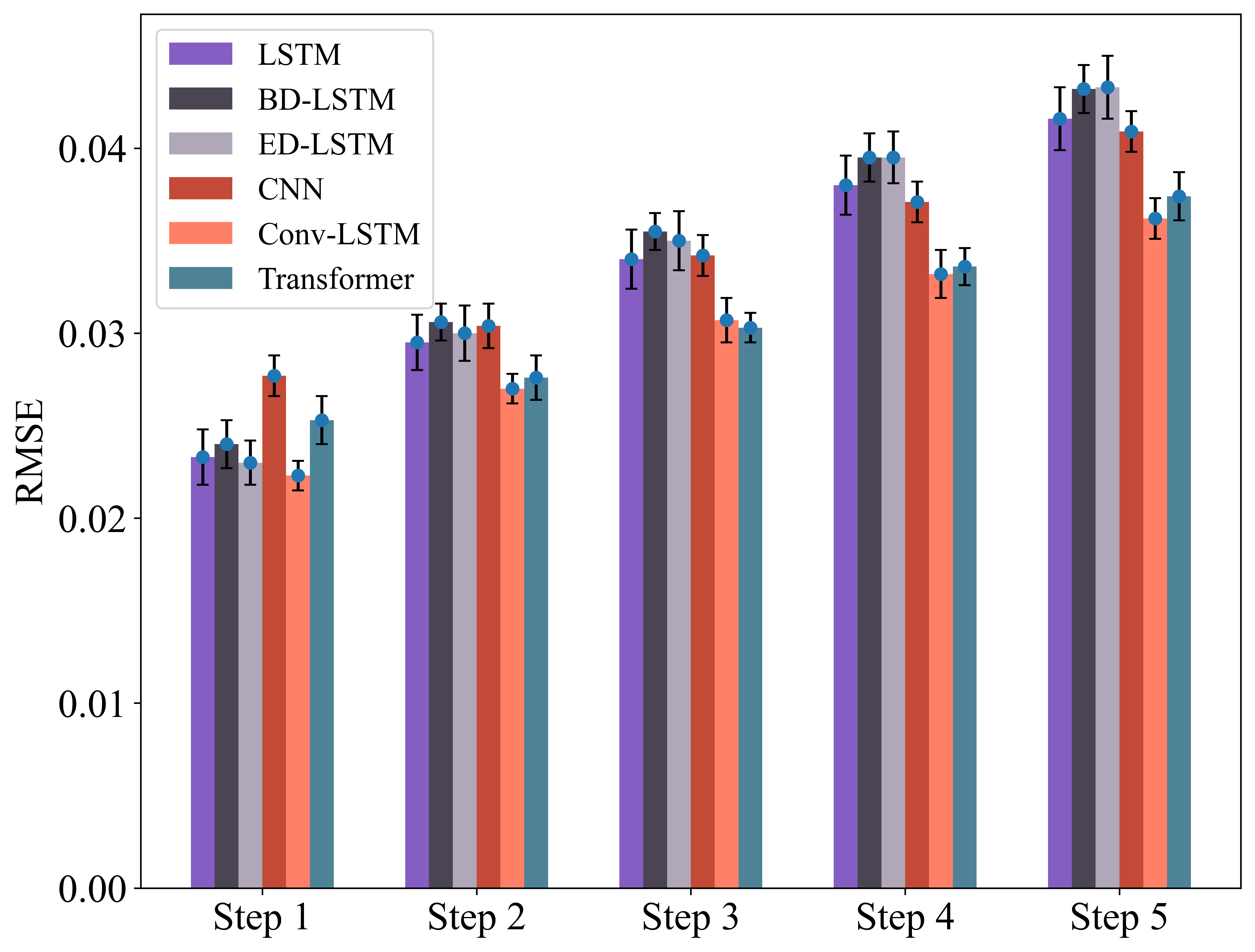}
}
\caption{Performance evaluation of respective Multivariate* methods (RMSE mean with 95\% CI for 30 experimental runs.) for Experiment 1: pre COVID-19.}
\label{fmultinew1}
\end{figure*}

 Furthermore, we present another approach (Multivariate*)  which features the Gold price and a highly correlated cryptocurrency as shown in Figure 13, i.e. in the case of Bitcoin, the Multivariate* model  features the Ethereum and Gold price along with Bitcoin high, low, opening and closing price.
 Figure \ref{fmultinew1} presents the results where we observe that the forecasting results improve significantly given the additional feature of a highly correlated  cryptocurrency, when compared to previous Multivariate results for the respective cryptocurrencies (Figure 15, 17, 19, and 21). In  Figure \ref{fmultinew1}, the prediction accuracy of all the respective models  exhibit a consistent decrease in accuracy (higher RMSE) with increasing prediction horizon. The prediction results for Bitcoin shows that Conv-LSTM is the best performing model in Figure \ref{fnew1} and CNN is the worst model. Figure \ref{fnew2} shows the results of each model predicting Ethereum where Conv-LSTM  maintains the best prediction accuracy. It is worth to mention that the ranking of LSTM among the 6 models gradually deteriorates as the prediction horizon increases. The prediction results of the models for Dogecoin are shown in Figure \ref{fnew3} where  Conv-LSTM has the best prediction performance in this case, followed by the Transformer model. Figure \ref{fnew4} presents the results of the Multivariate* for Litecoin   where we find that Conv-LSTM shows the best prediction accuracy.

 We summarise the results further in Tables \ref{t511a} and \ref{t511b} which feature the model prediction accuracy of the test dataset. We report the RMSE mean and 95\% confidence interval for the four cryptocurrencies, and the best models for the different steps are highlighted in bold. The test mean provides the average of the five steps. It is clear that the Univariate models are better than the Multivariate models; however, we find that the accuracy of both strategies is close (test mean) for Bitcoin and Dogecoin.  In general, Multivariate* strategy provides the best results for the four cryptocurrencies. 
 
 Table \ref{trank1} provides a rank of the models where BD-LSTM provides the best accuracy rank followed by  Conv-LSTM  for the case of the Univariate approach. Furthermore, Conv-LSTM provides the best accuracy rank followed by the LSTM and ED-LSTM models for the Multivariate approach. Finally, Conv-LSTM provides the best accuracy rank followed by the Transformer model for the Multivariate* approach.

\subsection{Results: Data featuring COVID-19}

\begin{figure*}[h]
\subfloat[Bitcoin]{
    \label{fnew5}
    \includegraphics[width=8.4cm]{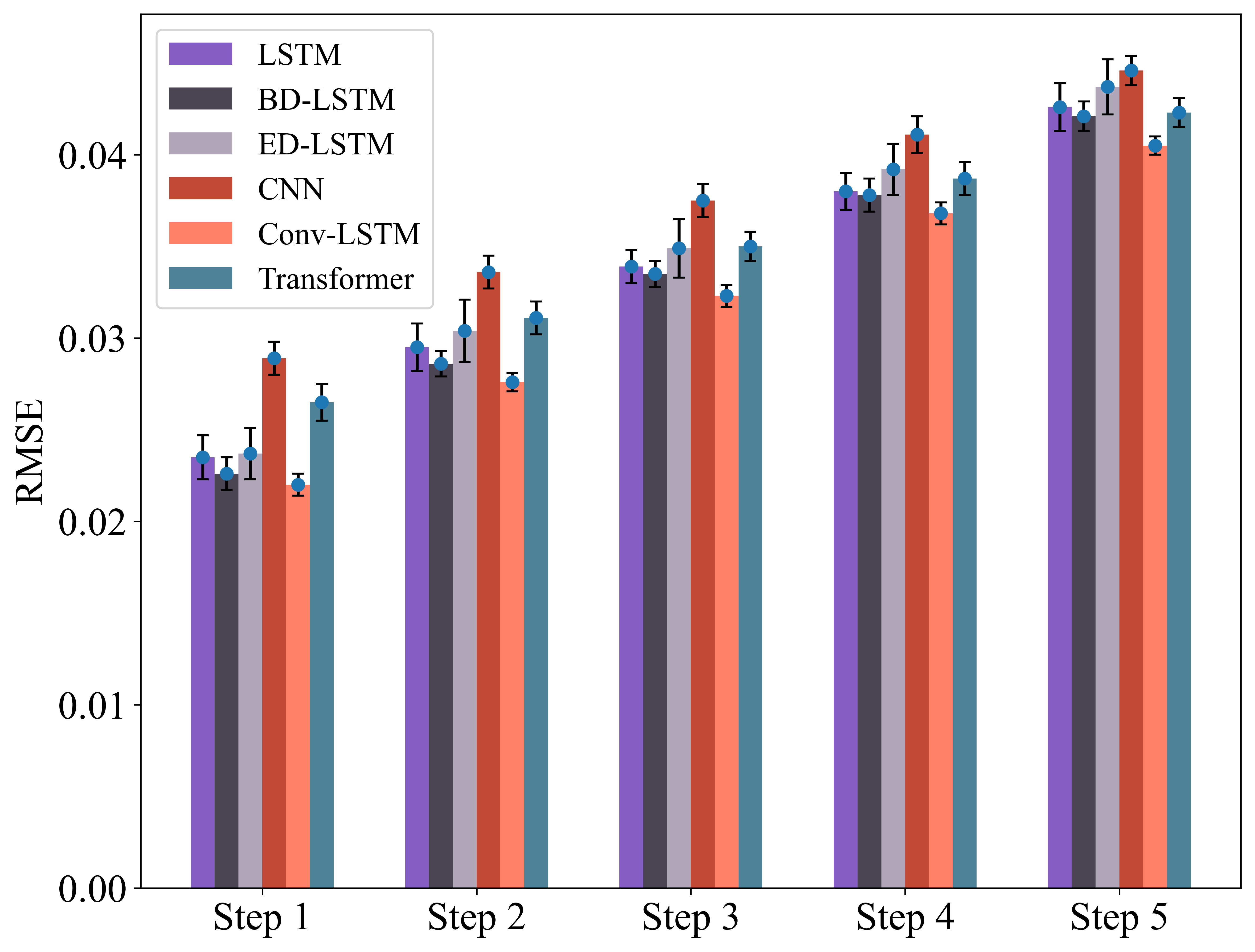}
}
\subfloat[Ethereum]{
    \label{fnew6}
    \includegraphics[width=8.4cm]{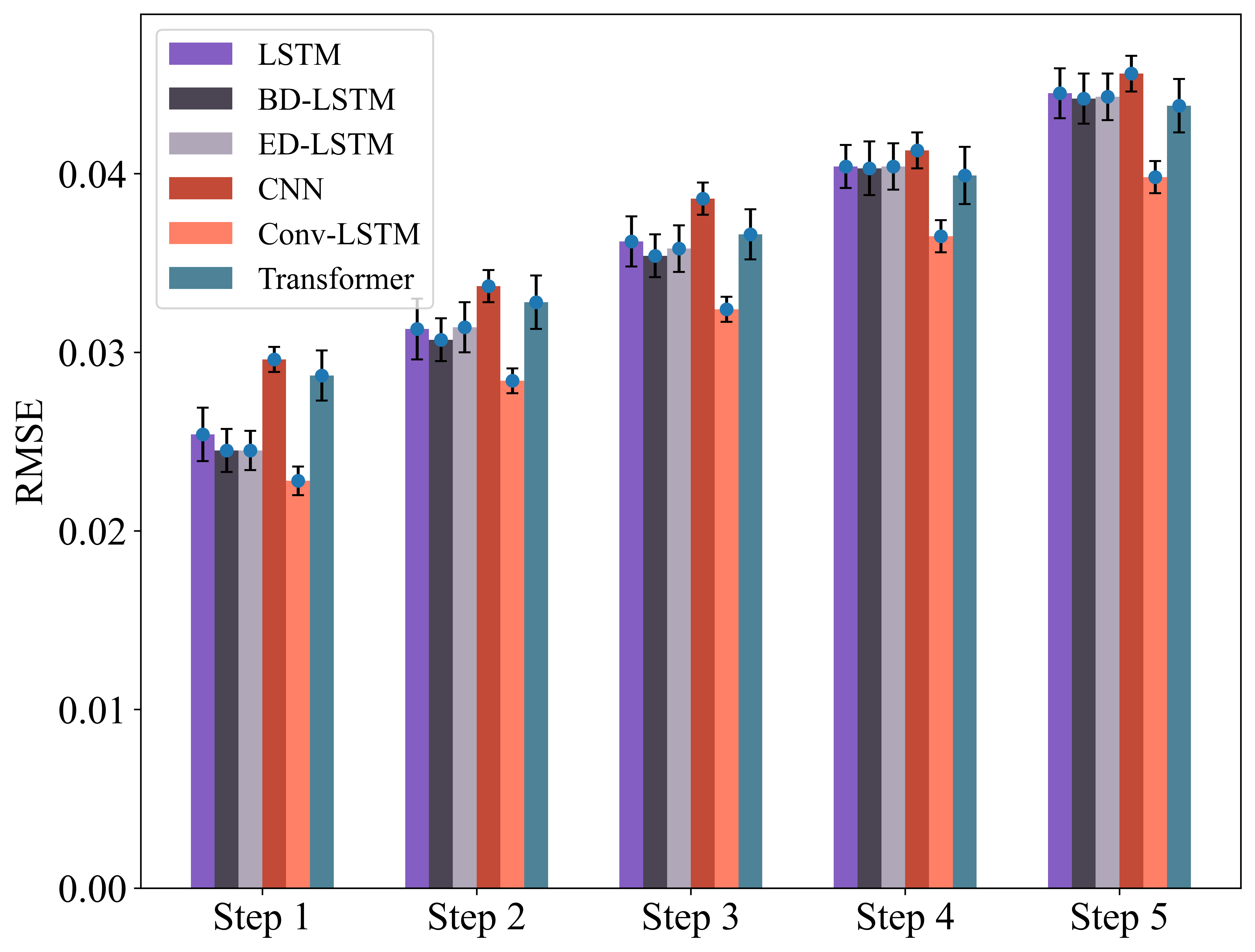}
} 
\\
\subfloat[Dogecoin]{
    \label{fnew7}
    \includegraphics[width=8.4cm]{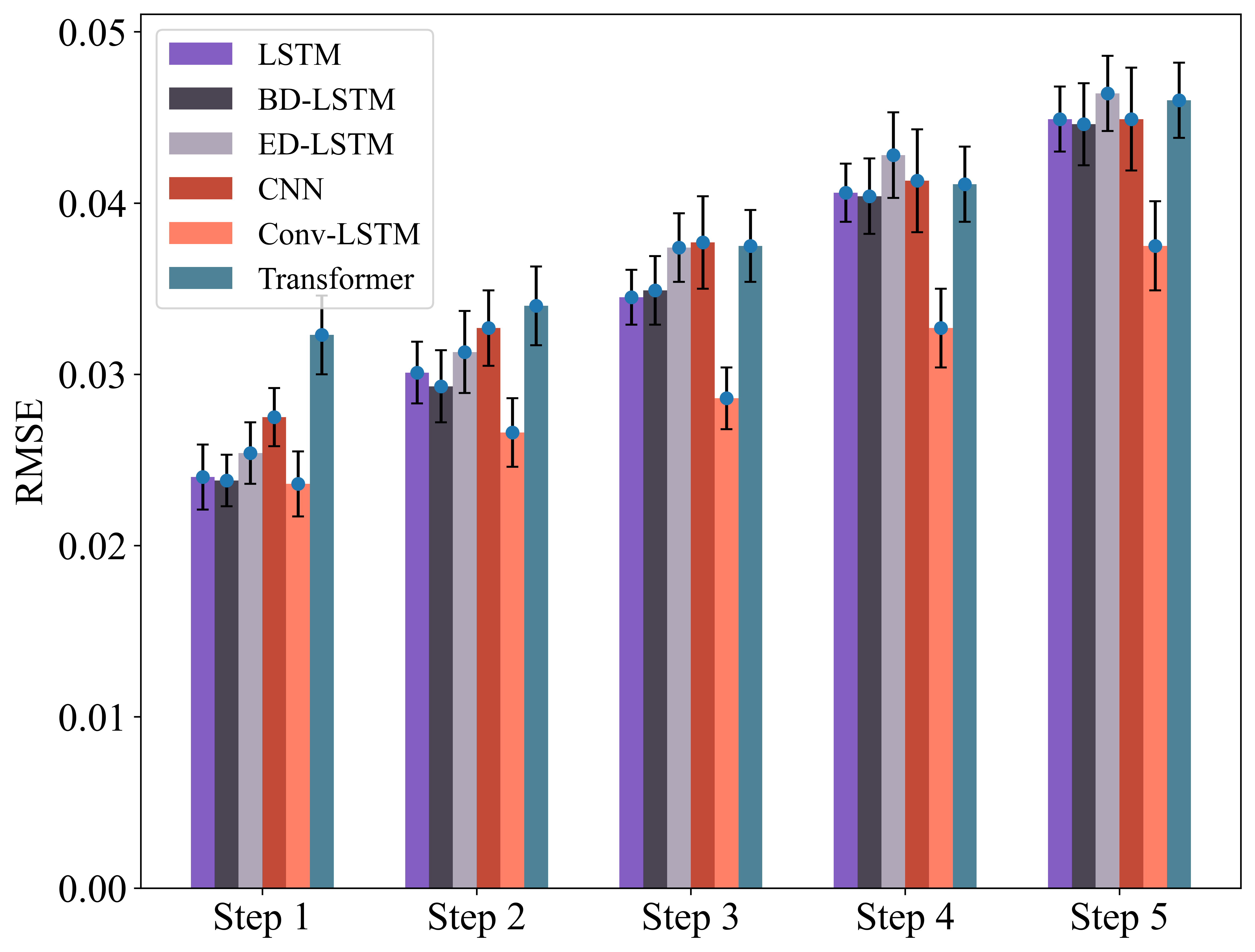}
}
\subfloat[Litecoin]{
    \label{fnew8}
    \includegraphics[width=8.4cm]{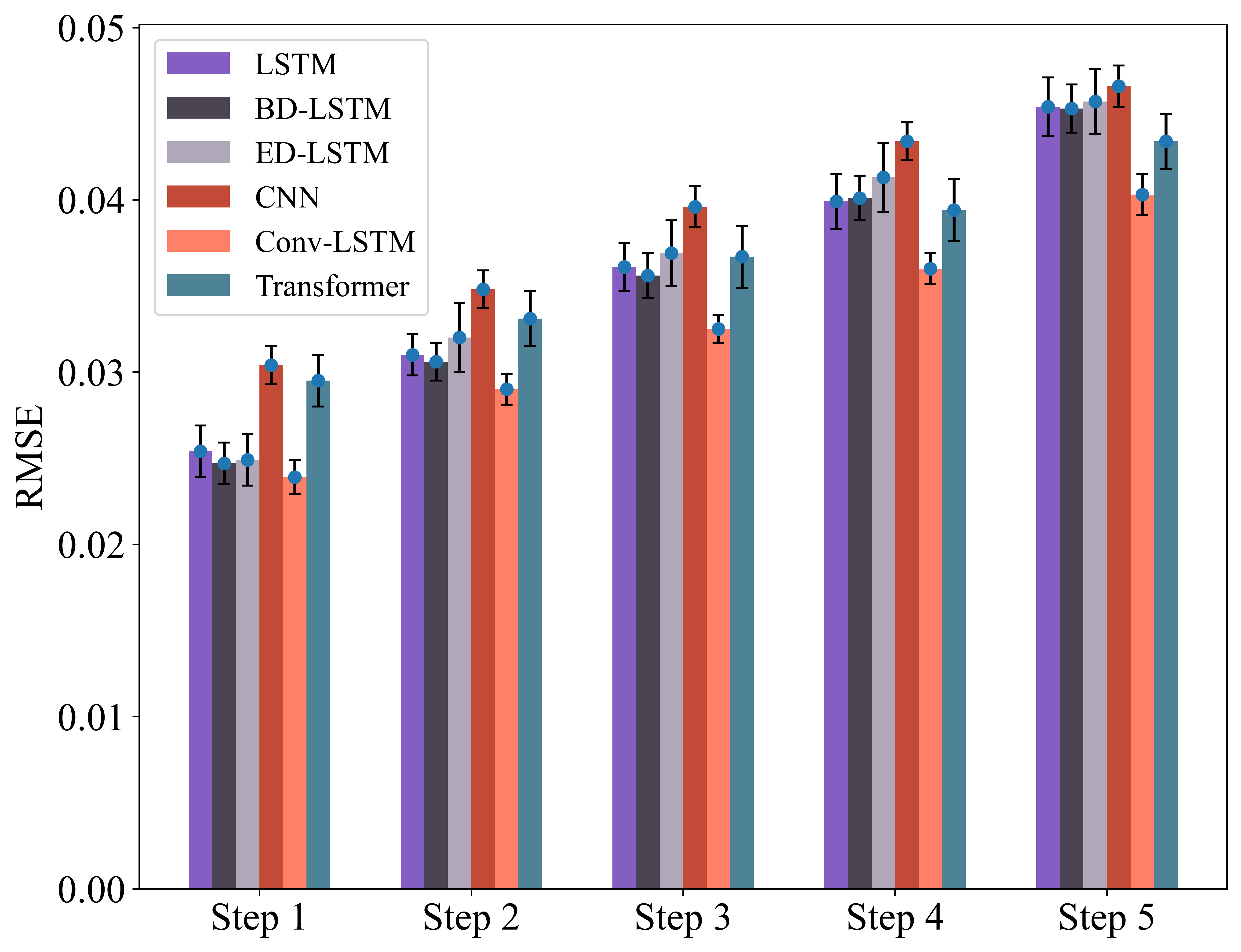}
}
\caption{Performance evaluation of respective Multivariate* methods (RMSE mean with 95\% CI for 30 experimental runs.) for Experiment 2: data featuring  COVID-19.}
\label{fmultinew2}
\end{figure*}

The previous section presented results given by the respective models using data before COVID-19. We found that the   Multivariate* strategy provided the best performance (Table \ref{t511a} and \ref{t511b}) in comparison to Multivariate and Univariate strategies. Therefore, in Experiment 2, i.e. data featuring COVID-19,  we only used the Univariate and Multivariate* strategies and presented the results in Table \ref{t461}. We further present the results for predicting the cryptocurrency at each prediction horizon in Figure \ref{fmultinew2}. We find that Conv-LSTM is the best model under the Multivariate* strategy. Figure \ref{fnew5} displays the performance of 6 deep learning models in forecasting Bitcoin. As the number of prediction days increases, the   forecast accuracy gradually decreases (higher RMSE). The models predict that the result of Ethereum (in Figure \ref{fnew6}) is similar to that of Bitcoin. Figure \ref{fnew7} demonstrates a decline in the model's robustness when making predictions for Dogecoin. The 95\% confidence interval of the model is larger than the other three cryptocurrencies. In terms of prediction accuracy, Dogecoin is almost the same as other cryptocurrencies. We also present the forecast results of Litecoin in Figure \ref{fnew8}, although the outcomes   remain consistent with Bitcoin, Ethereum and Dogecoin.

  We find that comparing the results of Experiment 2 with Experiment 1 (pre-COVID-19 data), the prediction accuracy of Bitcoin, Ethereum, and Dogecoin has generally deteriorated to a certain extent. Let us examine the case of Conv-LSTM with Multivariate*  which has provided the best results for Experiment 1 (Tables 10 and 11) and Experiment 2 (Table 12). Looking at the test mean, we get RMSE around 0.02 for Bitcoin and Ethereum in Experiment 1, but in Experiment 2, we get RMSE of around 0.03. Furthermore, in  Experiment 1 Multivariate* Conv-LSTM, we find that Dogecoin and Litecoin obtained RMSE of 0.019 and 0.03, respectively; in Experiment 2, these deteriorated to 0.029 and 0.032, respectively. This could be due to the high vitality in the dataset featuring COVID-19 as the markets were adjusting due to lockdowns and economic crisis around the world.  

 Table \ref{trank2} provides a rank of the models where BD-LSTM provides the best accuracy rank followed by  ED-LSTM  for the  Univariate strategy. Furthermore, Conv-LSTM provides the best accuracy rank followed by the BD-LSTM models for the Multivariate* approach.

\begin{table*}[htbp!]
    \centering
    \small
    \begin{tabular}{p{0.322in} p{0.48in} c c c c c c c}
    \hline \hline
 Data  & Strategy &Model&Step 1&Step 2&Step 3&Step 4&Step 5 & Test Mean \\
        \hline
  
        \hline
BTC  & Univariate& BD-LSTM & \textbf{0.0194$\pm$0.0002} & \textbf{0.0258$\pm$0.0003} & \textbf{0.0311$\pm$0.0004} & \textbf{0.0367$\pm$0.0003} & \textbf{0.0414$\pm$0.0004} & \textbf{0.0309$\pm$0.0003} \\
  & & ED-LSTM & 0.0199$\pm$0.0001 & 0.0284$\pm$0.0004 & 0.0339$\pm$0.0006 & 0.0381$\pm$0.0005 & 0.0418$\pm$0.0003 & 0.0324$\pm$0.0004\\
  & & LSTM & 0.0283$\pm$0.0004 & 0.0326$\pm$0.0003 & 0.0369$\pm$0.0003 & 0.0410$\pm$0.0003 & 0.0447$\pm$0.0002& 0.0367$\pm$0.0003 \\
  & & CNN & 0.0293$\pm$0.0006 & 0.0342$\pm$0.0006 & 0.0388$\pm$0.0005 & 0.0431$\pm$0.0004 & 0.0467$\pm$0.0003 & 0.0384$\pm$0.0005 \\
  & & Conv-LSTM & 0.0209$\pm$0.0004 & 0.0263$\pm$0.0004 & 0.0317$\pm$0.0004 & 0.0372$\pm$0.0003 & 0.0421$\pm$0.0002 & 0.0316$\pm$0.0004 \\
  & & Transformer & 0.0484$\pm$0.0055 & 0.0514$\pm$0.0051 & 0.0546$\pm$0.0047 & 0.0585$\pm$0.0047 & 0.0609$\pm$0.0047 & 0.0548$\pm$0.0050 \\
    \hline
  & Multivariate* & BD-LSTM & 0.0226$\pm$0.0009 & 0.0286$\pm$0.0008 & 0.0335$\pm$0.0007 & 0.0378$\pm$0.0009 & 0.0421$\pm$0.0008 & 0.0329$\pm$0.0008 \\
  & & ED-LSTM & 0.0237$\pm$0.0014 & 0.0304$\pm$0.0017 & 0.0349$\pm$0.0016 & 0.0392$\pm$0.0014 & 0.0437$\pm$0.0015 & 0.0344$\pm$0.0015 \\
  & & LSTM & 0.0235$\pm$0.0012 & 0.0295$\pm$0.0013 & 0.0339$\pm$0.0010 & 0.0380$\pm$0.0010 & 0.0426$\pm$0.0013 & 0.0335$\pm$0.0012 \\
  & & CNN & 0.0289$\pm$0.0010 & 0.0336$\pm$0.0009 & 0.0375$\pm$0.0009 & 0.0411$\pm$0.0010 & 0.0446$\pm$0.0008 & 0.0371$\pm$0.0009 \\
  & & Conv-LSTM & \textbf{0.0220$\pm$0.0007} & \textbf{0.0276$\pm$0.0006} & \textbf{0.0323$\pm$0.0006} & \textbf{0.0368$\pm$0.0007} & \textbf{0.0405$\pm$0.0006} & \textbf{0.0318$\pm$0.0006} \\
  & & Transformer & 0.0265$\pm$0.0010 & 0.0311$\pm$0.0009 & 0.0350$\pm$0.0008 & 0.0387$\pm$0.0010 & 0.0423$\pm$0.0008 & 0.0347$\pm$0.0009 \\
  \hline \hline
ETH & Univariate & BD-LSTM & 0.0230$\pm$0.0002 & \textbf{0.0304$\pm$0.0004} & \textbf{0.0361$\pm$0.0005} & \textbf{0.0426$\pm$0.0004} & 0.0484$\pm$0.0006 & \textbf{0.0361$\pm$0.0004} \\
  & & ED-LSTM & \textbf{0.0230$\pm$0.0001} & 0.0307$\pm$0.0004 & 0.0362$\pm$0.0004 & 0.0426$\pm$0.0004 & 0.0481$\pm$0.0006 & \textbf{ 0.0361$\pm$0.0004} \\
  & & LSTM & 0.0238$\pm$0.0005 & 0.0306$\pm$0.0005 & 0.0361$\pm$0.0005 & 0.0426$\pm$0.0005 & \textbf{0.0480$\pm$0.0005} & 0.0362$\pm$0.0005 \\
  & & CNN & 0.0321$\pm$0.0004 & 0.0397$\pm$0.0008 & 0.0481$\pm$0.0014 & 0.0536$\pm$0.0016 & 0.0585$\pm$0.0015 & 0.0464$\pm$0.0012 \\
  & & Conv-LSTM & 0.0250$\pm$0.0004 & 0.0332$\pm$0.0007 & 0.0405$\pm$0.0013 & 0.0481$\pm$0.0020 & 0.0550$\pm$0.0025 & 0.0404$\pm$0.0014 \\
  & & Transformer & 0.0269$\pm$0.0008 & 0.0359$\pm$0.0010 & 0.0411$\pm$0.0011 & 0.0486$\pm$0.0017 & 0.0542$\pm$0.0019 & 0.0413$\pm$0.0013 \\
  \hline
  & Multivariate* & BD-LSTM & 0.0245$\pm$0.0011 & 0.0307$\pm$0.0012 & 0.0354$\pm$0.0012 & 0.0403$\pm$0.0015 & 0.0442$\pm$0.0014 & 0.0350$\pm$0.0013 \\
  & & ED-LSTM & 0.0245$\pm$0.0011 & 0.0314$\pm$0.0014 & 0.0358$\pm$0.0014 & 0.0404$\pm$0.0013 & 0.0443$\pm$0.0013 & 0.0353$\pm$0.0013 \\
  & & LSTM & 0.0254$\pm$0.0015 & 0.0313$\pm$0.0017 & 0.0362$\pm$0.0014 & 0.0404$\pm$0.0012 & 0.0445$\pm$0.0014 & 0.0356$\pm$0.0014 \\
  & & CNN & 0.0296$\pm$0.0007 & 0.0337$\pm$0.0008 & 0.0386$\pm$0.0009 & 0.0413$\pm$0.0010 & 0.0456$\pm$0.0010 & 0.0378$\pm$0.0009 \\
  & & Conv-LSTM & \textbf{0.0228$\pm$0.0008} & \textbf{0.0284$\pm$0.0007} & \textbf{0.0324$\pm$0.0008} & \textbf{0.0365$\pm$0.0010} & \textbf{0.0398$\pm$0.0009} & \textbf{0.0320$\pm$0.0008} \\
  & & Transformer & 0.0287$\pm$0.0014 & 0.0328$\pm$0.0015 & 0.0366$\pm$0.0.014 & 0.0399$\pm$0.0016 & 0.0438$\pm$0.0015 & 0.0364$\pm$0.0015 \\
  \hline \hline
 DOGE& Univariate & BD-LSTM & 0.0291$\pm$0.0001 & 0.0618$\pm$0.0041 & 0.0673$\pm$0.0046 & 0.0742$\pm$0.0044 & 0.0795$\pm$0.0041 & \textbf{0.0624$\pm$0.0035} \\
 & & ED-LSTM & \textbf{0.0290$\pm$0.0001} & 0.0626$\pm$0.0030 & 0.0656$\pm$0.0029 & 0.0708$\pm$0.0026 & 0.0757$\pm$0.0024 & 0.0607$\pm$0.0022 \\
 & & LSTM & 0.0660$\pm$0.0039 & 0.0683$\pm$0.0042 & 0.0728$\pm$0.0047 & 0.0792$\pm$0.0048 & 0.0835$\pm$0.0043 & 0.0740$\pm$0.0044 \\
 & & CNN & 0.0646$\pm$0.0041 & 0.0630$\pm$0.0040 & 0.0655$\pm$0.0039 & 0.0762$\pm$0.0039 & 0.0806$\pm$0.0031 & 0.0700$\pm$0.0038 \\
 & & Conv-LSTM & 0.0538$\pm$0.0021 & 0.0593$\pm$0.0022 & \textbf{0.0612$\pm$0.0020} & \textbf{0.0662$\pm$0.0020} & \textbf{0.0726$\pm$0.0019} & 0.0626$\pm$0.0020 \\
 & & Transformer & 0.0536$\pm$0.0071 & \textbf{0.0586$\pm$0.0077} & 0.0639$\pm$0.0073 & 0.0724$\pm$0.0070 & 0.0789$\pm$0.0064 & 0.0655$\pm$0.0071 \\
  \hline
  & Multivariate* & BD-LSTM & 0.0238$\pm$0.0015 & 0.0293$\pm$0.00021 & 0.0349$\pm$0.0021 & 0.0404$\pm$0.0022 & 0.0446$\pm$0.0024 & 0.0346$\pm$0.0021 \\
  & & ED-LSTM & 0.0254$\pm$0.0018 & 0.0313$\pm$0.0024 & 0.0374$\pm$0.0020 & 0.0428$\pm$0.0025 & 0.0464$\pm$0.0022 & 0.0367$\pm$0.0022 \\
  & & LSTM & 0.0240$\pm$0.0020 & 0.0301$\pm$0.0018 & 0.0345$\pm$0.0016 & 0.0406$\pm$0.0017 & 0.0449$\pm$0.0019 & 0.0348$\pm$0.0018 \\
  & & CNN & 0.0275$\pm$0.0017 & 0.0327$\pm$0.0022 & 0.0377$\pm$0.0017 & 0.0413$\pm$0.0020 & 0.0449$\pm$0.0030 & 0.0368$\pm$0.0021 \\
  & & Conv-LSTM & \textbf{0.0236$\pm$0.0020} & \textbf{0.0266$\pm$0.0020} & \textbf{0.0286$\pm$0.0019} & \textbf{0.0327$\pm$0.0023} & \textbf{0.0375$\pm$0.0026} & \textbf{0.0298$\pm$0.0022} \\
  & & Transformer & 0.0323$\pm$0.0023 & 0.0340$\pm$0.0023 & 0.0375$\pm$0.0022 & 0.0411$\pm$0.0022 & 0.0460$\pm$0.0023 & 0.0382$\pm$0.0023 \\
  \hline \hline
 LTC & Univariate &  BD-LSTM & \textbf{0.0577$\pm$0.0007} & 0.0797$\pm$0.0022 & 0.0968$\pm$0.0018 & 0.1096$\pm$0.0015 & 0.1205$\pm$0.0016 & 0.0929$\pm$0.0016 \\
 & & ED-LSTM & 0.0578$\pm$0.0006 & \textbf{0.0797$\pm$0.0012} & \textbf{0.0962$\pm$0.0010} & \textbf{0.1096$\pm$0.0009} & \textbf{0.1198$\pm$0.0009} & \textbf{0.0926$\pm$0.0009} \\
 & & LSTM & 0.0587$\pm$0.0021 & 0.0804$\pm$0.0017 & 0.0971$\pm$0.0015 & 0.1100$\pm$0.0014 & 0.1207$\pm$0.0013 & 0.0934$\pm$0.0016 \\
 & & CNN & 0.0823$\pm$0.0011 & 0.1060$\pm$0.0042 & 0.1163$\pm$0.0025 & 0.1268$\pm$0.0029 & 0.1403$\pm$0.0049 & 0.1143$\pm$0.0031 \\
 & & Conv-LSTM & 0.0809$\pm$0.0073 & 0.1043$\pm$0.0095 & 0.1201$\pm$0.0087 & 0.1286$\pm$0.0086 & 0.1383$\pm$0.0072 & 0.1145$\pm$0.0083 \\
 & & Transformer & 0.0890$\pm$0.0056 & 0.1046$\pm$0.0048 & 0.1163$\pm$0.0042 & 0.1273$\pm$0.0038 & 0.1354$\pm$0.0033 & 0.1146$\pm$0.0043 \\
   \hline
  & Multivariate* & BD-LSTM & 0.0247$\pm$0.0012 & 0.0306$\pm$0.0010 & 0.0356$\pm$0.0013 & 0.0401$\pm$0.0013 & 0.0453$\pm$0.0014 & 0.0352$\pm$0.0012 \\
  & & ED-LSTM & 0.0249$\pm$0.0015 & 0.0320$\pm$0.0020 & 0.0369$\pm$0.0019 & 0.0413$\pm$0.0020 & 0.0457$\pm$0.0019 & 0.0362$\pm$0.0019 \\
  & & LSTM & 0.0254$\pm$0.0014 & 0.0310$\pm$0.0012 & 0.0361$\pm$0.0014 & 0.0399$\pm$0.0016 & 0.0454$\pm$0.0018 & 0.0357$\pm$0.0015 \\
  & & CNN & 0.0304$\pm$0.0010 & 0.0348$\pm$0.0011 & 0.0396$\pm$0.0012 & 0.0434$\pm$0.0011 & 0.0466$\pm$0.0012 & 0.0390$\pm$0.0011 \\
  & & Conv-LSTM & \textbf{0.0239$\pm$0.0009} & \textbf{0.0290$\pm$0.0010} & \textbf{0.0325$\pm$0.0008} & \textbf{0.0360$\pm$0.0010} & \textbf{0.0403$\pm$0.0012} & \textbf{0.0323$\pm$0.0010} \\
  & & Transformer & 0.0295$\pm$0.0015 & 0.0331$\pm$0.0016 & 0.0367$\pm$0.0018 & 0.0394$\pm$0.0019 & 0.0434$\pm$0.0017 & 0.0364$\pm$0.0017 \\
    \hline \hline
    \end{tabular}
    \caption{Prediction accuracy on the test dataset  reporting RMSE mean and 95\% confidence interval ($\pm$) for the four cryptocurrencies during COVID-19. }
    \label{t461}
\end{table*}

\begin{table*}[htbp!]
    \centering
    \small
    \begin{tabular}{c c c c c c c c}
    \hline \hline
    Data & Strategy & LSTM & ED-LSTM & BD-LSTM & CNN & Conv-LSTM &Transformer \\
    \hline
    BTC& \multirow{4}{*}{Univariate} &4&3&1&5&2&6 \\
    ETH&  &1&3&2&5&6&4 \\
    DOGE& &4&3&1&6&2&5 \\
    LTC&  &4&3&2&5&1&6 \\
    \hline
    \multicolumn{2}{c}{Mean Rank} &3.25&3&\textbf{1.5}&5.25&2.75&5.25 \\
    \hline
    BTC& \multirow{4}{*}{Multivariate} &3&1&2&5&6&4 \\
    ETH&  &5&4&3&1&2&6 \\
    DOGE&  &2&3&1&6&4&5 \\
    LTC&  &1&3&5&6&2&4 \\
    \hline
    \multicolumn{2}{c}{Mean Rank}
    &\textbf{2.75}&\textbf{2.75}&\textbf{2.75}&4.5&3.5&4.75 \\
    \hline
    BTC& \multirow{4}{*}{Multivariate*} &4&5&3&6&1&2 \\
    ETH&  &4&6&3&5&1&2 \\
    DOGE&  &3&5&4&6&1&2 \\
    LTC&  &3&5&6&4&1&2 \\
    \hline
    \multicolumn{2}{c}{Mean Rank} &3.5&5.25&4&5.25&\textbf{1}&2 \\
    \hline \hline
    \end{tabular}
    \caption{Performance (rank) of different models for predicting cryptocurrency for Experiment 1.}
    \label{trank1}
\end{table*} 

\begin{table*}[htbp!]
    \centering
    \small
    \begin{tabular}{c c c c c c c c}
    \hline \hline
    Data & Strategy & LSTM & ED-LSTM & BD-LSTM & CNN & Conv-LSTM &Transformer \\
    \hline
    BTC & \multirow{4}{*}{Univariate} & 4&3&1&5&2&6 \\
    ETH & &3&2&1&4&5&6 \\
    DOGE& &6&4&1&5&2&3 \\
    LTC & &3&1&2&4&5&6 \\
    \hline
    \multicolumn{2}{c}{Mean Rank} &4&2.5&\textbf{1.25}&4.5&3.5&5.25\\
    \hline
    BTC& \multirow{4}{*}{Multivariate*} &3&4&2&6&1&5 \\
    ETH&  &4&3&2&6&1&5 \\
    DOGE&  &3&4&2&5&1&6\\
    LTC& &3&4&2&6&1&5 \\
    \hline
    \multicolumn{2}{c}{Mean Rank} &3.25&3.75&2&5.75&\textbf{1}&5.25\\
    \hline \hline
    \end{tabular}
    \caption{Performance (rank) of different models for predicting cryptocurrency for Experiment 2.}
    \label{trank2}
\end{table*}

\section{Discussion} \label{ch5}


We first reviewed the results of the first experiment that investigated the model performance without COVID-19 data. Our results show that Conv-LSTM with Multivariate* strategy provided outstanding predictive performance across four different cryptocurrencies.  We also note that in all the cryptocurrencies, the Univariate model was poorer than the Multivariate* strategy. The Univariate Strategy was generally better in accuracy than the Multivariae strategy for Experiment 1 (Tables 10 and 11). We found that the  models with high forecast accuracy(lower RMSE)  were mostly accompanied by narrower confidence intervals (eg. Figure 22). On the contrary, the higher RMSE values usually resulted in lower robustness of the model, ie. higher confidence interval.    The accuracy of performance generally declines as the prediction horizons increase, which is natural for multistep ahead problems (Figure \ref{12}). The prediction is derived from the current values, and the information gap expands as the prediction step increases. This is because our task is defined as a direct approach for forecasting multisteps, rather than an iterated prediction strategy.  
We found that the CNN-related models using convolution operation provided better accuracy than other models in predicting cryptocurrency price. Later we will analyse the cause of this issue. Among the predictions for the four currencies in Experiment 1 (Table \ref{t511b}), Dogecoin has the worst prediction effect. The RMSE   values of the model are significantly higher than those of the other three cryptocurrencies. We believe this is due to the particularity of the Dogecoin price, which leads to large prediction errors. The first 70\% of the data fluctuates smoothly, while the last 30\% of the data fluctuates highly.  

In Experiment 2, we evaluated the two strategies (Univariate and Multivariate*)  for the dataset featuring COVID-19 and beyond. It has been discovered that the close-price forecasts for Bitcoin, Dogecoin, and Ethereum have all shown improvements. Also, as the prediction horizon rises, the prediction accuracy of the model deteriorates more slowly. We claim that there are two causes contributing to the decrease in the accuracy of forecasting Litecoin. The first is that because our evaluation criterion for the model is the overall performance of the model, we did not choose Conv-LSTM, which had the best performance in predicting Litecoin in the previous experiment. Maybe Conv-LSTM is more suitable for predicting Litecoin. The second reason is that Litecoin was highly volatile before COVID-19. Due to our design of the experiments, the data during this period was not included in the training set.

Next, we aim to investigate what might be contributing to the lower accuracy of multivariate models compared to univariate models in prediction. In our analysis (Figure \ref{fclose} and \ref{fvolatility}), we noticed that the price of cryptocurrencies is extremely unstable and is greatly influenced by several variables outside or inside of the market \cite{farell2015analysis}. Simply inserting some additional factors will not only be ineffective in assisting the model to accurately forecast outcomes, but it may also mislead the model into acquiring knowledge of irrelevant data features. 
Our analysis of volatility (Figure \ref{fclose} and \ref{fvolatility}) shows the high degree of volatility exhibited by cryptocurrencies throughout the COVID-19 pandemic. In both experiments (Tables 10, 11 and 12), we observed that the use of a highly correlated crypto as a feature in the Multivariate* strategy  provided better prediction accuracy for the model.


Next, we investigate the factors contributing to the advantages and disadvantages of each model. Conventional time series models and machine learning techniques are inadequate for addressing issues such as timing dependency and gradient explosion. As we used MLP and ARIMA models to predict Bitcoin, the prediction performance was not as good as the deep learning model. Jiang et al. \cite{jiang2018time} highlighted that there is a long-term memory in the cryptocurrency market.   The memory gate in the LSTM network can better capture information in time series with long-term dependencies and the convolutional layer in CNN is better at capturing local patterns and features for multivariate datasets. Hence,  the combination of CNN with LSTM (Conv-LSTM), provided the best results.  Next, we analyze the differences between the four models with LSTM layers (LSTM, ED-SLTM, BD-LSTM and Conv-LSTM). The ED-LSTM model has been created for language modelling problems, particularly for sequence modelling for language translation. In this model, an encoder LSTM is used to transform a source sequence into a fixed-length vector, while a decoder LSTM is employed to convert the vector representation back into a variable-length target sequence \cite{sutskever2014sequence}. In our study, the encoder function maps an input time series to a vector of fixed length. Subsequently, the decoder LSTM function translates the vector representation to several prediction horizons. Despite the differences in the application, the fundamental objective of mapping inputs to outputs remains unchanged. As a result, ED-LSTM models have proven to be highly efficient for multi-step ahead prediction. The BD-LSTM model utilises two LSTM models to capture both forward and backward information about the sequence at each time step \cite{graves2005framewise}. Although these models (BD-LSTM and LD-LSTM) have been shown effective for language modelling, our findings indicate that they're useful in both univariate and multivariate strategies and provided competitive performance with Conv-LSTM. Although Conv-LSTM uses convolutional layers as input  with  LSTM  memory cells in the hidden layer, it differs from the conventional LSTM models since the memory cells from different hidden layers update only in the time domain and are mutually independent \cite{wang2017predrnn}. Therefore, information at the top layer in time $t-1$ will be ignored by the bottom layer at time $t$. In cryptocurrency price prediction, the time information is crucial. This also explains why the prediction accuracy of Conv-LSTM in our experiments is often low and unstable at high prediction horizons. The Transformer model that has enhanced attention mechanism provided unsatisfactory results in Experiment 2 (lowest rank in Table 14)  and Experiment 1 (lowest rank in Table 13 for Univariate and Multivariate strategies. However, the Transformer model was number 2 rank in the case of Multivariate* for Experiment 1 which suggests that it works better with data that has less vitality (pre COVID-19). The issue with poor results may be attributed to the limited training data, as Transformer models are often better suited for handling large amounts of data.

 Our findings indicate that combining the closing prices of the highest-correlated cryptocurrencies into the multivariate model leads to a significant enhancement in the model's accuracy in forecasting. The Conv-LSTM model provides the most accurate predictions when employing this strategy for forecasting cryptocurrencies for both experiments. Under the Multivariate* approach, the scalability and accuracy of models becomes better. This means that  there is minimal difference in the accuracy of predicting 1 day and predicting 5 days.  Generally, we found that Experiment 2 has lower model accuracy than Experiment 1, since the latter features data during COVID-19 which features phases of high volatility.


The major limitation of the framework deals with a lack of uncertainty quantification which can be addressed by Bayesian deep learning, which includes MCMC sampling \cite{chandra2021bayesian,Chandra2024tut} and variational inference-based deep learning models \cite{ferianc2020vinnas,kapoor2023cyclone}. Furthermore, data augmentation methods \cite{shorten2019survey,khan2023review} can be explored to exploit segments of highly correlated crypto and stick prices, from different historical periods. We featured the prediction of daily price, but the model can be extended for regular hourly price prediction and also be extended to include extreme value forecast strategies, such as quantile regression \cite{koenker1978regression}. Furthermore, enhanced Bayesian data imputation methodology can extrapolate missing data in cases better, when highly correlated data sources have such issues. Finally, we only looked at unimodal data. Due to the complexity of the cryptocurrency market, further extension can be done using a multimodal data approach \cite{ngiam2011multimodal,tavakoli2023multi} that considers text data from news media and social media using large language models.

\section{Conclusions} \label{ch6}

In this study, we provide a rigorous evaluation of novel deep learning models for cryptocurrency price forecasting. We compared prominent deep learning models using  univariate and multivariate strategies.   The results show that the convolutional LSTM with multivariate strategy provides the highest accuracy in predicting cryptocurrency prices for two major experimental setup featuring data, period to and during COVID-19 pandemic. 
 Additionally, our findings indicate that the multivariate strategy featuring  a highly correlated cryptocurrency data enhanced prediction accuracy significantly, in comparison to the univariate strategy.   
In terms of the effect of COVID-19, we found that close-price volatility for cryptocurrency is quite apparent, which brings further challenges to the respective models. Our experimental results show that utilising a training data set with high volatility weakens the precision of our predictions.

In future work, it would be worthwhile to improve the multivariate model further using more correlated factors to enhance forecasts by employing techniques such as causal inference to find more correlated data in the multivariate strategy. We can also use a framework for predictions of other financial indicators, such as the volatility of cryptocurrency prices and market stocks. Further applications to other domains, such as predicting energy use and extreme weather forecasting, can also be done. 

\section{Code and Data}

We provide open source code and data using GitHub repository \footnote{\url{https://github.com/sydney-machine-learning/deeplearning-crypto}}.

 \bibliographystyle{elsarticle-num} 
 \bibliography{mybib}

\begin{thebibliography}{100}
\expandafter\ifx\csname url\endcsname\relax
  \def\url#1{\texttt{#1}}\fi
\expandafter\ifx\csname urlprefix\endcsname\relax\def\urlprefix{URL }\fi
\expandafter\ifx\csname href\endcsname\relax
  \def\href#1#2{#2} \def\path#1{#1}\fi

\bibitem{bose2019financial}
S.~Bose, G.~Dong, A.~Simpson, S.~Bose, G.~Dong, A.~Simpson, The financial ecosystem, Springer, 2019.

\bibitem{frankel2008fiscal}
J.~Frankel, B.~Smit, F.~Sturzenegger, Fiscal and monetary policy in a commodity-based economy 1, Economics of transition 16~(4) (2008) 679--713.

\bibitem{hayek1990denationalisation}
F.~A. Hayek, Denationalisation of money: the argument refined: an analysis of the theory and practice of concurrent currencies, Vol.~70, Institute of economic affairs, 1990.

\bibitem{gross2019money}
M.~M. Gross, C.~Siebenbrunner, Money creation in fiat and digital currency systems, International Monetary Fund, 2019.

\bibitem{boyd2001impact}
J.~H. Boyd, R.~Levine, B.~D. Smith, The impact of inflation on financial sector performance, Journal of monetary Economics 47~(2) (2001) 221--248.

\bibitem{milkau2015digitalisation}
U.~Milkau, J.~Bott, Digitalisation in payments: From interoperability to centralised models?, Journal of Payments Strategy \& Systems 9~(3) (2015) 321--340.

\bibitem{chaum1983blind}
D.~Chaum, Blind signatures for untraceable payments, in: Advances in Cryptology: Proceedings of Crypto 82, Springer, 1983, pp. 199--203.

\bibitem{nakamoto2008bitcoin}
S.~Nakamoto, Bitcoin: A peer-to-peer electronic cash system (2008).

\bibitem{manimuthu2019literature}
A.~Manimuthu, G.~Rejikumar, D.~Marwaha, et~al., A literature review on {Bitcoin:} transformation of crypto currency into a global phenomenon, IEEE Engineering Management Review 47~(1) (2019) 28--35.

\bibitem{farell2015analysis}
R.~Farell, An analysis of the cryptocurrency industry, Wharton Research Scholars 130 (2015) 1--23.

\bibitem{eyal2017blockchain}
I.~Eyal, Blockchain technology: Transforming libertarian cryptocurrency dreams to finance and banking realities, Computer 50~(9) (2017) 38--49.

\bibitem{jang2017empirical}
H.~Jang, J.~Lee, An empirical study on modeling and prediction of bitcoin prices with bayesian neural networks based on blockchain information, IEEE access 6 (2017) 5427--5437.

\bibitem{saad2019toward}
M.~Saad, J.~Choi, D.~Nyang, J.~Kim, A.~Mohaisen, Toward characterizing blockchain-based cryptocurrencies for highly accurate predictions, IEEE Systems Journal 14~(1) (2019) 321--332.

\bibitem{corbet2018datestamping}
S.~Corbet, B.~Lucey, L.~Yarovaya, Datestamping the bitcoin and ethereum bubbles, Finance Research Letters 26 (2018) 81--88.

\bibitem{bhosale2018volatility}
J.~Bhosale, S.~Mavale, Volatility of select crypto-currencies: A comparison of bitcoin, ethereum and litecoin, Annu. Res. J. SCMS, Pune 6~(1) (2018) 132--141.

\bibitem{katsiampa2019empirical}
P.~Katsiampa, An empirical investigation of volatility dynamics in the cryptocurrency market, Research in International Business and Finance 50 (2019) 322--335.

\bibitem{elendner2016cross}
H.~Elendner, S.~Trimborn, B.~Ong, T.~M. Lee, The cross-section of crypto-currencies as financial assets: An overview (2016).

\bibitem{seabe2023forecasting}
P.~L. Seabe, C.~R.~B. Moutsinga, E.~Pindza, Forecasting cryptocurrency prices using lstm, gru, and bi-directional lstm: A deep learning approach, Fractal and Fractional 7~(2) (2023) 203.

\bibitem{kyriazis2020systematic}
N.~Kyriazis, S.~Papadamou, S.~Corbet, A systematic review of the bubble dynamics of cryptocurrency prices, Research in International Business and Finance 54 (2020) 101254.

\bibitem{ammer2022deep}
M.~A. Ammer, T.~H. Aldhyani, Deep learning algorithm to predict cryptocurrency fluctuation prices: Increasing investment awareness, Electronics 11~(15) (2022) 2349.

\bibitem{murray2023forecasting}
K.~Murray, A.~Rossi, D.~Carraro, A.~Visentin, On forecasting cryptocurrency prices: A comparison of machine learning, deep learning, and ensembles, Forecasting 5~(1) (2023) 196--209.

\bibitem{wang2023machine}
Y.~Wang, G.~Andreeva, B.~Martin-Barragan, Machine learning approaches to forecasting cryptocurrency volatility: Considering internal and external determinants, International Review of Financial Analysis 90 (2023) 102914.

\bibitem{kyriazis2019survey}
N.~A. Kyriazis, A survey on empirical findings about spillovers in cryptocurrency markets, Journal of Risk and Financial Management 12~(4) (2019) 170.

\bibitem{stock2001vector}
J.~H. Stock, M.~W. Watson, Vector autoregressions, Journal of Economic perspectives 15~(4) (2001) 101--115.

\bibitem{duan1995garch}
J.-C. Duan, The garch option pricing model, Mathematical finance 5~(1) (1995) 13--32.

\bibitem{huynh2020small}
T.~L.~D. Huynh, M.~A. Nasir, X.~V. Vo, T.~T. Nguyen, “small things matter most”: The spillover effects in the cryptocurrency market and gold as a silver bullet, The North American Journal of Economics and Finance 54 (2020) 101277.

\bibitem{baltruvsaitis2018multimodal}
T.~Baltru{\v{s}}aitis, C.~Ahuja, L.-P. Morency, Multimodal machine learning: A survey and taxonomy, IEEE transactions on pattern analysis and machine intelligence 41~(2) (2018) 423--443.

\bibitem{wang2020deep}
S.~Wang, J.~Cao, S.~Y. Philip, Deep learning for spatio-temporal data mining: A survey, IEEE transactions on knowledge and data engineering 34~(8) (2020) 3681--3700.

\bibitem{lim2021time}
B.~Lim, S.~Zohren, Time-series forecasting with deep learning: a survey, Philosophical Transactions of the Royal Society A 379~(2194) (2021) 20200209.

\bibitem{jacques2022deep}
V.~Jacques-Dumas, F.~Ragone, P.~Borgnat, P.~Abry, F.~Bouchet, Deep learning-based extreme heatwave forecast, Frontiers in Climate 4 (2022).

\bibitem{mahjoub2022predicting}
S.~Mahjoub, L.~Chrifi-Alaoui, B.~Marhic, L.~Delahoche, Predicting energy consumption using lstm, multi-layer gru and drop-gru neural networks, Sensors 22~(11) (2022) 4062.

\bibitem{chandra2021bayesian}
R.~Chandra, Y.~He, Bayesian neural networks for stock price forecasting before and during covid-19 pandemic, Plos one 16~(7) (2021) e0253217.

\bibitem{livieris2020ensemble}
I.~E. Livieris, E.~Pintelas, S.~Stavroyiannis, P.~Pintelas, Ensemble deep learning models for forecasting cryptocurrency time-series, Algorithms 13~(5) (2020) 121.

\bibitem{ferdiansyah2019lstm}
F.~Ferdiansyah, S.~H. Othman, R.~Z. R.~M. Radzi, D.~Stiawan, Y.~Sazaki, U.~Ependi, A lstm-method for bitcoin price prediction: A case study yahoo finance stock market, in: 2019 international conference on electrical engineering and computer science (ICECOS), IEEE, 2019, pp. 206--210.

\bibitem{wu2018new}
C.-H. Wu, C.-C. Lu, Y.-F. Ma, R.-S. Lu, A new forecasting framework for bitcoin price with lstm, in: 2018 IEEE international conference on data mining workshops (ICDMW), IEEE, 2018, pp. 168--175.

\bibitem{hochreiter1997long}
S.~Hochreiter, J.~Schmidhuber, Long short-term memory, Neural computation 9~(8) (1997) 1735--1780.

\bibitem{yu2019review}
Y.~Yu, X.~Si, C.~Hu, J.~Zhang, A review of recurrent neural networks: {LSTM} cells and network architectures, Neural computation 31~(7) (2019) 1235--1270.

\bibitem{jiang2017cryptocurrency}
Z.~Jiang, J.~Liang, Cryptocurrency portfolio management with deep reinforcement learning, in: 2017 Intelligent systems conference (IntelliSys), IEEE, 2017, pp. 905--913.

\bibitem{sridhar2021multi}
S.~Sridhar, S.~Sanagavarapu, Multi-head self-attention transformer for dogecoin price prediction, in: 2021 14th International Conference on Human System Interaction (HSI), IEEE, 2021, pp. 1--6.

\bibitem{chandra2021evaluation}
R.~Chandra, S.~Goyal, R.~Gupta, Evaluation of deep learning models for multi-step ahead time series prediction, Ieee Access 9 (2021) 83105--83123.

\bibitem{tsay2005analysis}
R.~S. Tsay, Analysis of financial time series, John wiley \& sons, 2005.

\bibitem{fischer2018deep}
T.~Fischer, C.~Krauss, Deep learning with long short-term memory networks for financial market predictions, European journal of operational research 270~(2) (2018) 654--669.

\bibitem{sezer2020financial}
O.~B. Sezer, M.~U. Gudelek, A.~M. Ozbayoglu, Financial time series forecasting with deep learning: A systematic literature review: 2005--2019, Applied soft computing 90 (2020) 106181.

\bibitem{plakandaras2015market}
V.~Plakandaras, T.~Papadimitriou, P.~Gogas, K.~Diamantaras, Market sentiment and exchange rate directional forecasting, Algorithmic Finance 4~(1-2) (2015) 69--79.

\bibitem{nabipour2020predicting}
M.~Nabipour, P.~Nayyeri, H.~Jabani, S.~Shahab, A.~Mosavi, Predicting stock market trends using machine learning and deep learning algorithms via continuous and binary data; a comparative analysis, Ieee Access 8 (2020) 150199--150212.

\bibitem{kong2021predicting}
A.~Kong, H.~Zhu, R.~Azencott, Predicting intraday jumps in stock prices using liquidity measures and technical indicators, Journal of Forecasting 40~(3) (2021) 416--438.

\bibitem{elman1990finding}
J.~L. Elman, Finding structure in time, Cognitive science 14~(2) (1990) 179--211.

\bibitem{mehtab2021stock}
S.~Mehtab, J.~Sen, A.~Dutta, Stock price prediction using machine learning and lstm-based deep learning models, in: Machine Learning and Metaheuristics Algorithms, and Applications: Second Symposium, SoMMA 2020, Chennai, India, October 14--17, 2020, Revised Selected Papers 2, Springer, 2021, pp. 88--106.

\bibitem{rezaei2021stock}
H.~Rezaei, H.~Faaljou, G.~Mansourfar, Stock price prediction using deep learning and frequency decomposition, Expert Systems with Applications 169 (2021) 114332.

\bibitem{rilling2003empirical}
G.~Rilling, P.~Flandrin, P.~Goncalves, et~al., On empirical mode decomposition and its algorithms, in: IEEE-EURASIP workshop on nonlinear signal and image processing, Vol.~3, Grado: IEEE, 2003, pp. 8--11.

\bibitem{torres2011complete}
M.~E. Torres, M.~A. Colominas, G.~Schlotthauer, P.~Flandrin, A complete ensemble empirical mode decomposition with adaptive noise, in: 2011 IEEE international conference on acoustics, speech and signal processing (ICASSP), IEEE, 2011, pp. 4144--4147.

\bibitem{jing2021hybrid}
N.~Jing, Z.~Wu, H.~Wang, A hybrid model integrating deep learning with investor sentiment analysis for stock price prediction, Expert Systems with Applications 178 (2021) 115019.

\bibitem{mehtab2022stock}
S.~Mehtab, J.~Sen, Stock price prediction using machine learning and deep learning algorithms and models, Machine Learning in the Analysis and Forecasting of Financial Time Series (2022) 235--303.

\bibitem{li2022novel}
Y.~Li, Y.~Pan, A novel ensemble deep learning model for stock prediction based on stock prices and news, International Journal of Data Science and Analytics 13~(2) (2022) 139--149.

\bibitem{kanwal2022bicudnnlstm}
A.~Kanwal, M.~F. Lau, S.~P. Ng, K.~Y. Sim, S.~Chandrasekaran, Bicudnnlstm-1dcnn—a hybrid deep learning-based predictive model for stock price prediction, Expert Systems with Applications 202 (2022) 117123.

\bibitem{swathi2022optimal}
T.~Swathi, N.~Kasiviswanath, A.~A. Rao, An optimal deep learning-based lstm for stock price prediction using twitter sentiment analysis, Applied Intelligence 52~(12) (2022) 13675--13688.

\bibitem{ben2023forecasting}
H.~Ben~Ameur, S.~Boubaker, Z.~Ftiti, W.~Louhichi, K.~Tissaoui, Forecasting commodity prices: empirical evidence using deep learning tools, Annals of Operations Research (2023) 1--19.

\bibitem{baser2023gold}
P.~Baser, J.~R. Saini, N.~Baser, Gold commodity price prediction using tree-based prediction models, International Journal of Intelligent Systems and Applications in Engineering 11~(1s) (2023) 90--96.

\bibitem{deepa2023machine}
S.~Deepa, A.~Alli, S.~Gokila, et~al., Machine learning regression model for material synthesis prices prediction in agriculture, Materials Today: Proceedings 81 (2023) 989--993.

\bibitem{zhao2023deep}
Y.~Zhao, G.~Yang, Deep learning-based integrated framework for stock price movement prediction, Applied Soft Computing 133 (2023) 109921.

\bibitem{almeida2015bitcoin}
J.~Almeida, S.~Tata, A.~Moser, V.~Smit, Bitcoin prediciton using ann, Neural networks 7 (2015) 1--12.

\bibitem{mallqui2019predicting}
D.~C. Mallqui, R.~A. Fernandes, Predicting the direction, maximum, minimum and closing prices of daily bitcoin exchange rate using machine learning techniques, Applied Soft Computing 75 (2019) 596--606.

\bibitem{quek2022new}
S.~G. Quek, G.~Selvachandran, J.~H. Tan, H.~Y.~A. Thiang, N.~T. Tuan, et~al., A new hybrid model of fuzzy time series and genetic algorithm based machine learning algorithm: a case study of forecasting prices of nine types of major cryptocurrencies, Big Data Research 28 (2022) 100315.

\bibitem{radityo2017prediction}
A.~Radityo, Q.~Munajat, I.~Budi, Prediction of bitcoin exchange rate to american dollar using artificial neural network methods, in: 2017 international conference on advanced computer science and information systems (ICACSIS), IEEE, 2017, pp. 433--438.

\bibitem{greaves2015using}
A.~Greaves, B.~Au, Using the bitcoin transaction graph to predict the price of bitcoin, No data 8 (2015) 416--443.

\bibitem{cortes1995support}
C.~Cortes, V.~Vapnik, Support-vector networks, Machine learning 20 (1995) 273--297.

\bibitem{sovbetov2018factors}
Y.~Sovbetov, Factors influencing cryptocurrency prices: Evidence from bitcoin, ethereum, dash, litcoin, and monero, Journal of Economics and Financial Analysis 2~(2) (2018) 1--27.

\bibitem{guo2018bitcoin}
T.~Guo, A.~Bifet, N.~Antulov-Fantulin, Bitcoin volatility forecasting with a glimpse into buy and sell orders, in: 2018 IEEE international conference on data mining (ICDM), IEEE, 2018, pp. 989--994.

\bibitem{akcora2018forecasting}
C.~G. Akcora, A.~K. Dey, Y.~R. Gel, M.~Kantarcioglu, Forecasting bitcoin price with graph chainlets, in: Advances in Knowledge Discovery and Data Mining: 22nd Pacific-Asia Conference, PAKDD 2018, Melbourne, VIC, Australia, June 3-6, 2018, Proceedings, Part III 22, Springer, 2018, pp. 765--776.

\bibitem{roy2018bitcoin}
S.~Roy, S.~Nanjiba, A.~Chakrabarty, Bitcoin price forecasting using time series analysis, in: 2018 21st International Conference of Computer and Information Technology (ICCIT), IEEE, 2018, pp. 1--5.

\bibitem{derbentsev2019forecasting}
V.~Derbentsev, N.~Datsenko, O.~Stepanenko, V.~Bezkorovainyi, Forecasting cryptocurrency prices time series using machine learning approach, in: SHS Web of Conferences, Vol.~65, EDP Sciences, 2019, p. 02001.

\bibitem{aanandhi2021cryptocurrency}
S.~Aanandhi, S.~Akhilaa, V.~Vardarajan, M.~Sathiyanarayanan, et~al., Cryptocurrency price prediction using time series forecasting (arima), in: 2021 4th International Seminar on Research of Information Technology and Intelligent Systems (ISRITI), IEEE, 2021, pp. 598--602.

\bibitem{latif2023comparative}
N.~Latif, J.~D. Selvam, M.~Kapse, V.~Sharma, V.~Mahajan, Comparative performance of lstm and arima for the short-term prediction of bitcoin prices, Australasian Accounting, Business and Finance Journal 17~(1) (2023) 256--276.

\bibitem{maleki2023bitcoin}
N.~Maleki, A.~Nikoubin, M.~Rabbani, Y.~Zeinali, Bitcoin price prediction based on other cryptocurrencies using machine learning and time series analysis, Scientia Iranica 30~(1) (2023) 285--301.

\bibitem{kaelbling1996reinforcement}
L.~P. Kaelbling, M.~L. Littman, A.~W. Moore, Reinforcement learning: A survey, Journal of artificial intelligence research 4 (1996) 237--285.

\bibitem{lee2018generating}
K.~Lee, S.~Ulkuatam, P.~Beling, W.~Scherer, Generating synthetic bitcoin transactions and predicting market price movement via inverse reinforcement learning and agent-based modeling, Journal of Artificial Societies and Social Simulation 21~(3) (2018).

\bibitem{ly2018applying}
B.~Ly, D.~Timaul, A.~Lukanan, J.~Lau, E.~Steinmetz, Applying deep learning to better predict cryptocurrency trends, in: Midwest Instruction and Computing Symposium, 2018.

\bibitem{lucarelli2019deep}
G.~Lucarelli, M.~Borrotti, A deep reinforcement learning approach for automated cryptocurrency trading, in: Artificial Intelligence Applications and Innovations: 15th IFIP WG 12.5 International Conference, AIAI 2019, Hersonissos, Crete, Greece, May 24--26, 2019, Proceedings 15, Springer, 2019, pp. 247--258.

\bibitem{lahmiri2019cryptocurrency}
S.~Lahmiri, S.~Bekiros, Cryptocurrency forecasting with deep learning chaotic neural networks, Chaos, Solitons \& Fractals 118 (2019) 35--40.

\bibitem{patel2020deep}
M.~M. Patel, S.~Tanwar, R.~Gupta, N.~Kumar, A deep learning-based cryptocurrency price prediction scheme for financial institutions, Journal of information security and applications 55 (2020) 102583.

\bibitem{marne2020predicting}
S.~Marne, S.~Churi, D.~Correia, J.~Gomes, Predicting price of cryptocurrency--a deep learning approach, NTASU-9 (3) (2020).

\bibitem{nasekin2020deep}
S.~Nasekin, C.~Y.-H. Chen, Deep learning-based cryptocurrency sentiment construction, Digital Finance 2~(1) (2020) 39--67.

\bibitem{betancourt2021deep}
C.~Betancourt, W.-H. Chen, Deep reinforcement learning for portfolio management of markets with a dynamic number of assets, Expert Systems with Applications 164 (2021) 114002.

\bibitem{arulkumaran2017deep}
K.~Arulkumaran, M.~P. Deisenroth, M.~Brundage, A.~A. Bharath, Deep reinforcement learning: A brief survey, IEEE Signal Processing Magazine 34~(6) (2017) 26--38.

\bibitem{shahbazi2021improving}
Z.~Shahbazi, Y.-C. Byun, Improving the cryptocurrency price prediction performance based on reinforcement learning, IEEE Access 9 (2021) 162651--162659.

\bibitem{d2022deep}
V.~D’Amato, S.~Levantesi, G.~Piscopo, Deep learning in predicting cryptocurrency volatility, Physica A: Statistical Mechanics and its Applications 596 (2022) 127158.

\bibitem{schnaubelt2022deep}
M.~Schnaubelt, Deep reinforcement learning for the optimal placement of cryptocurrency limit orders, European Journal of Operational Research 296~(3) (2022) 993--1006.

\bibitem{parekh2022dl}
R.~Parekh, N.~P. Patel, N.~Thakkar, R.~Gupta, S.~Tanwar, G.~Sharma, I.~E. Davidson, R.~Sharma, Dl-guess: Deep learning and sentiment analysis-based cryptocurrency price prediction, IEEE Access 10 (2022) 35398--35409.

\bibitem{kim2022deep}
G.~Kim, D.-H. Shin, J.~G. Choi, S.~Lim, A deep learning-based cryptocurrency price prediction model that uses on-chain data, IEEE Access 10 (2022) 56232--56248.

\bibitem{goutte2023deep}
S.~Goutte, H.-V. Le, F.~Liu, H.-J. Von~Mettenheim, Deep learning and technical analysis in cryptocurrency market, Finance Research Letters 54 (2023) 103809.

\bibitem{yen2021economic}
K.-C. Yen, H.-P. Cheng, Economic policy uncertainty and cryptocurrency volatility, Finance Research Letters 38 (2021) 101428.

\bibitem{woebbeking2021cryptocurrency}
F.~Woebbeking, Cryptocurrency volatility markets, Digital finance 3~(3) (2021) 273--298.

\bibitem{cross2021returns}
J.~L. Cross, C.~Hou, K.~Trinh, Returns, volatility and the cryptocurrency bubble of 2017--18, Economic Modelling 104 (2021) 105643.

\bibitem{ftiti2023cryptocurrency}
Z.~Ftiti, W.~Louhichi, H.~Ben~Ameur, Cryptocurrency volatility forecasting: What can we learn from the first wave of the covid-19 outbreak?, Annals of Operations Research 330~(1) (2023) 665--690.

\bibitem{yin2021understanding}
L.~Yin, J.~Nie, L.~Han, Understanding cryptocurrency volatility: The role of oil market shocks, International Review of Economics \& Finance 72 (2021) 233--253.

\bibitem{catania2018predicting}
L.~Catania, S.~Grassi, F.~Ravazzolo, Predicting the volatility of cryptocurrency time-series, Mathematical and Statistical Methods for Actuarial Sciences and Finance: MAF 2018 (2018) 203--207.

\bibitem{catania2022forecasting}
L.~Catania, S.~Grassi, Forecasting cryptocurrency volatility, International Journal of Forecasting 38~(3) (2022) 878--894.

\bibitem{ma2020cryptocurrency}
F.~Ma, C.~Liang, Y.~Ma, M.~I.~M. Wahab, Cryptocurrency volatility forecasting: A markov regime-switching midas approach, Journal of Forecasting 39~(8) (2020) 1277--1290.

\bibitem{wei2023cryptocurrency}
Y.~Wei, Y.~Wang, B.~M. Lucey, S.~A. Vigne, Cryptocurrency uncertainty and volatility forecasting of precious metal futures markets, Journal of Commodity Markets 29 (2023) 100305.

\bibitem{box2015time}
G.~E. Box, G.~M. Jenkins, G.~C. Reinsel, G.~M. Ljung, Time series analysis: forecasting and control, John Wiley \& Sons, 2015.

\bibitem{hochreiter1998vanishing}
S.~Hochreiter, The vanishing gradient problem during learning recurrent neural nets and problem solutions, International Journal of Uncertainty, Fuzziness and Knowledge-Based Systems 6~(02) (1998) 107--116.

\bibitem{graves2005framewise}
A.~Graves, J.~Schmidhuber, Framewise phoneme classification with bidirectional lstm and other neural network architectures, Neural networks 18~(5-6) (2005) 602--610.

\bibitem{liu2016learning}
Y.~Liu, C.~Sun, L.~Lin, X.~Wang, Learning natural language inference using bidirectional lstm model and inner-attention, arXiv preprint arXiv:1605.09090 (2016).

\bibitem{chen2018end}
L.~Chen, J.~Tao, S.~Ghaffarzadegan, Y.~Qian, End-to-end neural network based automated speech scoring, in: 2018 IEEE international conference on acoustics, speech and signal processing (ICASSP), IEEE, 2018, pp. 6234--6238.

\bibitem{sutskever2014sequence}
I.~Sutskever, O.~Vinyals, Q.~V. Le, Sequence to sequence learning with neural networks, Advances in neural information processing systems 27 (2014).

\bibitem{cho2014learning}
K.~Cho, B.~Van~Merri{\"e}nboer, C.~Gulcehre, D.~Bahdanau, F.~Bougares, H.~Schwenk, Y.~Bengio, Learning phrase representations using {RNN} encoder-decoder for statistical machine translation, arXiv preprint arXiv:1406.1078 (2014).

\bibitem{gunduz2017intraday}
H.~Gunduz, Y.~Yaslan, Z.~Cataltepe, Intraday prediction of borsa istanbul using convolutional neural networks and feature correlations, Knowledge-Based Systems 137 (2017) 138--148.

\bibitem{di2016artificial}
L.~Di~Persio, O.~Honchar, et~al., Artificial neural networks architectures for stock price prediction: Comparisons and applications, International journal of circuits, systems and signal processing 10 (2016) 403--413.

\bibitem{hoseinzade2019cnnpred}
E.~Hoseinzade, S.~Haratizadeh, Cnnpred: Cnn-based stock market prediction using a diverse set of variables, Expert Systems with Applications 129 (2019) 273--285.

\bibitem{siripurapu2014convolutional}
A.~Siripurapu, Convolutional networks for stock trading, Stanford Univ Dep Comput Sci 1~(2) (2014) 1--6.

\bibitem{jiang2017face}
H.~Jiang, E.~Learned-Miller, Face detection with the faster r-cnn, in: 2017 12th IEEE international conference on automatic face \& gesture recognition (FG 2017), IEEE, 2017, pp. 650--657.

\bibitem{garcia2017review}
A.~Garcia-Garcia, S.~Orts-Escolano, S.~Oprea, V.~Villena-Martinez, J.~Garcia-Rodriguez, A review on deep learning techniques applied to semantic segmentation, arXiv preprint arXiv:1704.06857 (2017).

\bibitem{kingma2014adam}
D.~P. Kingma, J.~Ba, Adam: A method for stochastic optimization, arXiv preprint arXiv:1412.6980 (2014).

\bibitem{shi2015convolutional}
X.~Shi, Z.~Chen, H.~Wang, D.-Y. Yeung, W.-K. Wong, W.-c. Woo, Convolutional lstm network: A machine learning approach for precipitation nowcasting, Advances in neural information processing systems 28 (2015).

\bibitem{cho2014properties}
K.~Cho, B.~Van~Merri{\"e}nboer, D.~Bahdanau, Y.~Bengio, On the properties of neural machine translation: Encoder-decoder approaches, arXiv preprint arXiv:1409.1259 (2014).

\bibitem{bahdanau2014neural}
D.~Bahdanau, K.~Cho, Y.~Bengio, Neural machine translation by jointly learning to align and translate, arXiv preprint arXiv:1409.0473 (2014).

\bibitem{vaswani2017attention}
A.~Vaswani, N.~Shazeer, N.~Parmar, J.~Uszkoreit, L.~Jones, A.~N. Gomez, {\L}.~Kaiser, I.~Polosukhin, Attention is all you need, Advances in neural information processing systems 30 (2017).

\bibitem{amari1993backpropagation}
S.-i. Amari, Backpropagation and stochastic gradient descent method, Neurocomputing 5~(4-5) (1993) 185--196.

\bibitem{ruder2016overview}
S.~Ruder, An overview of gradient descent optimization algorithms, arXiv preprint arXiv:1609.04747 (2016).

\bibitem{duchi2011adaptive}
J.~Duchi, E.~Hazan, Y.~Singer, Adaptive subgradient methods for online learning and stochastic optimization., Journal of machine learning research 12~(7) (2011).

\bibitem{zeiler2012adadelta}
M.~D. Zeiler, Adadelta: an adaptive learning rate method, arXiv preprint arXiv:1212.5701 (2012).

\bibitem{hinton2012neural}
G.~Hinton, N.~Srivastava, K.~Swersky, Neural networks for machine learning lecture 6a overview of mini-batch gradient descent, Toronto University (2012).

\bibitem{buterin2014next}
V.~Buterin, et~al., A next-generation smart contract and decentralized application platform, white paper 3~(37) (2014) 2--1.

\bibitem{percival2016scrypt}
C.~Percival, S.~Josefsson, The scrypt password-based key derivation function, Tech. rep. (2016).

\bibitem{web:kaggle:cryptoprice}
\href{https://www.kaggle.com/datasets/sudalairajkumar/cryptocurrencypricehistory/data}{Cryptocurrency historical prices}, last accessed 13 Feburary 2024 (2021).
\newline\urlprefix\url{https://www.kaggle.com/datasets/sudalairajkumar/cryptocurrencypricehistory/data}

\bibitem{takens2006detecting}
F.~Takens, Detecting strange attractors in turbulence, in: Dynamical Systems and Turbulence, Warwick 1980: proceedings of a symposium held at the University of Warwick 1979/80, Springer, 2006, pp. 366--381.

\bibitem{gnauck2004interpolation}
A.~Gnauck, Interpolation and approximation of water quality time series and process identification, Analytical and bioanalytical chemistry 380 (2004) 484--492.

\bibitem{paszke2019pytorch}
A.~Paszke, S.~Gross, F.~Massa, A.~Lerer, J.~Bradbury, G.~Chanan, T.~Killeen, Z.~Lin, N.~Gimelshein, L.~Antiga, et~al., Pytorch: An imperative style, high-performance deep learning library, Advances in neural information processing systems 32 (2019).

\bibitem{mandaci2022herding}
P.~E. Mandaci, E.~C. Cagli, Herding intensity and volatility in cryptocurrency markets during the {COVID-19}, Finance Research Letters 46 (2022) 102382.

\bibitem{naeem2021asymmetric}
M.~A. Naeem, E.~Bouri, Z.~Peng, S.~J.~H. Shahzad, X.~V. Vo, Asymmetric efficiency of cryptocurrencies during {COVID19}, Physica A: Statistical Mechanics and its Applications 565 (2021) 125562.

\bibitem{tanwar2022prediction}
A.~Tanwar, V.~Kumar, Prediction of cryptocurrency prices using transformers and long short term neural networks, in: 2022 International Conference on Intelligent Controller and Computing for Smart Power (ICICCSP), IEEE, 2022, pp. 1--4.

\bibitem{wang2006forecasting}
D.~Wang, W.-Z. Lu, Forecasting of ozone level in time series using mlp model with a novel hybrid training algorithm, Atmospheric Environment 40~(5) (2006) 913--924.

\bibitem{cucinotta2020declares}
D.~Cucinotta, M.~Vanelli, Who declares covid-19 a pandemic, Acta bio medica: Atenei parmensis 91~(1) (2020) 157.

\bibitem{andersen2020proximal}
K.~G. Andersen, A.~Rambaut, W.~I. Lipkin, E.~C. Holmes, R.~F. Garry, The proximal origin of sars-cov-2, Nature medicine 26~(4) (2020) 450--452.

\bibitem{brodeur2021literature}
A.~Brodeur, D.~Gray, A.~Islam, S.~Bhuiyan, A literature review of the economics of covid-19, Journal of economic surveys 35~(4) (2021) 1007--1044.

\bibitem{leach2021post}
M.~Leach, H.~MacGregor, I.~Scoones, A.~Wilkinson, Post-pandemic transformations: How and why {COVID-19} requires us to rethink development, World development 138 (2021) 105233.

\bibitem{miao2023immunogen}
G.~Miao, Z.~Chen, H.~Cao, W.~Wu, X.~Chu, H.~Liu, L.~Zhang, H.~Zhu, H.~Cai, X.~Lu, et~al., From immunogen to {COVID-19} vaccines: Prospects for the post-pandemic era, Biomedicine \& Pharmacotherapy 158 (2023) 114208.

\bibitem{laskawiec2022post}
D.~{\L}askawiec, M.~Grajek, P.~Szlacheta, I.~Korzonek-Szlacheta, Post-pandemic stress disorder as an effect of the epidemiological situation related to the {COVID-19} pandemic, in: Healthcare, Vol.~10, 2022, p. 975.

\bibitem{jiang2018time}
Y.~Jiang, H.~Nie, W.~Ruan, Time-varying long-term memory in bitcoin market, Finance Research Letters 25 (2018) 280--284.

\bibitem{wang2017predrnn}
Y.~Wang, M.~Long, J.~Wang, Z.~Gao, P.~S. Yu, Predrnn: Recurrent neural networks for predictive learning using spatiotemporal lstms, Advances in neural information processing systems 30 (2017).

\bibitem{Chandra2024tut}
R.~Chandra, J.~Simmons, Bayesian neural networks via {MCMC:} a python-based tutorial, IEEE Access 12 (2024) 70519--70549.
\newblock \href {https://doi.org/10.1109/ACCESS.2024.3401234} {\path{doi:10.1109/ACCESS.2024.3401234}}.

\bibitem{ferianc2020vinnas}
M.~Ferianc, H.~Fan, M.~Rodrigues, Vinnas: Variational inference-based neural network architecture search, arXiv preprint arXiv:2007.06103 (2020).

\bibitem{kapoor2023cyclone}
A.~Kapoor, A.~Negi, L.~Marshall, R.~Chandra, Cyclone trajectory and intensity prediction with uncertainty quantification using variational recurrent neural networks, Environmental Modelling \& Software 162 (2023) 105654.

\bibitem{shorten2019survey}
C.~Shorten, T.~M. Khoshgoftaar, A survey on image data augmentation for deep learning, Journal of big data 6~(1) (2019) 1--48.

\bibitem{khan2023review}
A.~A. Khan, O.~Chaudhari, R.~Chandra, A review of ensemble learning and data augmentation models for class imbalanced problems: Combination, implementation and evaluation, Expert Systems with Applications (2023) 122778.

\bibitem{koenker1978regression}
R.~Koenker, G.~Bassett~Jr, Regression quantiles, Econometrica: journal of the Econometric Society (1978) 33--50.

\bibitem{ngiam2011multimodal}
J.~Ngiam, A.~Khosla, M.~Kim, J.~Nam, H.~Lee, A.~Y. Ng, Multimodal deep learning, in: Proceedings of the 28th international conference on machine learning (ICML-11), 2011, pp. 689--696.

\bibitem{tavakoli2023multi}
M.~Tavakoli, R.~Chandra, F.~Tian, C.~Bravo, Multi-modal deep learning for credit rating prediction using text and numerical data streams, arXiv preprint arXiv:2304.10740 (2023).

\end{thebibliography}





\end{document}